\documentclass[10pt]{article} % For LaTeX2e
\usepackage[preprint]{tmlr}
\usepackage{amsmath,amsfonts,bm}

\def\eqref#1{equation~\ref{#1}}
\def\1{\bm{1}}

\DeclareMathAlphabet{\mathsfit}{\encodingdefault}{\sfdefault}{m}{sl}
\SetMathAlphabet{\mathsfit}{bold}{\encodingdefault}{\sfdefault}{bx}{n}

\usepackage{hyperref}
\usepackage{url}

\usepackage{listings}
\usepackage{array}
\usepackage{graphicx}
\usepackage{booktabs}
\usepackage{tabularx}
\usepackage{amsmath}
\usepackage{soul}
\usepackage{tikz}
\usetikzlibrary{fit,positioning,calc}
\usepackage{threeparttable}
\usepackage{subcaption}

\usepackage{xcolor}
\definecolor{color1}{HTML}{FCB2AF}
\definecolor{color2}{HTML}{9BDFDF}
\definecolor{color3}{HTML}{FFE2CE}
\definecolor{color4}{HTML}{C4D8E9}
\definecolor{color5}{HTML}{38761D}

\renewcommand{\cite}{\citep}
\usepackage[edges]{forest}
\usepackage{tikz}
\usetikzlibrary{trees,positioning,shapes,shadows,arrows.meta}
\usepackage{amsfonts}
\usepackage{multirow}
\usepackage{pifont}
\newcommand{\yes}{{\color{color5}\ding{52}}}
\newcommand{\no}{{\color{red}\ding{55}}}
\tikzset{
    hyperlink node/.style={
        alias=sourcenode,
        append after command={
            let \p1 = (sourcenode.north west),
            \p2=(sourcenode.south east),
            \n1={\x2-\x1},
            \n2={\y1-\y2} in
            node [inner sep=0, outer sep=0, anchor=north west,at=(\p1)]
            {\hyperref[#1]{\XeTeXLinkBox{\phantom{\rule{\n1}{\n2}}}}}
        }
    },
}

\title{A Practical Review of Mechanistic Interpretability for Transformer-Based Language Models
}

\author{\name Daking Rai\thanks{Corresponding authors} \email drai2@gmu.edu \\
      \addr George Mason University
      \AND
      \name Yilun Zhou \email yilun@csail.mit.edu \\
      \addr Datadog AI Research
      \AND
      \name Shi Feng \email shi.feng@gwu.edu\\
      \addr George Washington University
      \AND
      \name Abulhair Saparov \email asaparov@purdue.edu\\
      \addr Purdue University
      \AND
      \name Ziyu Yao$^*$ \email ziyuyao@gmu.edu\\
      \addr George Mason University \\~\\
        \textbf{GitHub Paper Collection:} \href{https://github.com/Dakingrai/awesome-mechanistic-interpretability-lm-papers}{https://github.com/Dakingrai/awesome-mechanistic-interpretability-lm-papers}
      }

\begin{document}

\maketitle

\begin{abstract}
Mechanistic interpretability (MI) is an emerging sub-field of interpretability that seeks to understand a neural network model by reverse-engineering its internal computations. Recently, MI has garnered significant attention for interpreting transformer-based language models (LMs), resulting in many novel insights yet introducing new challenges. However, there has not been work that comprehensively reviews these insights and challenges, particularly as a guide for newcomers to this field. To fill this gap, we provide a comprehensive survey from a \textbf{task-centric} perspective, organizing the taxonomy of MI research around specific research questions or tasks.  We outline the fundamental objects of study in MI, along with the techniques, evaluation methods, and key findings for each task in the taxonomy. In particular, we present a \emph{task-centric taxonomy} as a \emph{roadmap for beginners} to navigate the field by helping them quickly identify impactful problems in which they are most interested and leverage MI for their benefit. Finally, we discuss the current gaps in the field and suggest potential future directions for MI research.
\end{abstract}

\newpage
\tableofcontents
\newpage

\section{Introduction}
In recent years, transformer-based language models (LMs) have achieved remarkable success in a wide range of natural language processing (NLP) tasks~\citep{radford2019language, brown2020language, achiam2023gpt, touvron2023llama, bubeck2023sparks}. Alongside these advancements, there are growing concerns over the safety, reliability, generalizability, and robustness of their usage and development~\cite{bengio2024managing, chang2024survey, yao2024survey, weidinger2022taxonomy}, especially as they are increasingly implemented in real-world applications. These concerns primarily stem from our limited understanding of these LMs and the difficulty in {interpreting their behavior}.

Recently, mechanistic interpretability (MI) has emerged as a promising technique that fills the gap in the research field of interpretability. This line of methods attempts to interpret LM by reverse-engineering the underlying computation into human-understandable mechanisms~\cite{olah2020zoom, elhage2021mathematical}. It has shown promise in providing insights into the functions of LM components (e.g., neurons, attention heads), offering mechanistic explanations for various LM behaviors, and enabling users to leverage the explanations to address LM shortcomings and improve their behavior~\cite{wang2022interpretability, marks2024sparse, templeton2024scaling}. Despite the promise, however, there are concerns about the scalability and generalizability of MI findings, as well as its practical applications in tackling critical problems such as AI safety~\cite{rauker2023toward, casper2023interpretability}.

Observing these promises and challenges, we aim to provide a comprehensive review of MI in its applications to interpret transformer-based LMs. 
{In particular, our survey offers a \textit{task-centric} perspective (see Section~\ref{sec:related_work} for a comparison with related surveys), organizing our survey around
% \zyc{our first version includes a taxonomy. but for this version, what does the taxonomy refer to?} 
MI research that investigates fundamental objects of MI study
% with specific research questions or tasks 
and enabling readers new to the field to quickly identify problems of their interest. Specifically, this is achieved by structuring our taxonomy as a \textit{Beginner's Roadmap to MI} (Section~\ref{sec: roadmap}),  where each category and subcategory are centered on overarching research questions and accompanied by actionable workflows. These workflows outline key steps and applicable techniques, guiding practitioners through the entire process from problem formulation to practical implementation. In addition, we discuss the pros and cons of applicable techniques (e.g., vocabulary project methods~[\citealt{nostalgebraist2020blog}] vs. sparse autonecoders~[\citealt{bricken2023monosemanticity}]) and present readers with case studies of how prior works fit within the workflow, aiming to aid them in understanding the practices and choosing the best-fit approaches to meet their need. By doing so, our survey serves as a friendly and practical guide for beginning researchers (e.g., new doctoral students) to quickly grasp the field of MI, and for experienced practitioners to more easily organize their study plan.}

{Our paper is organized as follows. We first discuss how our survey brings unique perspectives and contributions compared to existing surveys on related topics (Section~\ref{sec:related_work}), and then present the background knowledge needed to understand this survey (Section~\ref{sec:background}). Following this, we introduce the three \emph{fundamental objects of study} of MI, which conceptually structure the research explorations in MI~\cite{olah2020zoom} and also discuss the current (ambiguous) uses of the term ``MI'' in literature (Section~\ref{sec:what-is-MI}). Next, we introduce various techniques used in MI research (Section~\ref{sec: techniques}). For each technique, we introduce (1) its basic concepts and connections to other sub-fields of interpretability or to work that do not target transformer-based LMs, (2) recent advancements, particularly the limitations of the technique and what has or has not been addressed by further research, and (3) technique-specific evaluations. Then, we introduce \emph{Beginner's Roadmap}
, which provides a \emph{task-centric} overview of MI (Section~\ref{sec: roadmap}). Corresponding to the fundamental objects of study, the roadmap summarizes the actionable workflows along three main lines, i.e., the \emph{study of features}, the \emph{study of circuits}, and the \emph{study of universality}. Subsequently, we present case studies {that map the research activities of prior works to the workflow presented in the beginner's roadmap} (Section~\ref{sec: case-studies}). Next, we present an overview of findings and applications for each study in the roadmap (Section~\ref{sec: findings-application}). This section includes an in-depth discussion of the findings and novel contributions of various existing works. Finally, we provide careful discussion on the challenges and issues associated with the current development of MI, as well as suggestions for future work (Section~\ref{sec: discussion-future-works}).}

\vfill

% \begin{figure*}[t!]
%     \centering
%     \includegraphics[width=0.9\textwidth]{Figures/transformer.pdf}
%     \caption{Architecture of transformer-based LMs.}
%     % \vspace{-2mm}
%     \label{fig:transformer-architecture}
% \end{figure*}
\section{Related Work}\label{sec:related_work}

\begin{table}[t!]
    \centering
    \resizebox{\columnwidth}{!}{
    \newcolumntype{x}[1]{>{\raggedright\arraybackslash}m{#1}}
    \begin{tabular}{x{2.5cm} x{2.5cm} x{2.5cm} x{2.5cm} x{2.5cm} x{2.5cm} x{2.5cm}}\toprule
       \textbf{Related Survey} & \textbf{MI Focused?} & \textbf{Format} & \textbf{Techniques} & \textbf{LM Findings} & \textbf{Applications} & \textbf{Future Research} \\ \midrule
     \citet{zhao2024explainability} & \no & Technique-centric & \yes & \no & \yes & \yes \\\midrule
     \citet{dang2024explainable} & \no & Technique-centric & \no & \no & \yes & \yes \\\midrule
     \citet{ferrando2024primer} & \yes & Technique-centric & \yes & \yes & \no & \no \\\midrule
     \citet{bereska2024mechanistic} & \yes & Technique-centric & \yes & \yes & \yes & \yes \\\midrule
     \citet{shu2025survey}$^*$ & \yes (only SAEs) & Technique-centric & \yes & \no & \yes & \yes \\\midrule
     \citet{sharkey2025open}$^*$ & \yes & Task-centric & \no & \yes & \yes & \yes \\\midrule\midrule
     \textbf{Ours} & \yes  & Task-centric & \yes & \yes  & \yes & \yes \\\bottomrule
    \end{tabular}
    }
    \caption{Comparison of our survey with others on LM interpretability. ``MI Focused'': whether the survey has a strong focus on MI or whether it includes little or no coverage of MI research; ``Format'': whether the survey is elaborated from a \emph{technique-centric} or \emph{task-centric} angle (see Section~\ref{sec:related_work} for detail); ``Techniques'': whether the survey explains the involved technical approaches in detail (regardless of its MI/non-MI focus); ``LM Findings'': whether the survey includes findings and scientific discoveries about LMs' intrinsic properties by applying the techniques; ``Applications'': whether the survey covers how the techniques have been applied to practical tasks; and ``Future Research'': whether the survey includes a dedicated section discussing remaining challenges and future work of the field. (*surveys published after our submission.)
    }
    \label{tab:comparison}
\end{table}

{There exist surveys on relevant topics, but they differ from ours in a number of key aspects (Table~\ref{tab:comparison}). First, our survey focus is exclusively on MI for interpreting LMs, in contrast to some existing surveys~\cite{zhao2024explainability, dang2024explainable} that provide a broad overview of LM explainability (e.g., generating textual explanations for LMs). Secondly, among the surveys that focus on MI, one of the key distinctions between the existing surveys and ours is the format or organization of the survey. Specifically, most existing surveys follow an \emph{technique-centric} format where the central focus is on introducing various interpretability and explainability techniques, and they thus organize the survey around the technique family~\cite{zhao2024explainability, bereska2024mechanistic} or their general purpose~\cite{ferrando2024primer, dang2024explainable, shu2025survey}. In contrast, our survey adopts a \emph{task-centric} organization that centers around the practical objectives and workflows involved in MI study. Specifically, the task-centric organization of the survey is enabled by the \emph{Beginner’s Roadmap to MI} (overview in Section~\ref{subsec:roadmap-overview} and details in Section~\ref{sec: roadmap}), which (1) organizes the survey around three fundamental objectives of study in MI, i.e., how to perform \emph{feature study}, \emph{circuit study}, and the study of \emph{universality of features and circuits}---which we consider as three core \emph{tasks} in MI research and practices; (2) elaborates on each task category with its \emph{actionable workflows}, which distill key steps of the task and suggest appropriate MI techniques for each step; and (3) bridges the workflow concepts and practices through \emph{case studies}, showing how these workflows have been applied in prior work to solve real MI problems. 
While \emph{technique-centric} MI surveys such as \citet{bereska2024mechanistic} and \citet{ferrando2024primer} provide thorough introductions to various MI techniques and their technical developments, their technique-centric format does not include step-by-step workflows and illustrative case studies that can guide readers from research goals to method selection. Our survey addresses this gap by combining technical coverage with actionable guidance. We believe that our \emph{task-centric} organization provides a more intuitive entry point for newcomers who may feel overwhelmed by the breadth of technical terminology, while also providing experienced practitioners with a structured framework to align their research goals with the most relevant MI techniques.}

% By grounding the discussion of techniques in actionable workflows and case studies of the workflow, we aim to complement existing surveys with an organization that focuses on how interpretability techniques are applied in practice to address concrete research questions.}

{Beyond its task-centric organization, our survey also differs from other surveys~\cite{dang2024explainable, shu2025survey, sharkey2025open} in comprehensive coverage of MI techniques. Specifically, we explain the technical details of each technique, outline their limitations and technical advancements, and finally show how they are employed in the step-by-step workflows of the \emph{Beginner’s Roadmap to MI}. In contrast, \citet{sharkey2025open} and \citet{dang2024explainable} do not cover the technical introduction of MI techniques, and \citet{shu2025survey} is limited to a single MI technique, Sparse Autoencoder (SAE), which we cover in Section~\ref{sec: sae}. Similarly, our survey also provides 
% a comprehensive \emph{LM findings} on how MI techniques provides 
comprehensive discussions on \emph{LM findings} -- how MI techniques have been applied to discover and understand the properties of LMs, such as the functions of their components and their behaviors. 
This contrasts with surveys such as \citet{zhao2024explainability} and \citet{dang2024explainable} that primarily focus on the advancements and comparison of explanation techniques (partially also due to their lack of focus on MI), and \citet{shu2025survey} that has only a brief discussion about findings on LM behaviors and is limited to the application of SAEs.
% This contrasts with surveys like \citet{zhao2024explainability} and \citet{dang2024explainable} that primarily focus on the comparison and advancements of explanation techniques or \citet{shu2025survey} that have limited coverage. 
In addition, our survey also highlights studies that have applied MI to practical downstream applications such as knowledge editing, LM steering, and AI safety, which were addressed to a lesser extent in \citet{ferrando2024primer}.
% (among surveys focusing on MI \yilun{Most of the other surveys in Table 2 have this though?}).
% which have been addressed to a lesser extent in existing surveys~\cite{zhao2024explainability, ferrando2024primer}. 
Finally, our survey concludes by providing novel discussions on the important challenges associated with the current development of MI and directions for future research, in line with existing surveys.

\section{Background: Transformer-based LMs}
\label{sec:background}

% \label{sec: transformer-architecture}
\begin{figure*}[t!]
    \centering
    \includegraphics[width=\linewidth]{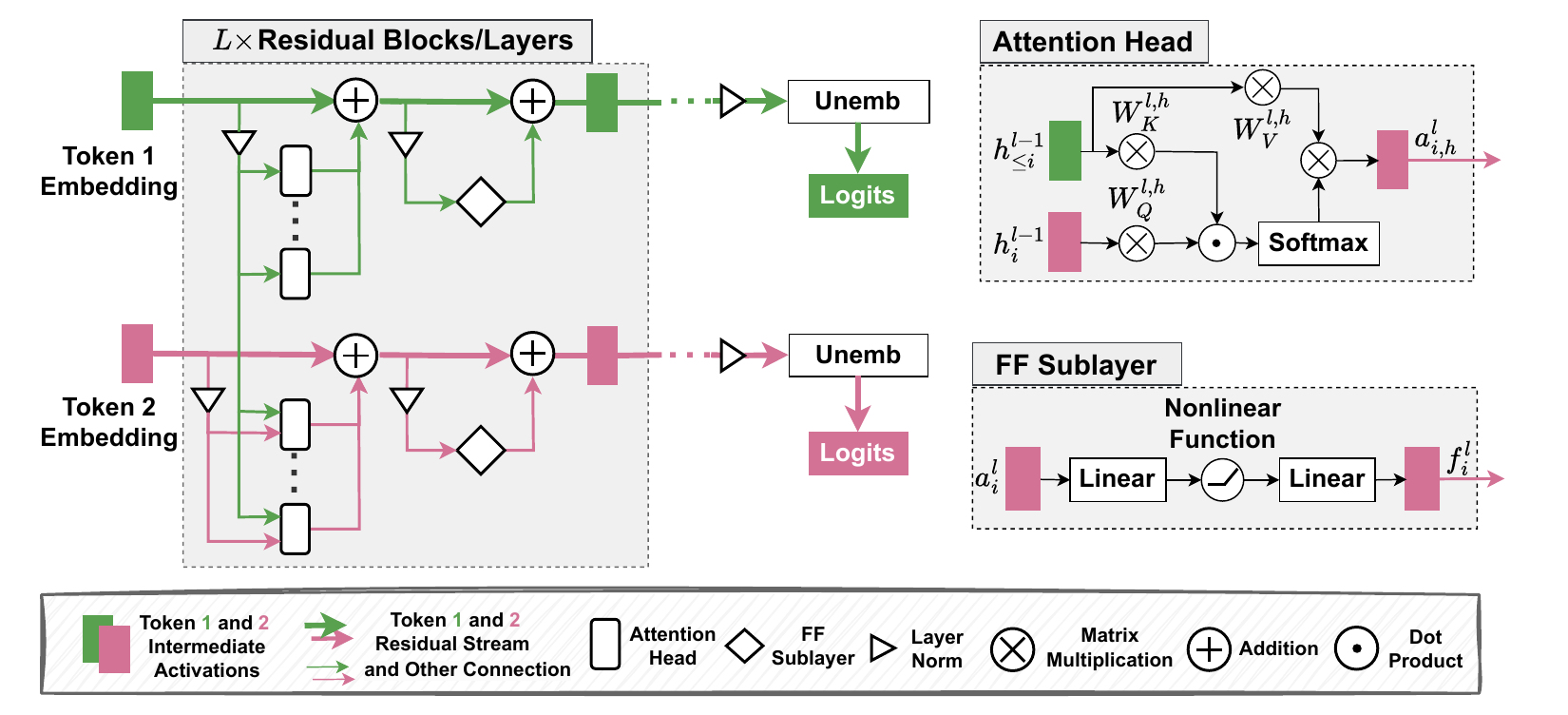}
    \caption{Architecture of transformer-based LMs.}
    \label{fig:transformer-architecture}
\end{figure*}
A transformer-based LM~\cite{vaswani2017attention} $M$ takes input tokens $X = (x_1,...,x_n)$ and outputs a vector in $\mathbb{R}^{|\mathcal{V}|}$, a probability distribution over the vocabulary $\mathcal{V}$, to predict the next token $x_{n+1}$. The model refines the representation of each token $x_i$ layer by layer (Figure~\ref{fig:transformer-architecture}). In the first layer, $h_i^0$ is an embedding vector of $x_i$, resulting from a lookup operation in an embedding matrix $W_E \in \mathbb{R}^{|\mathcal{V}| \times d}$.\footnote{For brevity, we omit components such as position embedding and layer normalization in transformer, as they will not affect our discussion of MI. Readers should refer to~\citet{vaswani2017attention} for a complete description.}

This representation is then updated layer-by-layer through the calculations of \emph{multi-head attention (MHA)} and \emph{feed-forward (FF)} sublayers in each layer, i.e.,
\begin{equation}
h_i^l = h_i^{l-1} + a_i^l + f_i^l, \label{eq: rs}
\end{equation}
where $h_i^l$ denotes the representation of token $x_i$ at layer $l$, $a_i^l$ is the attention output from the MHA sublayer, and $f_i^l$ is the output from the FF sublayer. The sequence of $h_i^l$ across the layers is also referred to as the \emph{residual stream} (RS) of the transformer in literature~\cite{elhage2021mathematical}.

Briefly, the MHA sublayer with $H$ attention heads is implemented via
\begin{align}
a_i^l &= \texttt{concat}(a_{i,0}^l, .., a_{i,H}^l)W_O^{l}, \\
a_{i,h}^l &= \texttt{softmax}\Bigg(\frac{(h_i^{l-1}W_Q^{l,h}) (h_{\leq i}^{l-1}W_K^{l, h})^\top}{\sqrt{d_k}}\Bigg) \nonumber \cdot (h_{\leq i}^{l-1} W_V^{l,h}), % \nonumber
\end{align} 
where $a_{i,h}^l$ is the attention output from the $h$-th head, $W_Q^{l,h}, W_K^{l,h}, W_V^{l,h}$ are the query, key, and value (learned) projection matrices, respectively, and $W_O^{l}$ projects the concatenated attention outputs from all heads to the model dimension $d$. 

The FF sublayer then performs two linear transformations over $h_i^{l-1} + a_i^l$ with an element-wise non-linear function $\sigma$ between them, i.e., 
\begin{equation}
    f_i^l = W_v^l \sigma \Big(W_k^l (h_i^{l-1} + a_i^l) + b^l_k \Big) + b^l_v,
\end{equation}
where $W_v^l$, $W_k^l$, $b^l_k$, and $b^l_v$ are learned parameter matrices and biases.
Finally, the RS of $x_n$ at the final layer, $h_n^L$, is projected into a probability distribution over $\mathcal{V}$ by applying an \emph{unembedding} matrix \(W_U \in \mathbb{R}^{d \times |\mathcal{V}| }\) and a softmax operation. 
\section{What is Mechanistic Interpretability?}\label{sec:what-is-MI} 
The goal of MI is to reverse-engineer the detailed computations performed by a model into human-understandable algorithms, similar to how a programmer might try to reverse-engineer complicated binaries into human-readable source code. To achieve this goal, MI takes a bottom-up approach by decomposing the model into smaller components and more elementary computations~\cite{zou2023representation}. By understanding these smaller components and their interactions, MI aims to build a comprehensive understanding of the full model. 
Below, we first present the three fundamental objects of study in MI, then give an overview of our devised \emph{beginner's roadmap to MI}, which summarizes the actionable workflows of current MI studies towards addressing the three fundamental objects, and finally discuss its connections and differences with respect to other sub-fields of AI interpretability.
% Specifically, MI decomposes the LMs into two fundamental objects of study, features and circuits, as discussed in Section~\cite{sec:objects of study}.}

\subsection{Fundamental Objects of Study in MI} \label{sec:objects of study}
% The fundamental objects of study of MI include \emph{features} and \emph{circuits}, as well as their properties \emph{universality} and the \emph{learning dynamics}.
Following~\citet{olah2020zoom}, one of the earliest studies in MI, we categorize research of MI into three areas: the study of features, circuits, and their universality.

\subsubsection{Features} \label{subsec: features}
A \emph{feature} is a human-interpretable input property that is encoded in LM activations.\footnote{We use the terms ``representations'' and ``activations'' interchangeably.} For example, when we provide an LM with the input token \emph{``dog''}, it may extract features like \emph{``animal''}, \emph{``pet''}, \emph{``has four legs''}, and other relevant features learned during pre-training, embedding these features in its activations. MI aims to interpret the LM representations by decoding the features encoded in them.

% To this end, many MI studies postulate the \emph{linear representation hypothesis}, which posits that neural networks have two properties -- \emph{linearity}, i.e. the network's activation space consists of meaningful (linear) vectors, each representing a feature, and \emph{decomposability}, i.e., network activations can be decomposed and described in terms of these independent features~\cite{elhage2022superposition}. This hypothesis resonates with earlier research that also shows linearity in word embeddings (e.g. $\overrightarrow{man\vphantom{k}} - \overrightarrow{woman\vphantom{k}} \approx \overrightarrow{king} - \overrightarrow{queen\vphantom{k}}$) ~\cite{mikolov2013distributed}. In Section~\ref{subsec: findings-features}, we will review some identified features in LMs and how they are represented. 

\subsubsection{Circuits} \label{subsec: circuits}

While the study of features helps us to understand what information is encoded in a model's activations, it does not inform us of how these features are extracted from the input, their interactions, or how they are used by the model to enable specific LM behaviors (e.g., reasoning). MI bridges this gap by investigating \emph{circuits} -- meaningful computational pathways that connect features and facilitate specific LM behaviors.

More formally, if we view an LM $M$ as a computational graph with features as nodes and the weighted connections between them as edges, a circuit is a sub-graph of $M$ responsible for implementing specific LM behaviors~\cite{olah2020zoom}.  
% \zyc{Why did we remove the following sentences? They are necessary for transitioning from "features" to "circuits".} 
Additionally, although circuits were initially defined as connections between features~\cite{olah2020zoom}, subsequent studies have generalized them as connections between the activation outputs of \emph{transformer components}~\cite{olsson2022context, wang2022interpretability}, where interpreting individual transformer components becomes part of the circuit interpretation. Therefore, we include research on interpreting individual transformer components as part of the study of circuits as well.
As we will introduce in Section~\ref{subsec:roadmap-circuit-study}, prior works have defined circuits with various assumptions and at different levels of granularity. Some works even generalized the definitions of circuits to cover general information flows in an LM~\cite{geva2023dissecting, nikankin2024arithmetic}. In this survey, we will consider all of these works as studies of circuits, as long as the goal is to find computational pathways connecting transformer components or features in an LM.

An example circuit discovered by \citet{elhage2021mathematical} in a toy LM is shown in Figure~\ref{fig:circuit}. This is an induction circuit consisting of two attention heads (previous token head and induction head) as nodes and input/output activations between them as edges of the circuit. The circuit implements the task of detecting and continuing repeated subsequences in the input (e.g., {``Mr D urs ley was thin and bold. Mr D'' -> ``urs''}), where the \emph{previous token head} encodes the information that \emph{``urs'' follows the ``D'' token} in the RS, which is then read by the \emph{induction head} to promote {``urs''} as the next token prediction.

\begin{figure*}[t!]
    \centering
    \includegraphics[width=0.9\textwidth]{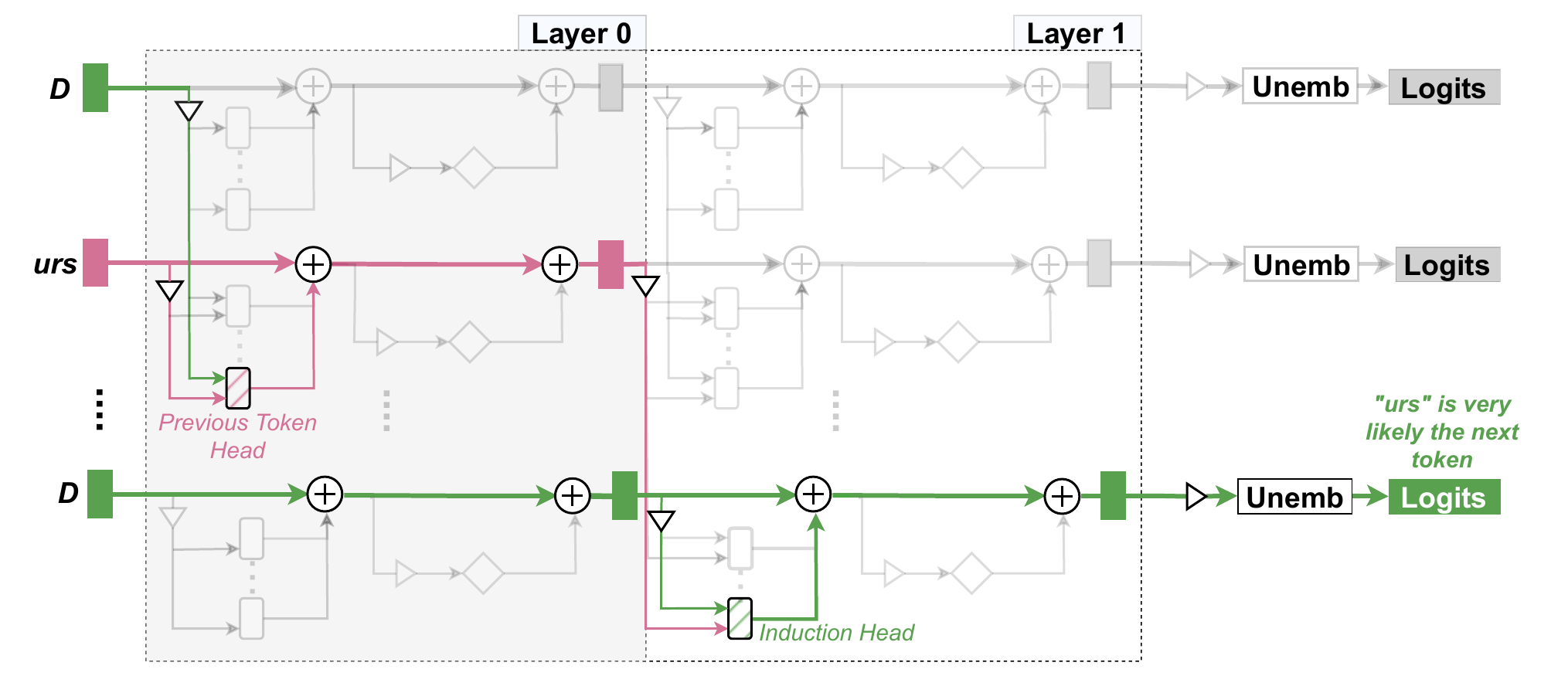}
    \caption{An example of an induction circuit discovered by \citet{elhage2021mathematical} that consists of outputs of two attention heads (previous token head and induction head) as nodes and connection between them as edges.}
    \label{fig:circuit}
\end{figure*}

% \daking{``pruning'' research such as \citet{xia2022structured} is related to circuit study!}

\subsubsection{Universality}  \label{subsec: universality}
For any feature or circuit that we have identified in an LM in one task, the critical question arises: \emph{Do similar features and circuits exist in other LMs or tasks?}
The investigation into this question has then given rise to the notion of \emph{universality}, i.e., {the extent to which} similar features and circuits are formed across different LMs and tasks~\cite{olah2020zoom, gurnee2024universal}. The implications of universal features and circuits can be significant. For instance, many studies on features and circuits~\cite{olsson2022context, elhage2022solu, elhage2022superposition} were performed with only toy or small LMs. If these features and circuits are universal, the insights from these studies can be transferred to other unexamined LMs and potentially state-of-the-art large LMs (LLMs). However, if they are not universal, then a significant amount of independent effort will be required to interpret each uninterpreted LM in each task.

\subsection{An Overview of \emph{Beginner's Roadmap to MI}}\label{subsec:roadmap-overview}
\begin{figure*}[t!]
    \centering
    \includegraphics[width=0.9\textwidth]{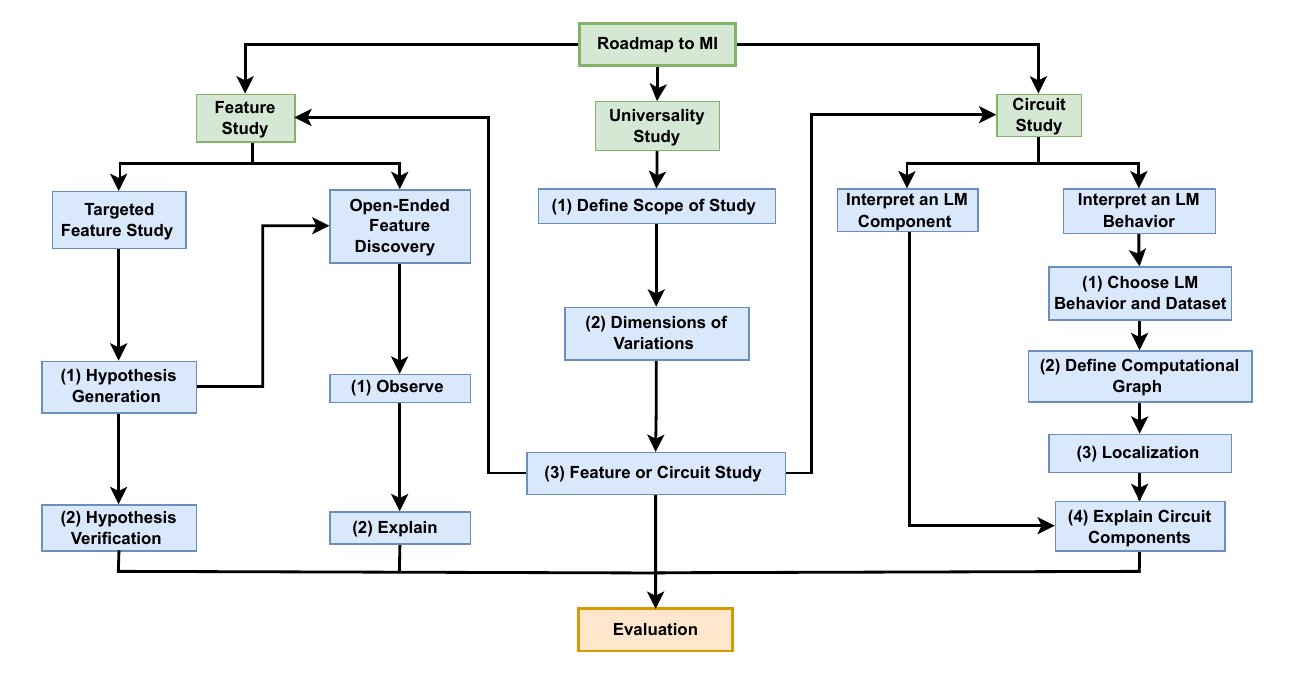}
    \caption{Beginner’s roadmap to MI, designed to help newcomers quickly pick up the field. The MI study is organized into three main categories -- feature study, circuit study, and universality study, corresponding to the fundamental three objects of MI study. The roadmap also provides a step-by-step actionable workflow for each category that distills the key steps for completing tasks in that category. More detailed figures and explanations for each category, together with the techniques available at each stage and their respective advantages and disadvantages, are provided in Section~\ref{sec: roadmap}.}
    \label{fig:overall-roadmap}
\end{figure*}
% TODO: Give an overview of the roadmap in one or two paragraphs. Add a new overview fig summarizing the three roadmaps (you can simply merge the three separate roadmaps and remove the sub-categories and approach details). Do not use subsections.

The MI community has explored various approaches to study the three fundamental objects of study. 
{A key motivation for this survey is to provide a friendly guide for researchers and developers interested in MI to quickly pick up the field. To achieve this, we organize our survey in a \emph{task-centric} format, which is enabled by \emph{beginner's roadmap to MI}. We present an overview of the roadmap in Figure~\ref{fig:overall-roadmap} and elaborate on its details in Section~\ref{sec: roadmap}. The roadmap categorizes MI research into three studies, corresponding to the three fundamental objects, i.e., feature study, circuit study, and the study of universality.
% categories (or practical objectives of MI) -- feature study, circuit study, and the study of universality, corresponding to the three objects of study.} 
{We provide a general step-by-step workflow for each category. For instance, we propose to further divide \emph{feature study} into two sub-categories, each with its own distinct workflow: \emph{targeted feature study} and \emph{open-ended feature study}. A \emph{targeted feature study} consists of two steps: \emph{Hypothesis generation}, in which we propose what feature may be present in the activations, and \emph{Hypothesis validation}, in which we test whether the feature is indeed represented. In contrast, an \emph{open-ended feature study} assumes no prior knowledge of what features are encoded and involves two steps: \emph{Observe}, where we collect information about model activations, and \emph{Explain}, where we annotate the features encoded based on the information gathered during the \emph{Observe} step.} 
{Similarly, we further categorize \emph{circuit study} into two sub-categories, each with its own workflow: \emph{interpret an LM behavior} and \emph{interpret an LM component}. To \emph{interpret an LM behavior}, the workflow consists of four steps: \emph{Choose LM behavior and dataset}; \emph{Define computational graph} of the selected LM; \emph{Localization} of the important nodes and edges of the circuit; and \emph{Explain circuit components} identified during the \emph{Localization} step. On the other hand, to \emph{interpret an LM component}, one can begin directly from the \emph{Explain circuit components} step.} 
{Finally, the \emph{universality study} workflow consists of three steps: \emph{Define scope of study}, which can be considered along two dimensions—universality of features or universality of circuits; \emph{Dimensions of variations}, referring to universality across LMs or across tasks; and conduct \emph{Feature or circuit study} based on the defined scope and variations.} 
{Importantly, all three categories also require a final \emph{Evaluation} step to verify the results of the study. The detailed workflows, together with the techniques available at each stage and their respective advantages and disadvantages, are presented in Section~\ref{sec: roadmap}. We further provide case studies in Section~\ref{sec: case-studies} that illustrate how prior work has applied the proposed roadmap.}

\subsection{How is MI Different From Other Sub-fields of Interpretability?}
{Although MI has become a popular sub-field within interpretability, it is still arguable what exactly makes an interpretability study \emph{mechanistic}. Historically, the term \emph{Mechanistic Interpretability} was coined to distinguish the field from interpretability approaches, such as saliency methods~\cite{ribeiro2016should, sundararajan2017axiomatic, lundberg2017unified}, that generated explanations solely by analyzing inputs and outputs and provided little to no insight into the internal mechanisms of the model~\cite{olah2020zoom, saphra2024mechanistic}. While this distinction clarifies that MI involves investigating the internals of a model, it does not address how MI is distinct from other interpretability research that also explores model internals. For instance, even prior to MI there was already a significant body of work, such as probing~\cite{belinkov2022probing} and attention analysis~\cite{vig2019visualizing}, that investigated the internal representation of models. Relevant to this discussion, \citet{zou2023representation} categorized interpretability work that investigates the internals of models into two broad categories, i.e., \emph{top-down} and \emph{bottom-up} approaches, with MI classified as bottom-up. Specifically, the \emph{bottom-up} approach begins by breaking down the model into its smallest units of analysis, such as neurons, aiming to first understand these components before piecing them together to understand how they combine to produce the model's overall behavior. In contrast, the \emph{top-down} approach starts with higher-level model behaviors or coarser components and works downward to explore finer-grained mechanisms. Despite these distinctions, in practice, the boundaries between the two approaches often blur, as both ultimately involve studying neurons and representations within models.}
% This lack of clarity has caused significant confusion and frustration within the community. MI studies often fail to connect with prior literature, leading to reinvented techniques, overlooking existing work as baselines, and missing opportunities for cumulative progress. To address this, \citet{saphra2024mechanistic} examined the historical evolution of the field and proposed four definitions of MI. In our survey, adopt the \emph{broad technical} definition from \citet{saphra2024mechanistic} and define \emph{MI to be any work that tries to reverse-engineer the internals of a model}. Furthermore, we aim to clarify existing confusion, draw connections to other related subfields, and offer a cohesive understanding of the techniques and findings within MI research.
 {This lack of clarity has led to confusion within the community, with several researchers publicly expressing concerns that many MI studies are ``not new'' and often disregard prior literature, reinvent or rediscover existing techniques, or overlook important baseline work~\cite{saphra2024mechanistic}. To provide a clearer historical context of MI, \citet{saphra2024mechanistic} examined the historical evolution of the field and proposed four definitions of MI. In our survey, we adopt the \emph{broad technical definition of MI} from \citet{saphra2024mechanistic} and define \emph{MI as any work that seeks to describe the internals of a model}. Additionally, we also urge the MI community to bridge the existing disconnect by more actively integrating studies from related non-MI research that explores similar topics.
 
 % Additionally, we aim to bridge MI research with the broader field of interpretability to foster a more cohesive understanding. 
 
 % In Section~\ref{subsec: mi-connect-broad}, we will further list the various areas of MI study in connection to the broader field of interpretability that investigates the same problem to illustrate their connections.
 
 % the challenge caused by the confusion of the definition of MI and the overlook of its connection to history 
 
 % \abu{what do we mean here by ``furthering the challenge''? is the intended meaning something like: we aim to clear the confusion caused by the non-standard definition of MI and to better situate MI research historically? (though my suggestion also may need rephrasing/improvement)}.
 % we will \zyc{revise this sentence:}discuss past and concurrent non-MI work relevant to MI.
 % Additionally, we aim to bridge MI research with the broader field of interpretability to foster a more cohesive understanding by providing relevant past and concurrent non-MI work to MI in Section~\ref{?}.
 }

\section{Techniques and Evaluation Methods}\label{sec: techniques}
\begin{table*}[t!]
\centering
\resizebox{\textwidth}{!}{
\begin{tabular}{
    >{\raggedright\arraybackslash}p{1.6cm}
    >{\raggedright\arraybackslash}p{4cm}
    >{\raggedright\arraybackslash}p{4cm}
    >{\raggedright\arraybackslash}p{5.5cm}
}
\toprule
\textbf{Techniques} & \textbf{Basic Concepts} & \textbf{Technique-specific Interpretation} & \textbf{Technical Advancements} \\
\midrule
Vocabulary Projection Methods (Sec~\ref{tech: vocab-proj-methods}) & 
Decode features from activations by projecting them into the vocabulary space, which is typically achieved by multiplying the activations with the unembedding matrix $W_U$, yielding logit distributions over the vocabulary. 
&
Infer encoded features by inspecting top/bottom tokens in the projected logit distribution (e.g., a majority of top-k tokens relate to ``breakfast'' implies a ``breakfast'' feature). &
(1) Reliability: improve faithfulness of projections, (2) Decoding positions: enable decoding from activations, weights, and gradients, and (3) Expressivity: decode features beyond next-token predictions. \\
\midrule
Intervention-based Methods (Sec~\ref{tech: intervention-methods}) & 
Investigate LM behavior causally by intervening on intermediate activations during the forward pass and comparing outputs before and after the intervention to infer their role. There are two types: \emph{noising-based} and \emph{denoising-based} interventions. 
&
Evaluate intervention effects by measuring changes in target token logit, target token probabilities, or the full logit distribution before and after intervention. If an activation is critical, intervening in its value should yield substantial performance changes. &
(1) Reliability of corruption: improve reliability by ensuring interventions do not take the model out of distribution, (2) Intervention for localization: adapting intervention to discover all behavior-related components, (3) Scaling up and automation: automate the localization technique to alleviate human effort, (4) Intervention for validating hypotheses, and (5) Integrate with vocabulary projection for richer activation decoding. \\
\midrule
Sparse Autoencoders (SAE, Sec~\ref{sec: sae}) & Map activations to a more interpretable but higher-dimensional sparse activations by training SAEs; one can then conduct feature studies in the sparse activations instead of original LM activations. & Fidelity of the sparse activations is measured by the reconstruction loss {and the interpretability of SAE features is evaluated manually by looking at the activation pattern of the feature or by automated interpretability score}.
% , and \zyc{is this true? It seems to contradict results/motivations of SAEBench etc.}more sparse activations are generally considered more interpretable. 
& (1) Balancing reconstruction fidelity with sparsity for interpretability, and (2) Discovering features that are functionally useful for downstream LM tasks. \\
\midrule
Probing (Sec~\ref{subsec: probing-tech}) & Test whether predefined features are present in LM activations by training a shallow classifier (i.e., \emph{probe}) on them. & High probe accuracy on held-out test set indicates feature presence on activations. & (Not a focus in this survey; we refer readers to \cite{belinkov2022probing} for more details.) \\
\midrule
Visualization (Sec~\ref{subsec: visualization}) & Prepare visualization (e.g., attention maps, neuron activation plots) to investigate model behavior. & Human examines visual information to form or refine hypotheses about the roles of components and activations. & (Not a focus in this survey.) \\
\bottomrule
\end{tabular}
}
\caption{Summary of MI techniques discussed in Section~\ref{sec: techniques}, along with their basic concepts, technique-specific interpretation (i.e., approaches to evaluate the technique outputs), and key technical advancements.}
\label{tab:techniques-summary}
\end{table*}

In this section, we review major techniques that have been developed to study the fundamental objects described in Section~\ref{sec:objects of study} for understanding transformer-based LMs. During the introduction of each technique, we will present its basic concepts, technique-specific interpretation, and recent advancements. A summary is also present in Table~\ref{tab:techniques-summary}.

\subsection{Vocabulary Projection Methods} \label{tech: vocab-proj-methods}

\subsubsection{Basic Concepts}
This class of techniques aims to decode the information within the LM representation by projecting it to the model's vocabulary space. Typically, during the LM inference, only the activation of the final layer ($h_i^L$) is multiplied by the unembedding matrix ($W_U$) to calculate the logit distribution for the next token prediction. The vocabulary projection method, however, proposes to multiply $W_U$ with intermediate representations (e.g., $h_i^l, 0 \leq l < L$), generating logit distributions from these intermediate layers as well.
By examining tokens with the highest logit values, one can infer the candidate tokens the LM is considering for the next prediction within that specific intermediate representation. 
% \abu{probably phrased too strongly? the intermediate representations may contain more information beyond that which this method provides. maybe instead we can say ``one can then infer some of the information in the intermediate representations''}.
% Once the logit distribution is obtained, the intermediate activations can be directly interpreted in the vocabulary space, as each logit corresponds to a token in the vocabulary $\mathcal{V}$, where the logits with higher values being considered as the candidates for next token prediction and vice versa. 

This technique can be viewed as a practical application of the \emph{iterative inference perspective}~\cite{jastrzkebski2017residual, elhage2021mathematical, geva2020transformer} on how an LM makes predictions, where each layer of an LM is seen as progressively refining a latent prediction of the next token, and the vocabulary projection technique enables us to examine how these predictions are refined and evolve across the model's intermediate layers.

\begin{figure}[t!]
    \centering
    \includegraphics[width=0.5\linewidth]{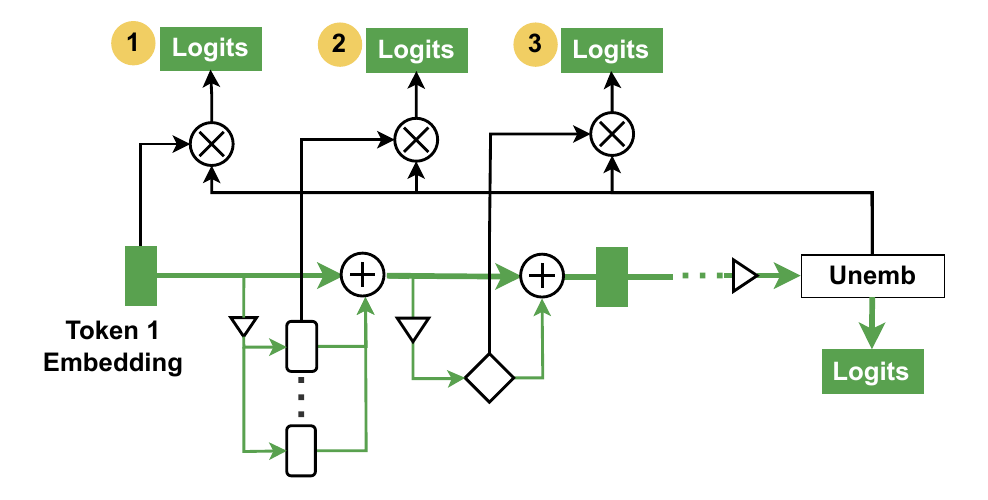}
    \caption{Logit lens implementation at (1) RS, (2) attention head, and (3) FF sublayer.}
    \vspace{-2mm}
    \label{fig:logit-lens}
\end{figure}

\textit{Logit lens}~\cite{nostalgebraist2020blog} was the first technique that proposed to employ $W_U$ for projecting the intermediate activations of an LM (GPT-2~[\citealt{radford2019language}] in the original work) to the vocabulary space. We illustrate the method in Figure~\ref{fig:logit-lens}. Specifically, given an activation $h_i^l$ in the residual stream, logit lens calculates the following:
% \citet{nostalgebraist2020blog} proposed to project the activation $h_i^l$ to vocabulary space as follows:\zyc{why do we highlight the layer norm here?} 
% \daking{This is the original formulation of LogitLens. It corresponds to the LayerNorm after the final transformer layer and before the unembedding operation.}\zyc{1) Do people apply LayerNorm in practice when they perform logit lens? If not, better include a footnote with clarification. 2) Is layer norm applied when projecting the activations of MHA and FF?} \daking{Yes, they are used in practice and also they are used when projecting the activation of MHA and FF as well. We can think of this layernorm as layernorm after the final layer. So, it's outside all the transformer layers and before unembedding matrix.}
\begin{equation}
\texttt{LogitLens}(h_i^l) = \texttt{LayerNorm}(h_i^l)W_U
\end{equation}
Similarly, the logit lens was also applied to project the output activations of MHA ($a_i^l$) and FF sub-layers ($f_i^l$) to the vocabulary space before they are added to the residual stream, so as to interpret the contribution of MHA and FF for the next token prediction. 
% After the projection, one can manually inspect the top tokens with the highest logits to under the LM's next token prediction and analyze how the LM's prediction evolves across layers.

\subsubsection{Technique-specific Interpretation}
As described earlier, once the logit distribution projected from the intermediate activation of interest is obtained, one can infer some of the information encoded within activation by examining the tokens with the highest logit values. This interpretation process is typically performed by human evaluators.
% The outputs of the vocabulary projection methods are logit distributions, which a human evaluator interprets by manually analyzing the tokens with the highest logits. 
For example, if the top tokens primarily relate to items that are associated with breakfast (e.g., pancakes, coffee, bread), it suggests that the activation encodes the ``breakfast'' feature.  To automate this manual evaluation process, \citet{bills2023language} proposed using a large LM (e.g., GPT-4) to automatically generate textual explanations based on the output of the logit lens. Similarly, \citet{rai2024investigation} proposed using an LM to automatically classify whether the logit distribution likely encodes a pre-defined feature category (e.g., arithmetic addition). Lastly, it is important to note that the explanation generated by vocabulary projection methods is only correlational rather than causal in nature~\cite{katz2024backward}, requiring caution when drawing inferences from these methods.

\subsubsection{Technical Advancements}

Since the introduction of the logit lens, several advancements have been made to enhance its capabilities or to address its limitations, leading to the development of various vocabulary projection methods. Specifically, the advancements can be categorized into the following three categories -- (1) Reliability: improving the faithfulness of the technique's output and enabling more reliable vocabulary projection
(2) Decoding position: adapting and applying the technique to decode features from various intermediate representations and model weights; and (3) Decoding expressivity: enhancing the degree to which the technique decodes features beyond immediate next-token predictions. We highlight representative approaches discussed in the technical advancements in Table~\ref{tab:vocab-tech-summary}.
% \daking{We summarize the technical advancements of vocabulary projection methods in Table~\ref{tab:vocab-tech-summary}.}

\begin{table*}[!t]
\centering
\resizebox{\textwidth}{!}{
\begin{tabular}{
    >{\arraybackslash}p{3.0cm}
    >{\arraybackslash}p{7.0cm}
    >{\arraybackslash}p{5.0cm}
}
\toprule
\textbf{Approach} & \textbf{Basic Concepts} & \textbf{Technical Advancements} \\
\midrule
Logit Lens + Translators~\cite{belrose2023eliciting, din2023jump, sakarvadia2023attention} & 
Propose to use \emph{learned translators} for mapping the intermediate activations to the final-layer representation space before applying the logit lens. &
Improve the \emph{reliability} of logit lens projection. \\
\midrule\midrule
Attention Lens~\cite{sakarvadia2023attention} & 
Interpret attention head outputs with the logit lens and learned head-specific translators. &
Adapt the logit lens to new \emph{decoding positions} (i.e., activations of attention heads). \\
\midrule
\citet{geva2022transformer} & {Showcase that the logit lens can also be used to decode the knowledge stored in the FF sub-layer by projecting the columns of the parameter matrix $W_v^l$ in vocabulary space.} & Adapt the logit lens to new \emph{decoding positions} (i.e., model weights). \\
\midrule
Backward Lens~\cite{katz2024backward} & 
Investigate the knowledge updates in FF neurons during model training by projecting their \emph{gradients} into the vocabulary space. &
Adapt the logit lens to new \emph{decoding positions} (i.e., model gradients). \\
\midrule\midrule
Attribute Lens~\cite{hernandez2023linearity} & 
Decode \emph{objects} encoded for a \emph{relation} from \emph{subject activations} (e.g., ``Space Needle — is located in → Seattle'') by learning a relation-specific linear map and using $W_U$ to project in vocabulary space. &
Improve \emph{expressivity} of technique beyond decoding next-token prediction. \\
\midrule
Future Lens \cite{pal2023future} & 
% Analyze whether an $n$-timestep activation encodes future output tokens beyond $n\ge2$ timesteps by training the linear map that maps $h_i^l$ to $h_{i+n}^L$ and using $W_u$ to project into vocabulary space. 
Analyze whether an activation at timestep $i$ encodes future output tokens at $i+n, n \geq 2$, with two types of approaches: learning a linear mapping from  $h_i^l$ to $h_{i+n}^L$ for projection, and patching $h_i^l$ to a separate LM under a pre-defined or learnable investigation prompt.
&
Improve \emph{expressivity} of technique beyond decoding next-token prediction. \\
\bottomrule
\end{tabular}
}
\caption{Representative approaches discussed in the technical advancements of vocabulary projection methods.}
\label{tab:vocab-tech-summary}
\end{table*}

\paragraph{Reliability}
\citet{belrose2023eliciting, din2023jump} noted that the logit lens may produce unreliable projection for certain LMs (e.g., BLOOM, OPT 125M), particularly in the earlier layers. For example, the top-1 projected token for BLOOM \citep{le2023bloom} in more than half of the earlier layers is often the input token, rather than a plausible continuation towards the final output token. The authors posit that the logit lens makes an oversimplified assumption that all layers operate in the same embedding space, which may not be true. Consequently, $W_U$, which is only pre-trained to project an LM's final-layer activation, cannot reliably project all the intermediate activations to the vocabulary space. This observation inspired \citet[i.e., \textit{tuned lens}]{belrose2023eliciting} and \citet{din2023jump} to train translators that transform the intermediate activations to an output representation that better aligns with the representation space of the final layer before the logit lens is applied. 
Specifically, both approaches linearly transformed the intermediate activation (e.g., $h_i^l$) to an activation that matches more closely with the basis of the final layer activation ($h_i^L$). However, \citet{din2023jump} achieved it by directly minimizing the difference between the transformed activation and the final-layer activation, while \citet{belrose2023eliciting} designed the loss function to encourage the matching between the logit distributions of the transformed activation and the final-layer activation. 
% \dakingrev{Specifically, \citet{din2023jump} proposed to learn layer-wise linear mapping as translators, which aims to transform any intermediate activations (e.g., $h_i^l$) to final layer activation $h_i^L$. Similarly, \citet{belrose2023eliciting} proposed the \textit{tuned lens} method that learns a layer-wise affine transformation as translators, which transforms intermediate activation $h_i^l$ to activation that matches closely with the basis of the final layer activations.}
Finally, \citet{sakarvadia2023attention} proposed the \emph{attention lens}, which similarly involved training translators for projecting output activations of attention heads with the logit lens.  

\paragraph{Decoding Position} 
% \zyc{Just to reiterate the comment: the writing should be clear if an approach can only be applied to a certain position (about its "capability limitation"), or has only been applied to the position (about its "application"). }
While the logit lens was initially proposed to decode an LM's prediction information from \emph{activations}, subsequent work has {applied or} adapted the technique to decode information from \emph{model weights} and \emph{gradients}. We summarize the various decoding positions as follows. (1) \textbf{Activations:} {In addition to the logit lens~\cite{nostalgebraist2020blog}, the \textit{tuned lens}~\cite{belrose2023eliciting} and the approach of \citet{din2023jump} were both proposed to project the activations of RS, FF, and MHA sub-layers into the vocabulary space. In contrast, the \textit{attention lens}~\cite{sakarvadia2023attention} was proposed specifically for projecting the output activations of individual attention heads within MHA sub-layers.} (2) \textbf{Model weights:} {\citet{geva2022transformer} showed that the logit lens can also be used to interpret the \emph{weights} of transformer components. Specifically, \citet{geva2022transformer} used it to understand the role of the FF sublayer in the LM's prediction by projecting columns of the parameter matrix $W_v^l$ to the vocabulary space. \citet{dar2022analyzing} further proposed to apply the technique to project the FF sub-layers ($W_k^l$ and $W_v^l$) and the query-key ($W_Q^{l,h} W_K^{l, h}$) and value-output ($W_V^{l,h} W_O^{l}$) interaction parameter matrices of individual attention heads to the vocabulary space, aiming to study how they transfer and mix information from source tokens to the target token.} (3) \textbf{Gradients:} As a representative work, \citet{katz2024backward} proposed \textit{backward lens}, which projects the gradient of FF sub-layers during back-propagation to the vocabulary space, enabling the investigation of how new information is stored in the FF sub-layers of LMs during training. 

% Existing work applying vocabulary project methods can be categorized based on where they decode the information....[talk about the three categories: activations, parameters, gradients]

% \subsubsection{Improving reliability across different LMs and layers} 
 
% \zyc{About writing: the use of italic font, bold font, and capitalization should be more consistent. Similarly, whether you make the first presentation of an approach in italic font or not, should be made consistently. In the paragraph, I can understand why you bold the text of "train" and "training", but readers won't. If you want to highlight anything, say it explicitly. The "tuned lens" and "attention lens" approaches have been introduced before; i don't see the reason to use italic fonts here, but if you have a consistent and clear logic of the font style, it is fine.}

\paragraph{Decoding Expressivity} {Several methods have extended the logit lens to decode information beyond the immediate next-token prediction. For instance, \citet{pal2023future} proposed the \emph{linear model approximation} technique that can be used to determine if an intermediate activation $h_i^l$ at $i^{th}$ decoding timestep has already encoded information about the future output tokens $h_{i+n}^L$, where $n\geq2$. Specifically, \citet{pal2023future} learned a linear model that transforms $h_i^l$ to $h_{i+n}^L$ and then used the unembedding matrix $W_U$ to project the transformed activations to vocabulary space.}
\citet{pal2023future} further proposed two other approaches integrating vocabulary projection with intervention, including the \emph{future lens} approach, and \citet{ghandeharioun2024patchscope} concurrently proposed a generalized framework called \emph{Patchscope}, which similarly performs intervention while projecting an activation to the vocabulary space; we will discuss both approaches in detail when we introduce intervention-based methods (Section~\ref{tech: intervention-methods}).
Similarly, \citet{hernandez2023linearity} proposed the \textit{attribute lens} that can be used to decode which objects (e.g., ``Seattle'') are encoded in the activation for a relation (e.g., ``is located''), given a subject (e.g., ``The space needle'') as the model input. {Specifically, they proposed to learn a linear function for a given relation, which can transform the intermediate activation when the subject is used as input into an object representation. Subsequently, the transformed object representation can be decoded into a vocabulary space or object-token distribution using the unembedding matrix $W_U$.}
Finally, \citet{cancedda2024spectral} introduced \emph{Logit Spectrology}, a method that applies singular value decomposition to $W_U$. Particularly, they found that the logit lens projection is dominated by the larger singular value vectors, while it fails to capture information projected by the other singular value vectors. Building on this observation, they proposed spectral filtering, a method that applies filters to selectively pass information encoded in specific bands of the representation. As a result, this approach allows one to discover information that is typically overlooked by the logit lens, such as how an LM avoids using outputs from irrelevant LM components during next-token predictions.

\subsection{Intervention-based Methods} \label{tech: intervention-methods}

\subsubsection{Basic Concepts}
Intervention-based methods investigate an LM's behaviors by directly altering the value of its intermediate representation during the forward pass (i.e., \emph{intervention}) and observing the resulting change of the model output.
In this approach, we view an LM as a computational or a causal graph~\cite{pearl2009causality}, where its transformer components {$\{C_0,..C_n\} \in \mathcal{C}$} 
are the nodes, and their connections represent the edges.  
By intervening on these nodes and edges, we can explore the causal relationships between different components and their influence on the model's output. In practice, the component $C_i$ can be defined at varying levels of granularity, such as a single neuron, an attention head, or an entire transformer layer. 

There are mainly two types of intervention-based methods: \emph{Noising}- and \emph{Denoising}-based intervention methods.

\paragraph{Noising-based Interventions} This line of methods involves removing (also referred to as \emph{noising}) the contribution of a specific component $C_i$ during the LM's forward pass and observing the resulting change in the model's output $Y$.
% \zyc{Notations need to be clarified: what is the index $i$ in $C_i$? In your earlier paragraph, you call a component $C$. What is $t$ and what is $y_t$?} 
The idea is that if the component $C_i$ is important to the original model output, there should be a significant change in the model output after the contribution of $C_i$ is removed from the model inference. In other words, noising-based methods aim to discover 
components that are \emph{necessary} for an LM to exhibit a certain behavior as removing the contributions of these components can break the LM's behavior~\cite{heimersheim2024use}. Noising-based interventions are also known as \emph{ablation}~\cite{chan2022causal} or \emph{knockout}~\cite{wang2022interpretability}.

\paragraph{Denoising-based Interventions} This line of methods typically involves two interventions to measure the importance of a component $C_i$ for the model output $Y$. {First, we perform \emph{noising} to all LM components $\mathcal{C}$ such that the model does not predict $Y$ anymore. Next, only $C_i$ is \emph{denoised} to see how much $C_i$ alone restores the likelihood (alternatively, logit; see Section~\ref{sec:intervention-tech-eval}) of the model predicting $y$.} If we observe a substantial increase in the likelihood of predicting $Y$ as the next token, it shows that $C_i$ plays an important role in the model's original prediction. In other words, denoising-based methods identify components that are \emph{sufficient} for restoring an LM's behavior to some degree. Denoising-based interventions have also been termed as \emph{causal tracing} or \emph{causal mediation analysis}~\cite{meng2022locating}. 
% \abu{are these method names typically capitalized? (similarly for Ablation and Knockout in the previous paragraph)}

\begin{table*}[t!]
    \centering
    \resizebox{\textwidth}{!}{%
    \begin{tabular}{>{\centering\arraybackslash}m{2cm}>{\centering\arraybackslash}m{3cm}>{\centering\arraybackslash}m{2.5cm}>{\centering\arraybackslash}m{3cm} m{5cm}}
    \toprule
    \textbf{Intervention Type} & \textbf{Corruption} \textbf{Technique} & \textbf{Clean Run} & \textbf{Corrupt Run} & \textbf{Patch Run}  \\
    \toprule

    \multirow[b]{4}{*}{\textit{Noising-based}\vspace{-0.7em}} & Zero Ablation \par \cite{olsson2022context} & Run with $X_{clean}$ & N/A & Perform clean run but replace activation of $C_i$ with zero-vector\\
    % \midrule
    \cline{2-5}\noalign{\vspace{0.3em}}
     & Random-noise Ablation \par \cite{rai2024investigation} & Run with $X_{clean}$ & N/A & Perform clean run but add random noise to activation of $C_i$ \\
    \cline{2-5}\noalign{\vspace{0.3em}}
     & Mean Ablation \par \cite{wang2022interpretability} & Run with $X_{clean}$ & Run with multiple $X_{corrupt}$ & Perform clean run but replace the activation of $C_i$ with the mean activation of multiple $X_{corrupt}$ \\
    \cline{2-5}\noalign{\vspace{0.3em}}
     & Resampling \par \cite{chan2022causal} & Run with $X_{clean}$ & Run with one $X_{corrupt}$ & Perform clean run but replace activation of $C_i$ with the activation from corrupt run \\

    \midrule
    \multirow[b]{2}{*}{\textit{Denoising-based}\vspace{-0.4em}} & 
    % Zero~\cite{} & Run with $X_{clean}$ & Replace the embedding of $X\_clean$ with zero vector & Perform corrupt run but replace activation of $C_i$ with its activation from the clean run\\
    % \cline{2-5}
    Random-noise \par \cite{meng2022locating} & Run with $X_{clean}$ & Add noise to the embedding of $X_{clean}$  & Perform corrupt run but replace the activation of $C_i$ with its activation from the clean run \\
    \cline{2-5}\noalign{\vspace{0.3em}}
     & Resampling \par \cite{hanna2024does} & Run with $X_{clean}$ & Run with $X_{corrupt}$ & Perform corrupt run but replace activation of $C_i$ with its activation from the clean run \\
     
    \bottomrule
    \end{tabular}
    }
    \caption{A summary of corruption techniques in intervention-based methods. 
    % \daking{[Note: Font-size should be the same as main text.]}
    % \zyc{Do we want to delete Zero Ablation from Denoising? It does not make sense to me and i can't recall works that used this approach. For Denoising, I'm not sure if we should call them "XX Ablation"} 
    }
    \label{tab: intervention-methods}
\end{table*}

\paragraph{Intervention Procedure}
The typical procedure for intervention-based methods, both noising and denoising, for measuring the contribution of component $C_i$ is outlined below. Conceptually, both types of intervention involve three runs.
\begin{itemize}
    \item \textbf{Clean run:} {Run the model $M$ with a prompt $X_{clean}$ (e.g., \emph{``The capital city of France is''}) that showcases the LM behavior $Y_{clean}$ (e.g., \emph{``Paris''}) and cache the activations of all LM components.}
    \item \textbf{Corrupt run:} {Run the model $M$ with a corrupted prompt $X_{corrupt}$ by selecting a prompt distinct from $X_{clean}$ (e.g., \emph{``The capital city of Italy is''}) which leads to model predicting $Y_{corrupt}$ (e.g., \emph{``Rome''}) instead of $Y_{clean}$. Cache the activations of all LM components from this corrupt run. In practice, one may implement the corrupt run in various ways. When patching $C_i$ with a zero vector, a corrupt prompt is not needed. One may not need a discrete prompt of $X_{corrupt}$ either and can instead apply continuous noise to the clean activation of $C_i$. We summarize the various corruption techniques in Table~\ref{tab: intervention-methods}.
    % There are other various corruption techniques listed in Table~\ref{tab: intervention-methods}.
    }
    % \item \textbf{Corrupt run:} Run the model $M$ with a corrupted prompt $X_{corrupt}$ which can be implemented either continuously by adding noise to the embeddings of $X_{clean}$, or discretely by selecting a contrastive prompt to $X_{clean}$ (e.g., \emph{``The capital city of Italy is''}) such that it leads to a decrease in the logit of the correct output (e.g., \emph{``Paris''}). Details about various other corruption techniques are discussed in Section~\ref{sub-sec: corruption-tech}). Cache the activations of all LM components from this corrupted run. \daking{Why did we say continuously and discretely?}
    \item \textbf{Patch run:} {This run can be performed in two ways based on whether you want to perform \emph{noising} or \emph{denoising} intervention: (a)\textbf{ Noising:} Run the model $M$ with clean prompt $X_{clean}$ but replace or \emph{patch} the activation of $C_i$ from the corrupt run. This patch removes $C_i$'s contribution towards the model predicting $Y_{clean}$ when it is provided with the input $X_{clean}$. 
    \mbox{(b)\textbf{ Denoising:}} Run the model $M$ with corrupted prompt $X_{corrupt}$ but replace or \emph{patch} the activation of $C_i$ from the \emph{clean run}. This patch includes only $C_i$'s contribution towards the model predicting $Y_{clean}$ when it is provided with the input $X_{clean}$. }
\end{itemize}

\subsubsection{Technique-specific Interpretation}\label{sec:intervention-tech-eval}
Once we perform the three runs, we measure the patching effect to evaluate the importance of a component $C_i$ for the LM output $Y$. For noising, we analyze the difference between the output of the clean and patch run to observe the resulting changes after removing the contribution of $C_i$ from the clean run. Alternatively, for denoising, we analyze the difference between the output of the corrupted and the patch run to observe how much patching $C_i$ recovers the clean run's output $Y_{clean}$. Specifically, we can measure the change in the LM output using the following metrics: 
% \abu{i think its fine to pick one of ``probability'' or ``logit'' in the below; it would be a lot easier to read} \daking{We have a discussion which metric to select, which I believe could be valuable for someone new to the field, as many prior works use all three metrics without clearly explaining their differences.}
\begin{itemize}
    \item \textbf{Probability or Logit:} Measure the change in the softmax probability or logit of $Y_{clean}$ before and after the patching.  
    
    \item \textbf{Probability or Logit Difference:} Measure the change in the difference between the probability or logit of $Y_{clean}$ and that of $Y_{corrupt}$ before and after the patching.

    \item \textbf{KL Divergence:} Measure the Kullback-Leibler (KL) divergence between the logit or probability distributions before and after the patching.
    % of the probability distribution of the output of $M$ and $M^{C_{patch}}$, i.e., $D_{KL}(Logit || Logit^{C_{patch}})$. 
    This metric compares the full output distributions rather than focusing solely on the change in the logit or probability of a single token.
\end{itemize}

\paragraph{Which Metric to Select?} 
In general, logit difference is recommended because it allows us to control things we do not want to measure (e.g. components that promote both $Y_{clean}$ and $Y_{corrupt}$), which is not possible using the absolute probability or logit. In addition, probability as a metric may fail to detect \emph{negative model components} that suppress the correct output, as empirically observed by \citet{zhang2023towards}. While KL divergence does not focus on specific token logits or probabilities, it is also a reasonable metric for assessing the effect of patching. For a more detailed comparison of the reliability of these metrics across different scenarios, we refer readers to \citet{heimersheim2024use, zhang2023towards}.

\subsubsection{Technical Advancements}\label{subsec:intervention-tech-advancements}
Intervention-based techniques have undergone several refinements to ensure their correct usage, including various adaptations to address diverse investigative objectives. Specifically, we summarize these technical advancements as follows: (1) Reliability of corruption: ensuring that the corrupt run does not take the model out of distribution (OOD); (2) Intervention for localization: adapting the techniques to identify important components associated with specific LM behavior; (3) Scaling up and automating localization techniques: automating the intervention process to alleviate the need for human effort; (4) Intervention for validating hypotheses: applying intervention to validate a hypothesized interpretation of the LM; and (5) Integrating intervention with vocabulary projection for activation decoding: devising intervention methods for more expressive vocabulary projections of activations. We present representative approaches discussed in the technical advancements in Table~\ref{tab:intervene-tech-summary}.
% \daking{We summarize the technical advancements of intervention techniques in Table~\ref{tab:intervene-tech-summary}.}

\begin{table*}[!t]
\centering
\resizebox{\textwidth}{!}{
\begin{tabular}{
    >{\arraybackslash}p{3.0cm}
    >{\arraybackslash}p{8.0cm}
    >{\arraybackslash}p{5.0cm}
}
\toprule
\textbf{Approach} & \textbf{Basic Concepts} & \textbf{Technical Advancements} \\
\midrule
% Zero-ablation~\cite{olsson2022context} & Replaces activation with a \emph{zero vector} to remove or corrupt its contribution from the model’s computation. & N/A. \\
% \midrule
% Random-ablation~\cite{rai2024investigation} & Replaces activation with a \emph{random vector} to remove or corrupt its contribution from the model’s computation. & N/A. \\
% \midrule
Mean-ablation~\cite{rai2024investigation} & Replace activation with its average across counterfactual inputs, preserving baseline task information while removing or corrupting input-specific contributions. & Improve \emph{reliability} of corruption by mitigating the OOD concern due to intervention. \\
\midrule
Resampling-ablation~\cite{rai2024investigation} & Replace activation with its activation of a randomly sampled counterfactual input. & Improve \emph{reliability} of corruption by mitigating the OOD concern due to intervention. \\
\midrule\midrule
Activation Patching~\cite{meng2022locating} & Investigate the importance of a node in the circuit by performing noising or denoising intervention on the node activation. & Intervention for \emph{localizing} an important node in a circuit.\\
\midrule
Path Patching~\cite{wang2022interpretability} & Investigate the importance of edges between two components ($C_1$ and $C_2$) by applying activation patching only along the computational path from $C_1$ to $C_2$. & Intervention for \emph{localizing} important edges in a circuit.\\
\midrule
Distributed Interchange Interventions (DII)~\cite{geiger2024finding} & Target interventions to specific subspaces of a representation rather than replacing the entire activation. & Intervention at a more \emph{granular} level than activation patching. \\
\midrule
Distributed Alignment Search (DAS)~\cite{geiger2024finding} & A supervised method for discovering lower-dimensional subspaces associated with specific causal variables.  & Intervention at a more \emph{granular} level than activation patching. \\
\midrule\midrule
ACDC~\cite{conmy2023towards} & Automate circuit discovery by iteratively removing edges in the computational graph and discarding those with minimal effect on a target metric. & \emph{Automate} intervention for localization by reducing manual effort.
\\
\midrule
EAP~\cite{syed2023attribution} \& EAP-IG~\cite{hanna2024have} & Approximate edge patching with gradient-based attribution to efficiently locate important edges in a circuit, trading some faithfulness for much greater scalability. & \emph{Automate} and \emph{scale up} intervention for localization by reducing the computational requirement.
% Reduces the computational requirement for the circuit discovery process.
\\
\midrule\midrule
Causal Scrubbing~\cite{chan2022causal} & Rigorously test a hypothesis about how a model works by systematically intervening in its internal activations and then observing if the outputs (before and after intervention) match what the hypothesis would predict. & Intervention for \emph{validating hypotheses}. \\
\midrule\midrule
Fixed/Learned Prompt Causal Intervention~\cite{pal2023future} & Decode features from intermediate activations by patching them into a separate LM run on a fixed or learned prompt, then interpreting the resulting generation as the features encoded in the original activation. The initial goal of the approach is to decode \emph{future} tokens from the end of the input. &  \emph{Integrate intervention with vocabulary projection} for activation decoding. \\
\midrule
Patchscope~\cite{ghandeharioun2024patchscope} & Decode features from intermediate activations by patching them from a \emph{source LM and prompt} into a \emph{target LM} run on an \emph{inspection (target) prompt}, then interpreting the resulting generation as the features encoded in the original activation.  &  \emph{Integrate intervention with vocabulary projection} for
activation decoding. \\
\bottomrule
\end{tabular}
}
\caption{Representative approaches in the technical advancements of intervention techniques.}
\label{tab:intervene-tech-summary}
\end{table*}

\paragraph{Reliability of Corruption}
The corruption of activations during the corrupt run can push the model out of distribution (OOD) as the modified activations may not resemble anything the model encountered during training. As a result, any decline in model performance may be due to the off-distribution behavior of the model, rather than the removal of $C_i$'s contribution to the behavior of interest~\cite{zhang2023towards}. Specifically, Zero and Random-noise ablations can be unreliable because both replacement values are arbitrary choices and the model may not have encountered them during training. Mean ablation attempts to address this issue by replacing the activation of $C_i$ with the average activation of $C_i$ computed from multiple samples drawn from the same data distribution. This approach partially mitigates the OOD concern as it obtains the replacement value from in-distribution activations, making it more reliable than Zero and Random-noise ablation. 
However, the mean value could still push the LM off distribution if the activation distribution is non-linear~\cite{chan2022causal, geiger2021causal}. For example, if the activation distribution of $C_i$ consists of points along the circumference of a circle, the mean of these points would be at the center, rather than at another point along the circumference~\cite{chan2022causal}. Resampling ablation addresses this issue by simply replacing the activation of $C_i$ with the activation from a counterfactual example sampled from the same distribution~\cite{chan2022causal}. 

% \paragraph{Improving scope of localization} 
% \paragraph{\zy{Scope of Interventions}}
\paragraph{Interventions for Localization}
% The investigation of LM behavior typically begins with localization i.e. identifying the key components within $M$ that contribute to the behavior under study. 
\begin{figure*}[t!]
    \centering
    \includegraphics[width=0.9\textwidth]{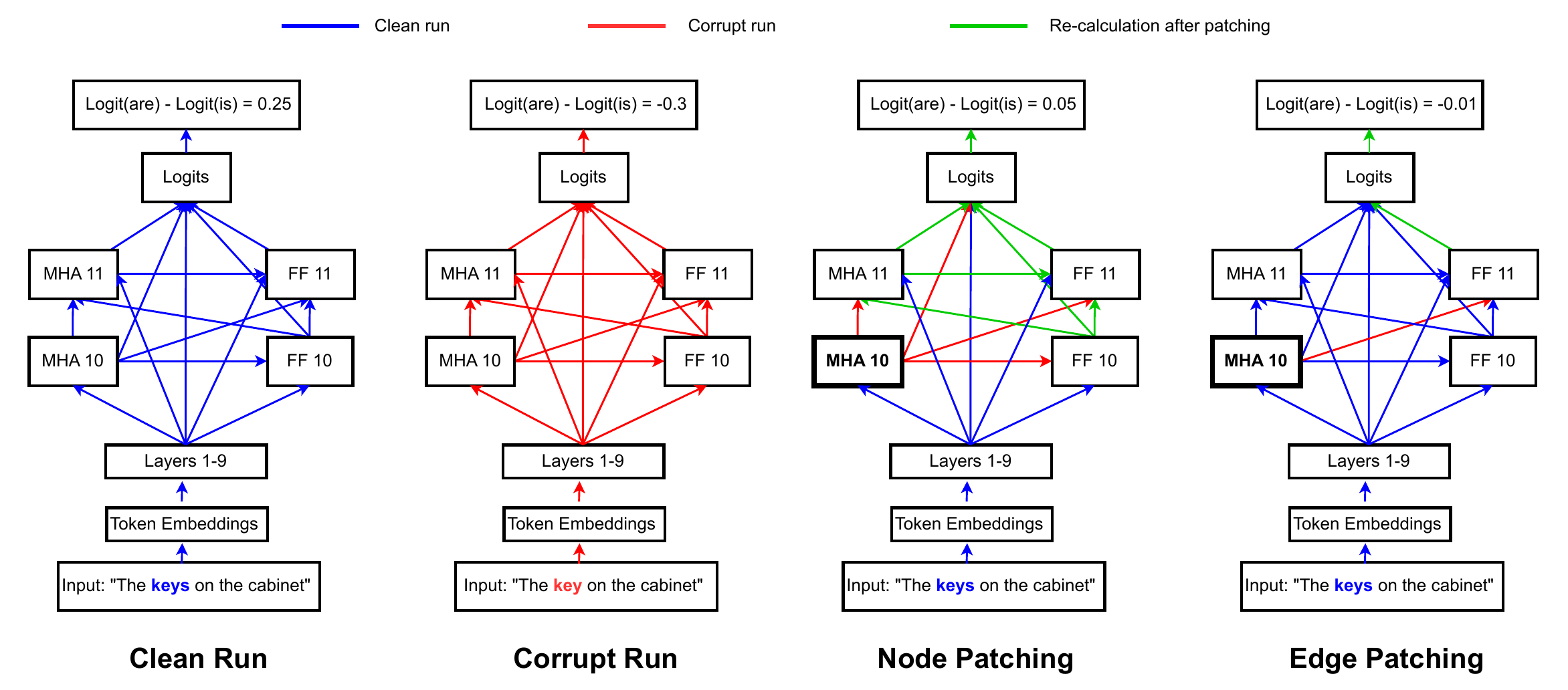}
    \caption{An example of \emph{noising-based} intervention on GPT-2 at the \emph{node} and the \emph{edge} levels. In the example, a corrupt prompt, with the singular \textit{``key''} replacing the plural \textit{``keys''} in the clean prompt, was introduced to discover the circuit in GPT-2 for associating the verb with different forms of the noun. 
    (1) Clean Run: The computational graph of GPT-2 when run on clean input. (2) Corrupt Run: The computational graph of GPT-2 when run on corrupt input. (3) Node Patching: The computational graph of GPT-2 when run on clean input, with the output activation of MHA10 replaced by its activation from the corrupt run (i.e., patching the node of ``MHA10''). (4) Edge Patching: The computational graph of GPT-2 when run on clean input, with activations along the edge of (MHA10, FF11) replaced by activations from the corrupt run (i.e., patching the edge of ``(MHA10, FF11)''). Colored connections indicate activations calculated from the clean run (Blue), the corrupt run (Red), and the re-calculation after patching (Green), respectively.}
    % \zyc{update the caption}\abu{it would be clearer to split ``Token Embeddings'' and ``Layers 1-9'' into two separate nodes}}
    % \abu{minor: i think `key' should be `keys' in the corrupt run example}
    \label{fig:patching}
\end{figure*}

As we will introduce in our roadmap (Section~\ref{sec: roadmap}), intervention is commonly used to localize important components (i.e., \emph{nodes} of circuits) and computational pathways (i.e., \emph{edges} of circuits) when one tries to discover circuits in an LM. Depending on the goal, the scope of intervention varies. \textbf{(1) Patching nodes vs. edges:} 
% \abu{we should try to be consistent with capitalization of these numbered titles; for example, the numbered titles here are capitalized in 4.1.3 or 4.2.3}
\textit{Activation patching} is a widely used technique employed to localize key components~\cite{meng2022locating}. This method evaluates the importance of a component \( C_i \) by patching the activation of \( C_i \) from the \emph{clean run} into the \emph{corrupt run}, one component at a time, to observe its impact on the model's output, as illustrated in Figure~\ref{fig:patching}. The higher impact implies $C_i$ is an important component of the behavior. \emph{Path Patching} is an extension of activation patching to localize important {edges} between components \cite{wang2022interpretability, goldowsky2023localizing}. For instance, to assess whether the connection between two components, $C_1$ and $C_2$, is significant, path patching applies activation patching to the output of $C_1$ \emph{but, notably, only along paths serving as input to $C_2$}. In other words, when performing path patching, we allow for the effect of a patched activation only along the computational path we want to investigate, while any other directed nodes will not receive the patched activation despite the connection. Similar to patching nodes, if a change in the LM's behavior is observed, the connection between the two components is considered important. \textbf{(2) Patching at various granularity levels:} As discussed in Section~\ref{subsec: features}, it is hypothesized that LMs encode features within a linear subspace of the representation space, potentially represented in a non-standard basis, as indicated by the discovery of polysemantic neurons (see Section~\ref{para: polysemantics} for further discussion). 
{However, activation patching focuses on replacing the entire hidden representation and is typically analyzed under a \emph{localist} assumption that each high-level causal variable corresponds to a disjoint set of neurons~\citep{geiger2024finding, geiger2025causal}.  To overcome this limitation, \textit{Distributed Interchange Interventions (DII)} proposes to perform intervention in rotated subspaces of the representation rather than in the raw neuron basis~\citep{geiger2024finding}. Concretely, DII performs the following operations: (i) it first applies a change-of-basis rotation transformation to rotate the model’s representation into a new basis; (ii) within this rotated basis, specific subspaces corresponding to high-level causal variables are identified, and interventions are performed on those subspaces; (iii) after the intervention, the representation is transformed back into the original neural basis. By operating at the subspace level, DII reveals interpretable distributed structure and enables more fine-grained interventions of model activations. In addition to DII, \citet{geiger2024finding} propose another technique \textit{Distributed Alignment Search (DAS)}, a supervised method for automatically discovering the rotation matrix and the $k$-dimensional subspaces within the rotated representation that best align with the high-level causal variables.} DAS involves learning an orthogonal transformation matrix that maps neural representations onto $k$-dimensional subspaces aligned with the causal variable. The transformation is trained on clean-counterfactual pairs via interchange interventions, with the objective of maximizing interchange intervention accuracy under the $k$-dimensional constraint.

% \daking{However, activation patching focuses on replacing the entire representation and relies on a \emph{localist} assumption, namely that high-level causal variables are encoded within disjoint groups of neurons~\citep{geiger2024finding, geiger2025causal}. To this end, \textit{Distributed Interchange Interventions} (DII) were proposed, which do not limits itself in localist assumption and perform interventions in rotated subspaces of the representation rather than on the full representation, enabling more fine-grained tests of how high-level causal variables are encoded within model activations~\citep{geiger2024finding}. Building on this idea, the authors also proposed \textit{Distributed Alignment Search} (DAS), a supervised method for discovering $k$-dimensional subspaces associated with a specific causal variable. DAS involves learning an orthogonal transformation matrix that maps neural representations onto $k$-dimensional subspaces aligned with the causal variable. The transformation is trained on clean-counterfactual pairs via interchange interventions, with the objective of maximizing interchange intervention accuracy under the $k$-dimensional constraint.}

\paragraph{Scaling Up and Automating Interventions for Localization}
Performing interventions to localize important components or connections requires humans in the loop, as it involves iteratively patching each component or connection within an LM to assess its significance, which can be time-consuming.
To automate this process, \citet{conmy2023towards} introduced \emph{ACDC (Automatic Circuit DisCovery)}, which iteratively knocks out edges in the computational graph and removes any edge whose effect on the target metric is less than a specified threshold. However, ACDC is not scalable to large LMs as it requires independent inferences for every iteration. To address the issue, \citet{syed2023attribution} proposed \textit{Edge Attribution Patching (EAP)} for locating important edges, where it employs \emph{Attribution Patching}~\cite{nandaattribution} to approximate activation patching for locating important edges, which requires only two forward passes and one backward pass for measuring all model components. In addition, \citet{hanna2024have} recently augmented this approach with Integrated Gradient~\cite{sundararajan2017axiomatic} to address the concern of zero gradients, leading to the approach of \textit{EAP-IG}.

% To address the issue, \citet{nandaattribution} \abu{this was published two years prior; so how was it to address the issue? is there an earlier reference for ACDC?} proposed \emph{Attribution Patching} to approximate activation patching for locating important components, which requires only two forward passes and one backward pass for measuring all model components. \citet{syed2023attribution} further extended it to \textit{Edge Attribution Patching (EAP)} for locating important edges, which outperforms ACDC in circuit discovery. \citet{hanna2024have} recently augmented this approach with Integrated Gradient~\cite{sundararajan2017axiomatic} to address the concern of zero gradients, leading to the approach of \textit{EAP-IG}.

% \paragraph{Hypothesis Formulation and Validation}
\paragraph{{Interventions for Validating Hypotheses}}
% \paragraph{\dakingrev{Interventions for Interpreting LM Components}}
% Once the important components for a given LM behavior have been identified through localization, the next step involves interpreting the function of each LM component and how they interact with each other to implement the LM behavior ultimately. 
% \zyc{Can you revert the first few sentences to my revisions for "Interventions for Validating Hypotheses"?}
% Once the key components for the LM behavior have been localized, the next step is to investigate the role each component plays in this behavior. This investigation process involves two main steps, as shown in Figure~\ref{fig: roadmap}: (1) generating hypotheses about the roles of individual components, and (2) validating these hypotheses. To support this process, various techniques have been proposed for hypothesis formulation and validation of hypotheses. 
Intervention-based methods are widely used to causally validate hypotheses about how specific LM components contribute to its behavior.
For instance, \emph{Causal Scrubbing}~\cite{chan2022causal} is a popular technique for formalizing and validating a hypothesis about the function of an LM component for the given behaviors. Specifically, causal scrubbing involves formalizing a hypothesis $(G, I, c)$, where $G$ is the model's original computational graph, $I$ is an interpretation graph that reflects the hypothesized roles of the model's components, and $c$ is a function that maps between nodes of $I$ to $G$.
For example, when an LM is tasked to perform a two-digit addition, the interpretation graph $I$ could specify the hypothesized computational pathway, where only certain transformer components are used for implementing this addition while others do not contribute to this task.
% \abu{i was confused about this before: isn't the definition of $G$ itself a hypothesis? or are you considering $G$ to be ``latent''? i think a concrete example (with a concrete definition of each of $G$, $I$, and $c$) would help to better convey/refine this concept; i think the example below is too abstract} \daking{$(G, I, c)$ all together is the hypothesis while $G$ is the model itself represented or viewed as a computational graph and $I$ is the sparse circuit hypothesized to be responsible for certain behavior or role. I have tried to present a concrete example of addition}. 
To validate that $I$ is a faithful interpretation of $G$ for a given behavior, \emph{resampling ablations} of component activations in $G$ are performed. 
% For example, if the hypothesis states that some model components are unimportant for a given LM behavior (e.g.,$1+2 \xrightarrow{} 3$), then replacing activations of these model components with resampled example activations (e.g., $2+5$) should not change the LM behavior (e.g., $3$).
In our example, it means that replacing activations of model components that do not contribute to the addition task should not change the LM's prediction behavior.
This provides a more rigorous way to check our intuitions about how models work, compared to just looking at the model or manually changing its parts. Similarly, \textit{Interchange Intervention}~\cite{geiger2021causal} is another hypothesis verification technique that similarly formulates the hypothesis by representing the model $M$ as a high-level conceptual model and performing resampling ablation on the low-level (i.e., the original) model to verify the hypothesis. 

% \paragraph{\dakingrev{Interventions for Interpreting LM Activations}}
\paragraph{{Integrating Intervention with Vocabulary Projection for Activation Decoding}}
{The intervention technique has also been adapted by \citet{pal2023future, ghandeharioun2024patchscope} to decode features from intermediate activations, which can be viewed as augmented vocabulary projection (Section~\ref{tech: vocab-proj-methods}) with more powerful expressivity. For instance, to decode features from the end-position activation $h_n^l$, the \emph{fixed prompt causal intervention}~\cite{pal2023future} runs another LM inference on a fixed generic prompt (e.g., ``Tell me something about'') with its activation at the same location being patched by the target one, and then let the LM continue the generation. This generation can then be viewed as the projected features encoded in the activation $h_n^l$. When this fixed prompt is replaced by one learned for effective information elicitation, it results in the variant of \emph{learned prompt causal intervention} (or \emph{future lens}, as the initial goal of \citealt{pal2023future} was to probe the future tokens from the end-position activation). 
% \emph{fixed prompt causal intervention}~\cite{pal2023future} involves decoding a LM activations generated from an original input (e.g., \emph{``Madison Square Garden is located in''}) by replacing those activations with ones obtained from decoding an unrelated prompt (e.g., ``Tell me Something about''). If this replacement causes the model to generate tokens (e.g., \emph{``New York City''}) related to the original input, it suggests that the intervened activations encoded the newly generated token features. 
% Concurrently, PatchScope~\cite{ghandeharioun2024patchscope} provides a unifying framework that combines vocabulary projection and activation patching~\cite{wang2022interpretability} to decode features of an intermediate activation. Specifically, PatchScope involves patching activations from a source prompt to an inspection prompt carefully designed to elicit the interested feature information. 
Concurrently, \emph{Patchscope}~\cite{ghandeharioun2024patchscope} provides a unifying framework that similarly combines vocabulary projection with intervention methods but is more generalized than the approaches of \citet{pal2023future}. Specifically, Patchscope defines a source prompt and a source LM, which produces the activation to be interpreted, as well as a target prompt and a target LM, which facilitates the activation interpretation. Patchscope works by patching the activation derived from the source to the corresponding activation in the target. By carefully designing the target prompt (also termed as ``inspection prompt''), Patchscope allows one to {decode} the interested feature information. 
% \abu{unclear what ``eliciting information'' means; ``elicit'' is used to describe something causing a response/behavior}.
% For instance, to check if an LM's activation obtained from the end position of the source prompt \emph{``Eiffel Tower is located in''} encodes the feature of \emph{``Paris''}, an inspection prompt (or $X_{corrupt}$) can be \emph{``The largest city in X is''}.
% % , as shown in Figure~\ref{fig:patchscope}. 
% If the patched LM run on the inspection prompt generates \emph{``Paris''} as the next token, we can infer that the activation contains information related to the token \emph{``Paris''}.
For instance, to check if an LM's activation obtained from the end position of the source prompt \emph{``Amazon's former CEO''} encodes the feature of \emph{``Jeff Bezos''}, an inspection prompt (or $X_{corrupt}$) can be \emph{``cat $\rightarrow$ cat; 135 $\rightarrow$ 135; hello $\rightarrow$ hello; ?''}, which comprises few-shot demonstrations encouraging the model to explain its internal representation, with the final \emph{``?''} served as a placeholder. If the LM with the activation at the position of \emph{``?''} patched by the source activation can generate \emph{``Jeff Bezos''} as the continuation, we can infer that the activation contains information about this person's name.

\subsection{Sparse Autoencoder (SAE)} \label{sec: sae}

\begin{figure}[t!]
    \centering
    \includegraphics[width=0.5\linewidth]{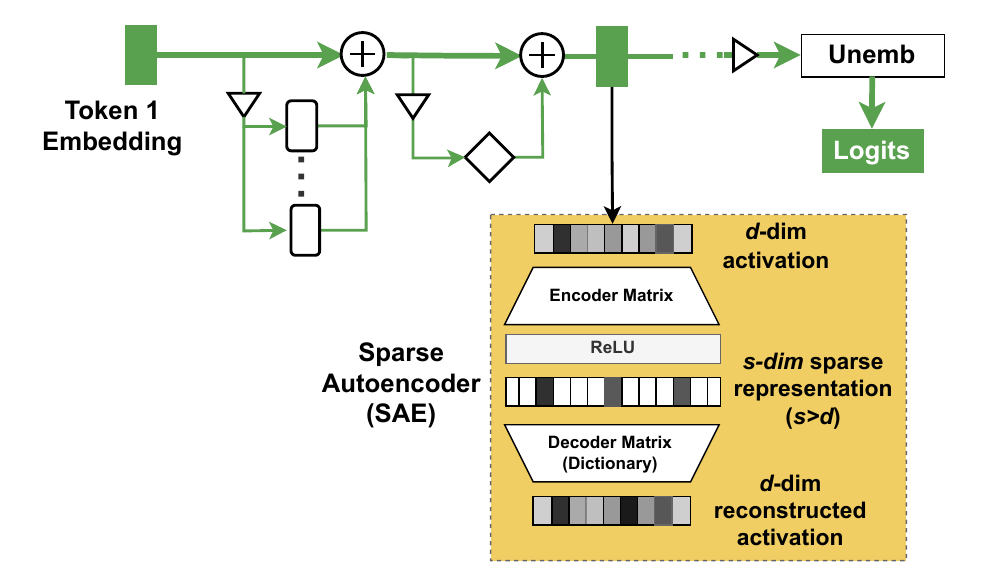}
    \caption{Sparse Autoencoder (SAE) applied to activation on RS.}
    \vspace{-2mm}
    \label{fig:sae}
\end{figure}

\subsubsection{Basic Concepts}
Sparse Autoencoders (SAEs) are employed in feature discovery to tackle the problem of \emph{superposition}, where LM activations encode more features than their dimensions, leading to \emph{polysemantic} neurons that activate for multiple unrelated concepts. SAEs tackle this issue by mapping a \(d\)-dimensional activation into an \(s\)-dimensional sparse representation (\(s > d\)), where this new \(s\)-dimensional representation consists of neurons that are more \emph{monosemantic}, each associated with only a single feature, making them more interpretable~\cite{bricken2023monosemanticity, gao2024scaling, rajamanoharan2024jumping}. In other words, the primary goal of SAEs is to transform the less interpretable LM activation into a more interpretable sparse activation.

% Sparse Autoencoders (SAEs) are employed in feature discovery to address the issue of \emph{polysemanticity}, where individual neurons represent multiple unrelated features, complicating the interpretation of how activations encode features~\cite{sharkey2023taking, bricken2023monosemanticity, cunningham2023sparse, templeton2024scaling, rajamanoharan2024improving, gao2024scaling, rajamanoharan2024jumping}.

% \zyc{I revised notations in the following paragraphs to be more consistent with the background section: use $h$ rather than $x$ to denote an activation. $X$ is a single input and $\mathcal{X}$ is a set of inputs. Use $g$ (rather than $\sigma$, which typically refers to the sigmoid function) to denote a non-linear function.}

\paragraph{SAE Architecture} An SAE (Figure~\ref{fig:sae}) consists of an \emph{encoder} that maps the \(d\)-dimensional input activation into an \(s\)-dimensional sparse activation and a \emph{decoder} which reconstructs the original input activation from the sparse representation, defined as follows:
\begin{equation}
\begin{split}
    f(h) = g(W_{enc}h + b_{enc})  \\
    \hat{h} = W_{dec}f(h) + b_{dec}
\end{split}
\label{eq: sae}
\end{equation}
Here, the encoder consisting of weights $W_{enc} \in \mathbb{R}^{d \times s}$,
% \zyc{should be $s \times d$?} \daking{$d$ is the dimension of activation. So, I think ${d \times s}$ is correct}\zyc{but $h$ has a dim of $\mathbb{R}^d$; when applying a matrix to multiply with $h$, the second dim should match with $h$'s. Eventually SAE will have encoder weights $W_{enc} \in \mathbb{R}^{s \times d}$ and decoder weights $W_{dec} \in \mathbb{R}^{s \times d}$ with reversed shapes.} 
a bias term $b_{enc} \in \mathbb{R}^s$, and a non-linear activation function $g$ (e.g., ReLU~\cite{agarap2018deep}), maps the input activation $h \in \mathbb{R}^d$ to sparse activation $f(h) \in \mathbb{R}^{s}$. The size of the sparse representation $s$ is a hyperparameter, decided by the researcher. Subsequently, the decoder consisting of weights $W_{dec} \in \mathbb{R}^{s \times d}$ and a bias term $b_{dec} \in \mathbb{R}^d$, takes the sparse activation $f(h)$ from the encoder and reconstructs the input activation $\hat{h}$. The decoder is also commonly referred to as the \emph{dictionary}, as it comprises a set of learned feature vectors encoded in its weight matrix after training. Consequently, the size of the sparse representation $s$ is referred to as the dictionary size.

\paragraph{Loss Function}
An SAE is trained in an unsupervised approach using a dataset $\mathcal{X}$ with a loss function defined as:
\begin{equation}
% \begin{split}
    \mathcal{L}(h, \hat{h}) = \frac{1}{|\mathcal{X}|} \sum_{X \in \mathcal{X}}  \Big( \underbrace{\| h(X) - \hat{h}(X) \|_2^2}_{\mathcal{L}_{reconstruct}}  +\underbrace{\lambda \|f(h(X))\|_1}_{\mathcal{L}_{sparsity}} \Big) 
% \end{split}
\label{eq: sae-loss}
\end{equation}
% \zyc{align the two lines for beautification; note that the prev $\mathcal{L}_{sparsity}$ had a redundant summation.}
where $\mathcal{L}_{reconstruct}$ ensures the faithfulness or fidelity of the reconstructed activation to the input, while the sparsity penalty $\mathcal{L}_{sparsity}$ restricts the encoder to activate only a small number of neurons for a given $h$, encouraging the SAE to learn a sparse interpretable representation for $h$ during training. Specifically, without a sparsity penalty, the encoder could simply memorize the input activations since \( f(h) \) has more dimensions than \( h \). However, introducing a sparsity penalty forces the encoder to reconstruct \( h \) while activating only a small subset of neurons. This constraint encourages the SAE to capture the most important features of \( h \) in \( f(h) \), as doing so minimizes the overall loss. As a result, the SAE learns to disentangle and represent the most salient features of \( h \) more effectively. In addition, these two objectives (fidelity and sparsity) are balanced by the \emph{$L_1$ coefficient} $\lambda$.

\paragraph{Decoding Positions}
SAEs can be trained on activations from any LM component. For example, \citet{templeton2024scaling, bricken2023monosemanticity, templeton2024scaling, sharkey2022taking} trained SAEs on intermediate activations from the residual stream, \citet{kissane2024interpreting, krzyzanowski2024we} trained SAEs on attention outputs, while \citet{braun2024identifying} trained SAEs on the outputs of FF sub-layers. 
% \abu{i wonder if the sections on LM positions should be moved earlier, as it is a common question that a new practitioner may have immediately after learning about SAEs (i.e., how can they be used?). in addition, this isn't really a technical advancement}

\subsubsection{Technique-specific Interpretation}
The fidelity of the SAE is measured using the reconstruction loss, where a low reconstruction loss implies high fidelity. However, evaluating the interpretability or quality of the features learned by an SAE remains an open research challenge. While increased sparsity can enhance interpretability, it does not necessarily guarantee highly interpretable features. Existing work has focused on analyzing the activation patterns of features with particular emphasis paid to sequences that a feature activates most strongly \cite{bills2023language, bricken2023monosemanticity, cunningham2023sparse, templeton2024scaling, rajamanoharan2024improving, gao2024scaling, rajamanoharan2024jumping}. The rating of a feature’s interpretability is usually either done by human raters~\cite{bricken2023monosemanticity} or automatically by prompting a language model to do the evaluation~\cite{bills2023language}.

\begin{table*}[!t]
\centering\resizebox{0.9\textwidth}{!}{
\begin{tabular}{
    >{\arraybackslash}p{3.0cm}
    >{\arraybackslash}p{8.0cm}
    >{\arraybackslash}p{4.0cm}
}
\toprule
\textbf{Approach} & \textbf{Basic Concepts} & \textbf{Technical Advancements} \\
\midrule
Standard SAE~\cite{bricken2023monosemanticity} & Train a standard SAE with ReLU activation and $L1$ sparsity that maps model activations into higher-dim sparse activations with more monosemantic units, which helps open-ended feature discovery. & N/A \\
\midrule\midrule
TopK SAE~\cite{gao2024scaling} & Enforce sparsity without using L1 penalties by retaining only the top-K activations per input, reducing feature suppression caused by L1 regularization. & Improve \emph{sparsity reconstruction trade-off}.  \\
\midrule
JumpReLU SAE~\cite{rajamanoharan2024jumping} & Replace ReLU with the JumpReLU activation function, which sets an activation threshold to remove false-positive sparse activation.  & Improve \emph{sparsity reconstruction trade-off}. \\
\midrule
Gated SAE~\cite{rajamanoharan2024improving} & Introduce a modified encoder architecture with two separate pathways: a gating path, which determines active features and is subject to a sparsity penalty, and a magnitude path, which estimates the strength of each active feature and is not penalized for sparsity & Improve \emph{sparsity reconstruction trade-off}. \\
\midrule
Matryoshka SAE~\citep{bussmann2025learning} & Trains multiple nested sub-SAEs of increasing dictionary sizes simultaneously to learn a common feature space that consists of both general and specific features. & Improve \emph{sparsity reconstruction trade-off}. \\
\midrule\midrule
End-to-End SAE~\cite{braun2024identifying} & Minimize KL divergence between the original model’s output and that of the model with SAE-generated activations, rather than reconstruction loss. & Improve \emph{discovery of functionally important features}. \\
\bottomrule
\end{tabular}
}
\caption{Representative approaches in the technical advancements of sparse autoencoders (SAE).}
\label{tab:sae-tech-summary}
\end{table*}

\subsubsection{Technical Advancements}
%The majority of research explorations about SAEs has been focused on balancing the trade-off between fidelity and sparsity of the autoencoder training.  In addition, existing work has applied SAEs to learn sparse representations for activations from various decoding positions of the LM. 
{SAEs have emerged as a dominant approach for conducting open-ended feature studies (Section~\ref{sub-sec: open-ended-workflow}) and have garnered significant attention within the MI community. The technical advancements in SAEs can be broadly categorized into two categories:
(1) Balancing the trade-off between reconstruction and sparsity loss; (2) Discovering features that are functionally useful for the LM to implement downstream applications. We summarize representative approaches discussed in the technical advancements of SAEs in in Table~\ref{tab:sae-tech-summary}.
% \daking{We summarize the technical advancements of SAE in Table~\ref{tab:sae-tech-summary}.}
% functionally useful features: meaningful features useful for LM to implement downstream applications 
% (3) Decoding positions, where prior research has explored the application of SAEs to various positions within an LM.
% : to determine which LM activations can be effectively analyzed using SAEs.}

\paragraph{Balancing the Trade-off between Reconstruction and Sparsity Loss}
{The loss function of SAEs consists of two objectives: \emph{reconstruction loss}, i.e., the sparse representation should accurately preserve information from input activation, and \emph{sparsity loss}, i.e., only a small number of elements in the sparse representation should be active for any given input activation. These two objectives can conflict, as greater sparsity can reduce reconstruction loss~\cite{gurnee2024sae}. 
For instance, ~\citet{wright2024addressing} noted that $L_1$ penalty in sparsity loss can lead to \textit{feature suppression}, a systematic underestimation of feature activation magnitudes, particularly those with weak activations but high frequency, negatively impacting reconstruction loss. To address this, a line of research has focused on improving SAE architectures and training methods to better balance the trade-off between sparsity and reconstruction loss~\cite{gao2024scaling, rajamanoharan2024jumping, erichson2019jumprelu}.
% \abu{what is the Pareto frontier? it is undefined}. 
For instance, \citet{gao2024scaling} proposed \textit{TopK SAE} that enforces sparsity by selecting only the K most active features and setting the rest to zero, where the sparse representation is obtained by $f(h) = \texttt{TopK}(g(W_{enc}h + b_{enc}))$. This approach eliminates the need for an $L_1$ penalty to achieve sparsity, mitigating the issue of feature suppression. Similarly, \citet{rajamanoharan2024jumping} introduced \textit{JumpReLU SAEs}, which leverage the JumpReLU function \cite{erichson2019jumprelu} in place of the ReLU function. The JumpReLU function sets activations below a positive threshold to zero, effectively removing false positives and increasing sparsity, while leaving activations above the threshold intact, thus preventing feature suppression and improving fidelity. \textit{Gated SAEs}~\cite{rajamanoharan2024improving} address the issue of feature suppression by using a gated ReLU encoder that decouples the detection of active features from the estimation of their magnitudes, and applying the $L_1$ penalty only to the feature detection. This decoupling allows the Gated SAE to achieve sparsity without underestimating feature magnitudes. {In addition to feature suppression, the other known issues include (1) \textit{feature splitting}~\citep{bricken2023monosemanticity}, where a general features (e.g., punctuation marks) fragments into more specific features (comma, period, question mark); (2) \textit{feature absorption}~\citep{chanin2024absorption}, where parent feature only partially splits leading to general feature with holes (e.g., a feature that activates on all tokens starting with an ``E'', except if the word is ``Elephant''); and (3) \textit{feature composition}~\citep{anders2024sparse}, where distinct features are entangled into a single feature (e.g., ``red triangle'' instead of ``red'' and ``triangle'' separately). To address these issues, \citet{bussmann2025learning} proposed \textit{Matryoshka SAE}, a novel hierarchical approach to SAE training that mirrors the hierarchical structure of real-world features. Specifically, Matryoshka SAE trains multiple nested sub-SAEs of increasing dictionary sizes simultaneously to learn a common feature space that consists of both
general and specific features. Each sub-SAE is optimized to reconstruct the input using only a subset of the total latents. This design prevents later, more specialized features from absorbing the roles of earlier, more general ones, thereby regularizing the SAE to capture features across multiple levels of abstraction.}} 

\paragraph{Discovering Functionally Important Features}
{The primary goal of training SAEs is to discover functionally-important features used by LMs to implement their behavior. Specifically, a feature is considered functionally important if it facilitates the explanation of
% is \dakingrev{helpful?} for explaining 
the model behavior on the training distribution
% \abu{this definition is somewhat circular (using ``important'' to define ``important''), and non-precise}. 
However, standard SAEs trained to minimize the reconstruction and sparsity loss may not strongly correlate with the goal of discovering functionally important features~\cite{braun2024identifying, makelov2024towards}. To address this, \citet{braun2024identifying} proposed \textit{end-to-end (e2e) SAEs}, which are trained to minimize the KL divergence between the output distributions of the original model and the model with SAE-generated activations inserted, instead of minimizing the reconstruction loss. This approach ensures that e2e SAEs are optimized to identify features that influence the model's predictions, rather than focusing on features that only accurately reconstruct activations.}

% \subsubsection{Other Variants SAEs}
% tied and untied. end-to-end SAEs

% SAEs (Figure~\ref{fig:sae}) serve as an unsupervised technique for discovering features from activations, especially those that demonstrate \emph{superposition},
% a phenomenon in LMs where their $d$-dimensional representation encodes more than $d$ features \cite{elhage2022superposition, sharkey2023taking, cunningham2023sparse, bricken2023monosemanticity, yun2021transformer}.   
% In contrast to dimensionality reduction techniques (e.g., Principal Component Analysis), SAEs seek to embed the activation vectors into a much higher-dimensional space, but with strong sparsity.
% Specifically, an encoder maps the $d$-dimensional input into an $s$-dimensional vector ($s > d$), which the decoder then maps back to the $d$-dimension. The encoder and decoder are jointly trained for input reconstruction and sparsity of the $s$-dimensional representation. This $s$-dimensional representation, owing to its sparsity, makes the discovery of independent (or \emph{monosemantic}) features more easily.

\subsection{Others}

In addition to the three methods we present above, i.e., vocabulary projection, intervention, and SAEs, two other techniques are also widely used to aid the mechanistic interpretation of LMs. Both techniques have a long-standing history in the broad research topic of interpretable AI and machine learning. Below, we provide a brief introduction to them and their applications to interpreting LMs.

\subsubsection{Probing} \label{subsec: probing-tech}
The probing technique (Figure~\ref{fig:probing}) was developed and used extensively prior to the introduction of the MI field for investigating whether model activations encode various linguistic features such as part-of-speech~\cite{conneau2018you, tenney2019bert, tenney2019you, antverg2021pitfalls}. MI studies have also adopted probing as an important tool for investigating whether a pre-defined feature is present in intermediate activations. Specifically, it involves training a shallow (or linear) classifier, known as a \emph{probe}, to predict whether a feature is present in those activations \cite{gurnee2023finding, nikankin2024arithmetic, nanda2023emergent}.
However, it is important to note that the results of probing analyses only indicate a correlation, not a causal relation, between the feature and activations. We refer readers to \citet{belinkov2022probing} for further study on various types of probing.

\begin{figure}[t!]
    \centering
    \includegraphics[width=0.6\linewidth]{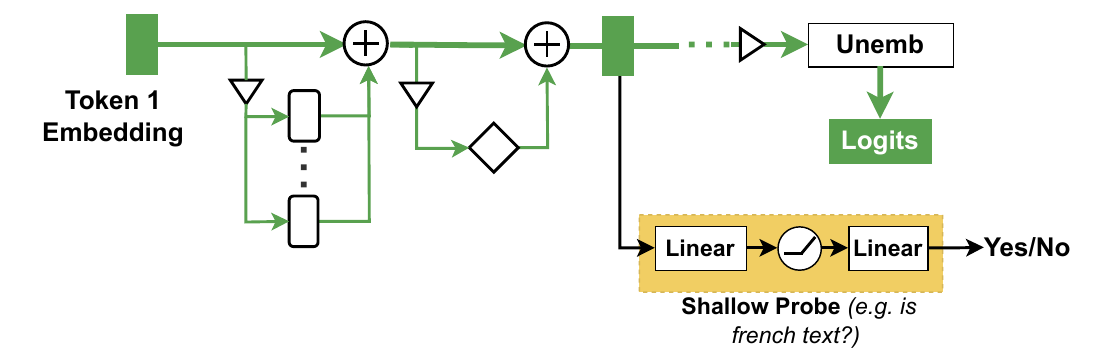}
    \caption{Probing on RS to detect whether it encodes a ``French text'' feature.}
    % \vspace{-3mm}
    \label{fig:probing}
\end{figure}
\begin{figure}[t!]
    \centering
    \includegraphics[width=0.85\linewidth]{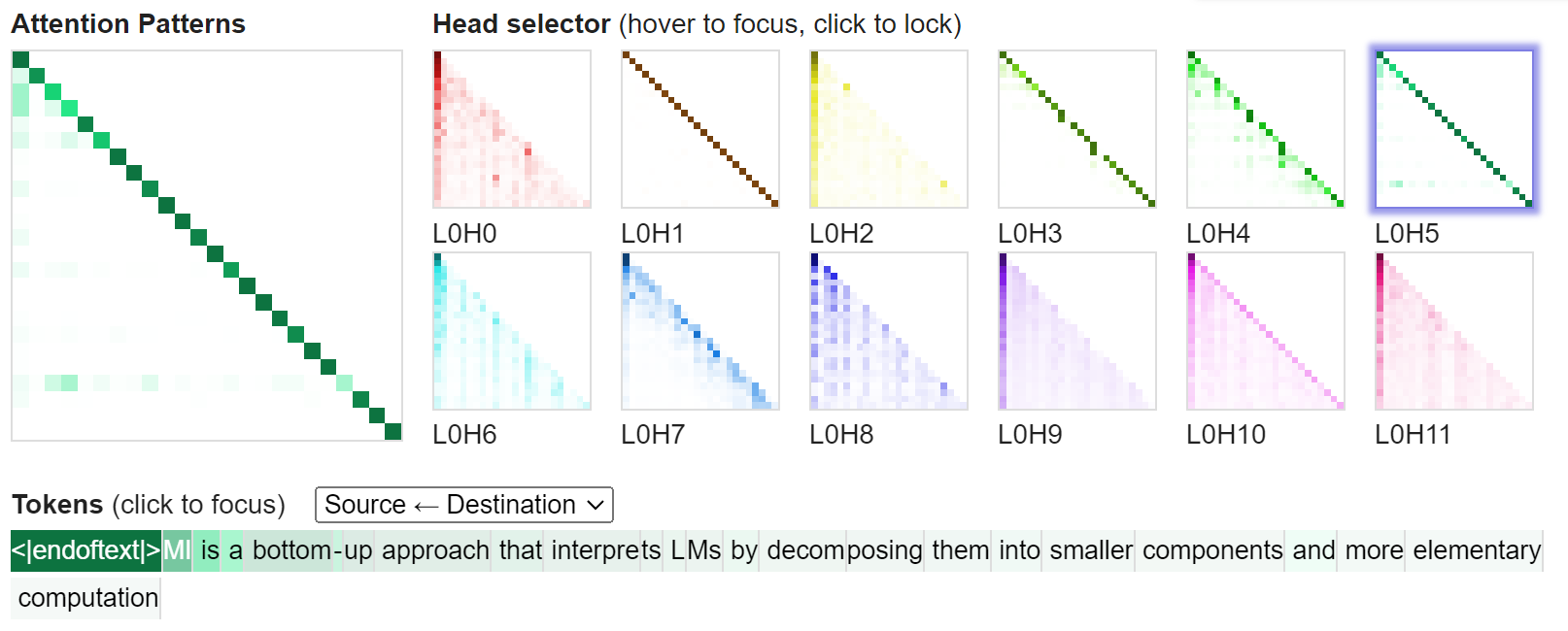}
    \caption{Attention visualization, created using the tool of \citet{cooney2023circuitsvis}.
    % \abu{this figure is rather zoomed out and difficult to read. perhaps this would be better depicted as a two-column figure?}
    }
    % \vspace{-2mm}
    \label{fig:visualization}
\end{figure}

\subsubsection{Visualization} \label{subsec: visualization}
Visualization (Figure~\ref{fig:visualization}) is employed across various stages of an MI investigation, from generating initial hypotheses to refining them, conducting qualitative analyses, and validating results. For instance, attention patterns are often visualized to understand attention heads \cite{lieberum2023does, olsson2022context}, and a neuron activation across the input text is visualized to identify its functionality~\cite{elhage2022solu, bricken2023monosemanticity}. While visualization can be highly useful, it requires human effort to interpret results and carries the risks of overgeneralization. Thus, any claims need to be substantiated with further experimentation and analysis.

\section{A Beginner's Roadmap to MI} \label{sec: roadmap}

A key motivation for this survey is to provide a friendly guide for researchers and developers interested in MI to quickly pick up the field. To this end, we provide a \emph{beginner's roadmap} in Figure~\ref{fig: feature-study-roadmap}, \ref{fig: circuit-study-roadmap}, and \ref{fig: universality-roadmap}, where we categorize MI research into three categories -- feature study, circuit study, and the study of universality, corresponding to the three objects of studies (Section~\ref{sec:objects of study}).
% \abu{weren't these defined as ``objects'' rather than ``objectives''? i was interpreting ``object'' in the paper so far in the sense of a ``thing'' rather than in the sense of a goal (e.g., ``the ball is an object'' vs ``the object of the game is to get the most points'')}
Each category is further divided into sub-categories with their corresponding MI workflow and associated techniques. 
% In Section~\ref{sec: case-studies}, we will then present \emph{case studies} of the MI workflows based on prior work, such that beginners can easily map between the workflow action and the actual research activity. 

% Due to the space limitation, techniques involved in the roadmap will only be briefly described. Details about each technique and additional case studies are included in Appendices~\ref{app:techniques}-\ref{?}.}
% More details are included in Appendix~\ref{app:techniques}.}

\subsection{Feature Study}
\begin{figure*}[t!]
    \centering
    \includegraphics[width=0.9 \linewidth]{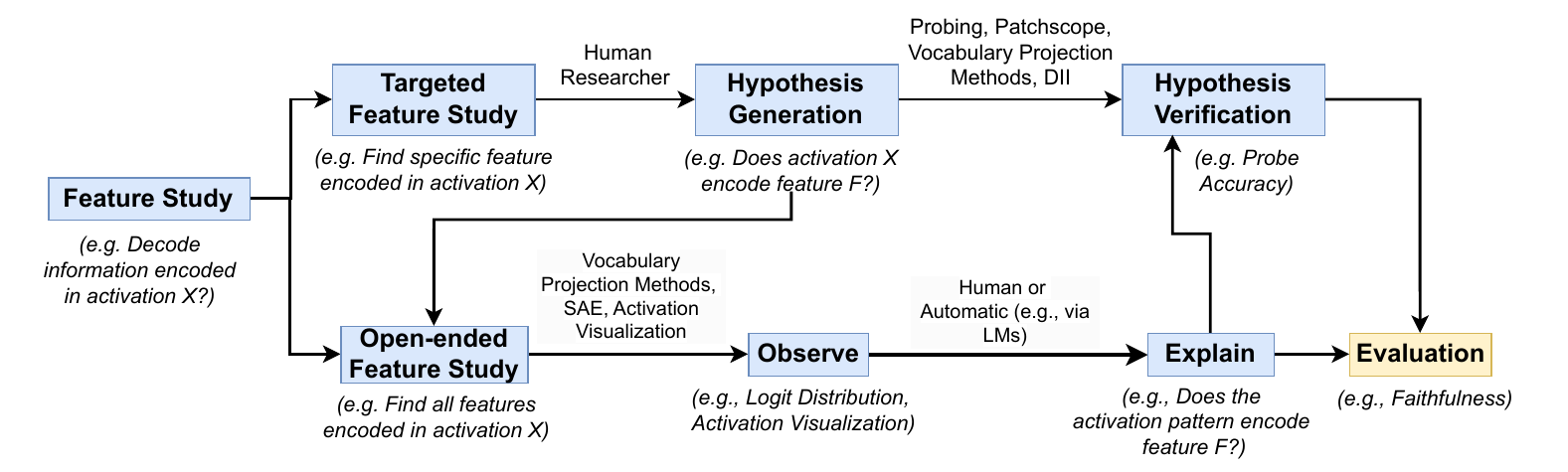}
    \caption{A task-centric beginner's roadmap for feature study.}
    \vspace{-3mm}
    \label{fig: feature-study-roadmap}
\end{figure*}
Feature studies can be broadly categorized into \emph{targeted} and \emph{open-ended} feature studies. 

\paragraph{Targeted Feature Study}
A targeted feature study aims to determine whether a \emph{pre-defined feature} is encoded in the representation of an LM. These pre-defined features are typically intuitive guesses or hypotheses made by humans, and we perform a targeted feature study when we want to verify the presence or absence of these features. Therefore, a general research question in a targeted feature study is, \emph{``Does this LM representation (or a subset of its neurons) encode the feature $F$?''}

\paragraph{Open-ended Feature Study} 
While targeted feature discovery is useful for determining whether a specific pre-defined feature is present in an activation, it relies on human intuition to hypothesize which features might be encoded. This could be problematic, as there may be features encoded within LM activations that do not align with human intuition. Consequently, targeted feature discovery alone is insufficient for fully interpreting an LM's activations. To address this issue, {open-ended feature discovery} aims to enumerate all the features encoded within the activations of an LM. A general research question in an open-ended feature study is thus, \emph{``What are all the features encoded in this LM representation?''}

\subsubsection{General Workflow for Targeted Feature Study}\label{subsec: feature-workflow}

% \zyc{I added "hypothesizing a pre-defined feature...". Both this step and the "Verifying the Hypothesis" step are missing in the current roadmap diagram.}

% \paragraph{Hypothesizing a Pre-defined Feature and Forming Research Questions}
\paragraph{Hypothesis Generation}
A targeted feature study begins with a hypothesis about the presence or absence of a pre-defined feature $F$ in an LM representation. For example, suppose we are interested in knowing whether an LM activation encodes the \emph{``is\_python\_code''} feature in its activations when the input contains a Python code snippet. Then, our hypothesis would be: \emph{``Does this LM activation (e.g., $h_i^{1}$) encode the \emph{`is\_python\_code'} feature?''} We can refine our hypothesis to localize the feature encoding to a more granular level: \emph{``Does this neuron in this activation (e.g., Neuron 1 in $h_i^{1}$) encode \emph{`is\_python\_code'} feature?''}. Note that we can iteratively pose this question to all the LM activations when we want to identify a complete set of activations that encode the targeted feature. 

% \paragraph{Verifying the Hypothesis} 
\paragraph{Hypothesis Verification} 
To verify whether or not a pre-defined feature exists in a given LM representation, there are typically two workflows to consider. The first workflow directly probes the LM representation and inquires the presence of the targeted feature $F$, whereas the second workflow follows a similar procedure as the open-ended feature study, which first identifies all features encoded in the representation and then confirms whether the feature $F$ is among them. Below, we mainly present the first workflow. Details of the second workflow will be discussed in Section~\ref{sub-sec: open-ended-workflow}.
For the first workflow, one option is to use the \emph{Probing} method (Section~\ref{subsec: probing-tech}) to verify the hypothesis, where a probe determining the existence of the targeted feature $F$ will be trained. Probing is a lightweight and easy-to-implement technique with a substantial body of literature discussing its strengths and limitations.  \emph{PatchScope}~\cite{ghandeharioun2024patchscope} is another technique that can also be used for verifying the hypothesis. This method is especially useful to verify whether certain attributes associated with a given subject (e.g., attributes associated with the ``largest city of'' from the representation of ``United States'') are encoded in the LM representation. Unlike probing, these methods require no training. {However, PatchScope relies on humans to craft an expressive inspection prompt capable of eliciting the desired feature, which can be particularly challenging for abstract features. Besides PatchScope, other vocabulary projection methods can also be used to check hypotheses about features that are expressible in the vocabulary space. {Beyond these approaches, \emph{DII} \cite{geiger2024finding} offers a more fine-grained method for hypothesis verification. DII operates at the subspace level, enabling interventions when a hypothesis posits that a given causal variable is encoded not in the full activation but in a distinct subspace of the activation. Concretely, DII applies a change-of-basis transformation to rotate the representation, intervenes on a $k$-dimensional subspace corresponding to the high-level causal variable, and then maps the representation back to the original basis. Complementary to DII, \emph{DAS} \cite{geiger2024finding} is a supervised approach that automatically discovers both the change-of-basis transformation and the $k$-dimensional subspace best aligned with a given causal variable. In short, DII enables fine-grained causal interventions, while DAS identifies the appropriate subspaces in which those interventions should be applied. The hypothesis can be validated based on the intervention effect of DII or whether DAS can identify an appropriate subspace for the targeted causal variable. However, both DII and DAS require access to training data composed of clean-counterfactual pairs.}
Therefore, the choice of these techniques depends on the type of feature being investigated as well as the available computational and data resources. We also summarize the advantages and limitations of the two techniques in Table~\ref{tab:feature-study-tech-comparison}. 

\begin{table*}[t!]
    \centering
    \resizebox{\textwidth}{!}{%
    \begin{tabular}{
        >{\centering\arraybackslash}m{4.2cm}
        >{\raggedright\arraybackslash}m{7cm}
        >{\raggedright\arraybackslash}m{7cm}
    }
    \toprule
    \textbf{Techniques} & \textbf{Advantages} & \textbf{Limitations} \\
    \toprule

    \multicolumn{3}{c}{\textbf{Targeted Feature Study}} \\
    \midrule

    Probing &
    Established methodology with a large body of literature; useful for identifying a wide range of feature types (e.g., syntactic or semantic features).  &
    Require training data; \emph{correlational only} (not causal evidence); confirms only feature presence (not usage). \\
    \midrule

    PatchScope &
    No additional training; lightweight. &
    Require well-crafted \emph{inspection prompts}; unsuitable for discovering abstract features (e.g., syntactic features). \\
    \midrule

    Vocabulary Projection Methods &
    No additional training; lightweight. & Only decode features that can be represented in vocabulary space; early-layer projections can be unreliable. \\

    \midrule 
    Distributed Alignment Search (DAS) & Discovers the specific $k$-dimensional subspace associated with a causal variable from a given representation. & Dataset with clean-counterfactual pairs required; computationally expensive. \\
    \midrule

    Open-ended Feature Discovery &
    Can uncover unexpected related features. &
    Computationally expensive; human-in-the-loop labeling. \\
    \midrule

    \multicolumn{3}{c}{\textbf{Open-Ended Feature Study — Observe (Step 1)}} \\
    \midrule

    Neuron Activation Visualization &
    Intuitive; no additional training. &
    Neurons encode polysemantic features that are challenging to interpret. \\
    \midrule

    Sparse Autoencoders (SAE) &
    Project activations to sparse activation with \emph{more monosemantic} features, improving interpretability of activations. &
    Require SAE training; high compute cost; sparsity–reconstruction trade-offs. \\
    \midrule

    Vocabulary Projection Methods &
    Provide important signals on how a given activation or feature impacts LM output. &
    Only decode features that can be represented in vocabulary space; early-layer projections can be unreliable without translators. \\
    \midrule

    \multicolumn{3}{c}{\textbf{Open-Ended Feature Study — Explain (Step 2)}} \\
    \midrule

    Human &
    Gold standard in current practice. &
    Expensive and time-consuming; subjective; polysemantic features are hard to label consistently across annotators. \\
    \midrule

    Automatic LM &
    Reduce human effort; scalable first-pass explanations and scoring; consistent formatting. &
    Faithfulness and reliability of explanations must be validated; risk of model bias or hallucination; may bias human reviewers. \\
    \midrule

    Human + Automatic LM &
    Efficient division of labor: LMs propose/cluster, humans verify/refine; improve efficiency while maintaining quality &
    Still need human oversight; potential anchoring on LM suggestions. \\
    \bottomrule
    \end{tabular}
    }
    \caption{Comparison of techniques across \emph{Targeted} and \emph{Open-ended feature study} stages (Observe \& Explain).}
    \label{tab:feature-study-tech-comparison}
\end{table*}

 \subsubsection{General Workflow for Open-ended Feature Study} \label{sub-sec: open-ended-workflow}
 Open-ended feature study typically involves two steps: \textbf{(1) Observe:} generate an \emph{intermediate explanation} (e.g., logit distribution, activation visualization) for each activation; and \textbf{(2) Explain:} interpret the intermediate explanation to discover features encoded within the activations.
 % ; and (3) \textbf{Evaluation:} assessing the faithfulness of the discovered features.

 \paragraph{Observe} In this step, we employ various MI techniques to gather various information about the activation to guide open-ended feature discovery. The gathered information can be referred to as ``intermediate explanations'' because they require further interpretation during the Explain step to discover the exact features. {Techniques such as vocabulary projection methods (Section~\ref{tech: vocab-proj-methods}), neuron activation visualization (Section~\ref{subsec: visualization}), and SAEs (Section~\ref{sec: sae}) are commonly used techniques for generating intermediate explanations. We provide the advantages and limitations of these techniques in Table~\ref{tab:feature-study-tech-comparison}.}

 The vocabulary projection method generates a {logit distribution} as the intermediate explanation. As we elaborated in Section~\ref{tech: vocab-proj-methods}, one can then infer the encoded features by examining tokens with the highest logits from the distribution. However, it is important to note that the vocabulary projection method can only discover features in the prediction space. In other words, it only discovers features that are directly used for the next token prediction. {For example, while completing the following input \emph{``English translation for nourriture is''} with \emph{``food''}, the LM might have extracted the \emph{is\_french\_text} feature from the input and encoded it in its activations. However, the vocabulary projection method is unlikely to reveal the presence of the \emph{is\_french\_text} feature.} Consequently, intermediate explanations from vocabulary projection methods are not sufficient for discovering all features encoded in a given activation. 

 Neuron activation visualization (Sec~\ref{subsec: visualization}) is another technique that involves developing an interface to visually highlight the text or tokens that elicit information about a specific neuron. The idea here is that, for example, if a neuron consistently activates for text written in French, one can then label the neuron as encoding a French text detection feature. It is important to note that neuron activation visualization is based on an important assumption that features are represented by neurons in a one-to-one correspondence.
However, subsequent studies~\cite{elhage2022solu, marks2024sparse} have found \emph{polysemantic neurons}, i.e., neurons that activate for multiple unrelated features, rendering the assumption of one-to-one correspondence to be false. The discovery of polysemantic neurons also complicates the {Explain step}, since polysemantic neurons can easily confuse the explainer when they activate for multiple unrelated features simultaneously. To address this challenge, subsequent studies have proposed training {SAEs} (Section~\ref{sec: sae}) that will transform the original activation into a higher-dimensional sparse activation. The goal of this projection is to generate a new representation that faithfully retains all the features from the original activation but encodes them in a way that each neuron corresponds to a single feature, resulting in a representation composed of more \emph{monosemantic neurons}. The neuron activation visualization can then be applied to the sparse activation, rather than the original activation, for generating more interpretable intermediate explanations. However, training SAEs for specific LMs requires non-trivial computation, and SAEs, as we discussed in Section~\ref{sec: sae}, face weaknesses such as the sparsity-reconstruction trade-off.

Although each of these techniques can be used individually to generate the intermediate explanation in the {Observe step}, in practice, they are often combined to provide a more complete explanation of the LM activation. Specifically, the intermediate explanations from various MI techniques are generated and presented in an interface for the human evaluators to annotate the feature description in the Explain step. In Section~\ref{sec:observation_interfaces}, we will discuss three examples of interfaces from prior work.

 \paragraph{Explain} The intermediate explanation generated in the {Observe} step is then interpreted by explainers to uncover features encoded in the representation. For instance, when a vocabulary projection method is used in the Observe step, a human explainer views the top tokens in the projected {logit distribution} and labels the feature implied by the tokens (e.g., if top tokens are all associated with arithmetic addition, then a human evaluator infers that the activation is encoding features relevant to additive operation in arithmetic [\citealt{rai2024investigation}]).
 % the ``arithmetic addition'' feature~
 % \abu{what is the definition of the addition feature?}
 Similarly, for neuron and SAE activation visualization, a human explainer is instructed to examine the activation visualization for each neuron, and then indicate whether they have found a plausible theory to explain the activations. For instance, \citet{elhage2022solu} provided the following instructions to human explainer -- ``mark INTERPRETABLE if 80\% or more of the strongest firings can be explained by a single rule or category (e.g. the word `apple', or any phrase relating to music), and NOT INTERPRETABLE otherwise''. 
 While the explainers are often humans, we can also automate this Explain step using machines, such as \emph{LLMs as explainers}.
 This involves providing LLMs with intermediate explanations and prompting them to label the corresponding features, if applicable. For instance, \citet{bills2023language} provided intermediate explanations from both the vocabulary projection method and the neuron activation visualization to a GPT-4 model for feature labeling; the authors reported a strong correlation between human and GPT-4-generated explanations, supporting the feasibility of leveraging LLMs for feature explanation. {In an interface like Neuronpedia~\cite{neuronpedia}, human evaluators and automatic LM explanations (i.e., the ``auto-interp explanation'' on the Neuronpedia interface in Figure~\ref{fig: neuronpedia-interface}) approach are combined to improve the efficiency of annotating feature explanations, as discussed in Section~\ref{sec:observation_interfaces}. Specifically, human evaluators are provided with LM-generated explanations as intermediate references alongside other intermediate explanations when formulating the final explanation. While this human–LM combination can accelerate the annotation process, it also risks biasing evaluators, who may become overly reliant on the LM’s output instead of producing independent and precise explanations.} 
 % \zyc{@Daking, briefly talk about the human+automatic LM approach in the technique. If this is not done by prior work and is more like a potential, we can directly say it, e.g., "Potentially, one can combine humans and LMs for a hybrid explanation, which ...[advantages]. However, we also imagine that ..[limitations]"}

\subsubsection{Example Interfaces for Making Observations in Open-Ended Feature Study}\label{sec:observation_interfaces}
{We highlight two examples of interfaces from prior work that enable open-ended feature discovery: (1) Annotating neurons with features by \citet{elhage2022solu} (Figure~\ref{fig: neuron-activation}), and (2) Feature interpretation using Neuronpedia~\cite{neuronpedia} (Figure~\ref{fig: neuronpedia-interface}).

% and (3) Anthropic’s tool for interpreting SAE features~\citep{bricken2023monosemanticity} (Figure~\ref{fig: anthropic-interface}).}

\begin{figure}[t!]
    \centering
    \includegraphics[width=0.80\linewidth]{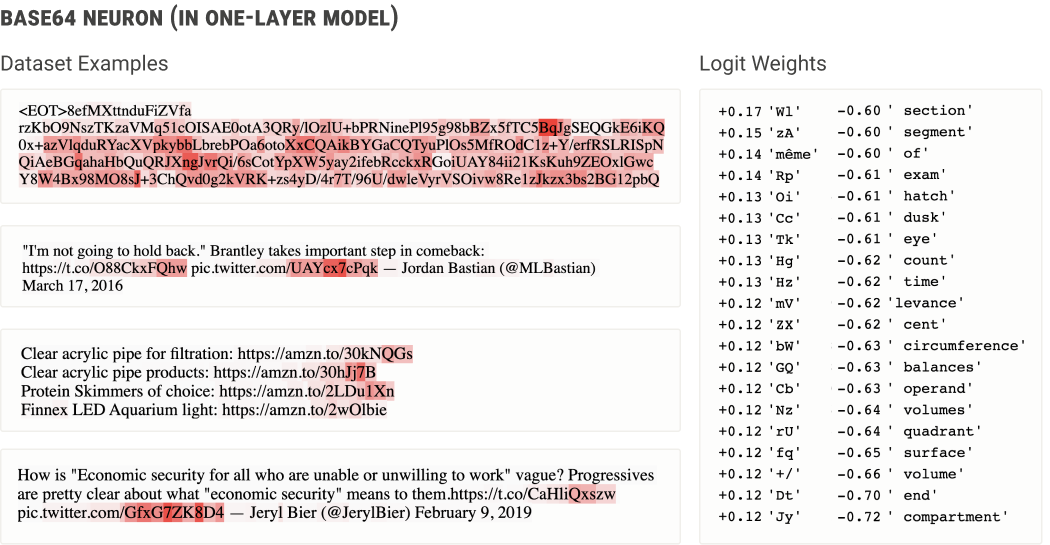}
    \caption{The neuron activation visualization developed by \citet{elhage2022solu}, which labels neurons with features by analyzing the text on which they activate.
    }
    \vspace{-2mm}
    \label{fig: neuron-activation}
\end{figure}

 {\paragraph{Annotating Neurons with Features}  \citet{elhage2022solu} designed a simple interface to perform open-ended feature discovery, when they analyzed the FF neurons of a one-layer transformer model, as shown in Figure~\ref{fig: neuron-activation}. The interface shows the intermediate explanations generated by neuron activation visualization (left) and vocabulary projection (right). Neuron activation visualization consists of highlighted text snippets that were sampled from a large corpus, focusing on paragraphs where the neuron exhibits the highest activation. We can see that the neuron seems to consistently activate on text encoded in base 64. Besides neuron activation visualization, the vocabulary projection method is applied to the same neuron to identify which output tokens it promotes or suppresses. The neuron seems to increase the logits of random mixed-case string tokens and suppresses common English words. This further supports the hypothesis that the neuron encodes base64-related features, as base64 strings often occur in sequence—making it sensible for the neuron to promote base64-like tokens as the next prediction. In summary, looking at the intermediate explanations from both the vocabulary projection method and neuron activation visualization, human evaluators can annotate the neuron to encode the base-64 feature.}

 {\paragraph{Feature Interpretation using Neuronpedia} 
  \begin{figure}[t!]
    \centering
    \includegraphics[width=0.80\linewidth]{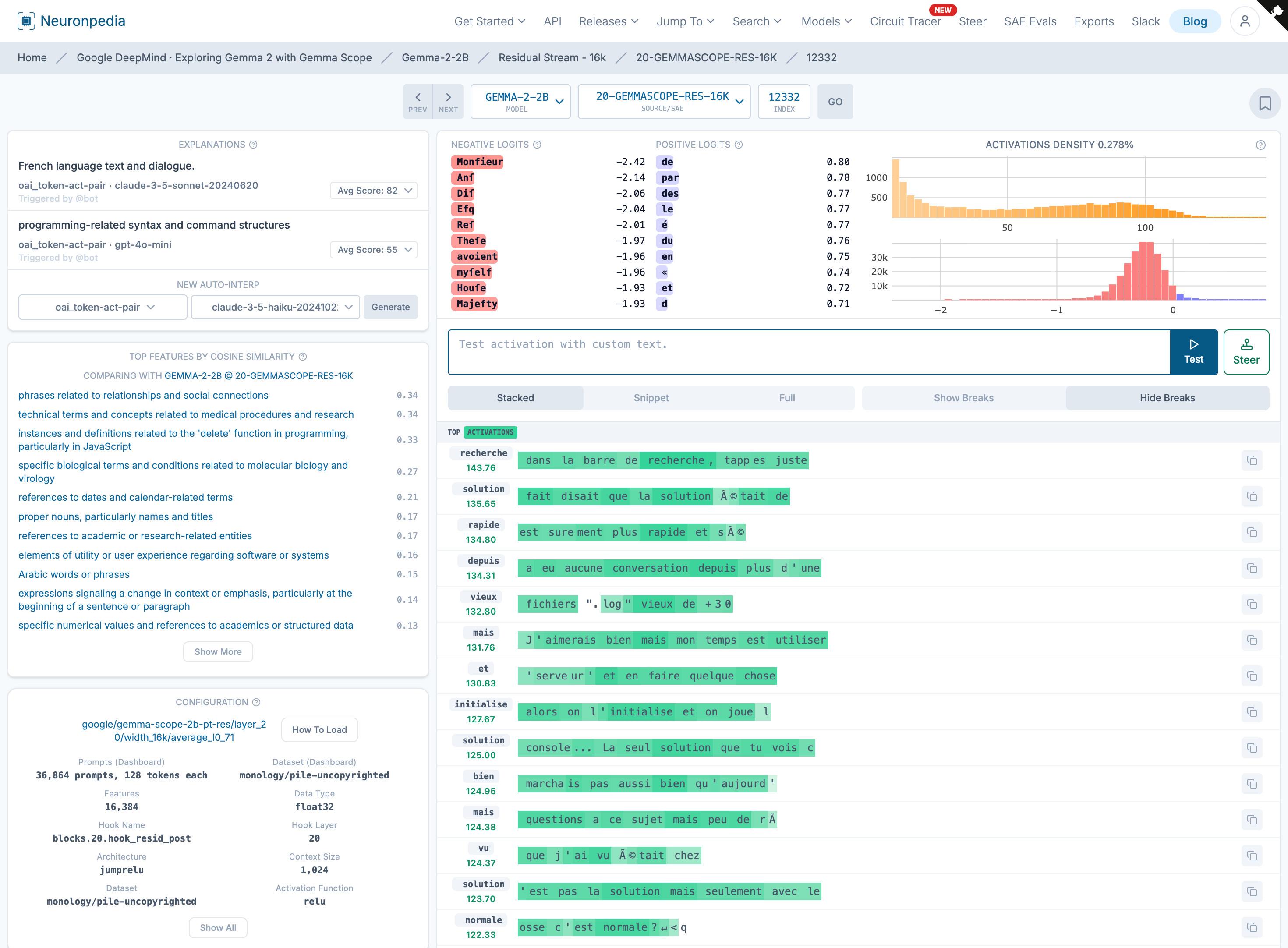}
    \caption{Example interface of Neuronpedia \citep{neuronpedia} for conducting open-ended feature studies.
    % Example interface designed for conducting an open-ended feature study, taken from Neuronpedia~\citep{neuronpedia}. 
    The example demonstrates a collection of intermediate explanations for interpreting an SAE feature (index 12,332) encoded in the residual stream activation at layer $20$ of the Gemma2-2b model. A human evaluator labels or explains the feature based on information present in this interface. 
    % \zyc{The caption makes me think that Neuronpedia is a general-purpose feature study platform, but the main text describes it only from the perspective of SAE feature interpretation. This is confusing.}
    % A human evaluator uses this interface to label or explain the feature, based on the information presented within the interface.
    }
    \vspace{-2mm}
    \label{fig: neuronpedia-interface}
\end{figure} 
Neuronpedia~\citep{neuronpedia} is a popular open-source platform for performing open-ended feature discovery. {In particular, Neuronpedia serves as a general-purpose platform where researchers can upload collections of feature vectors discovered through different techniques (e.g., SAEs, custom probe vectors, etc.) for a given LM. The platform then generates an interactive dashboard for each feature vector by showcasing various intermediate explanations about the feature to support its open-ended feature study, facilitating both the \emph{Observe} and the \emph{Explain} steps.} Figure~\ref{fig: neuronpedia-interface} shows an example interface for conducting an open-ended feature study of residual stream activation at layer $20$ ($h^{20}_i$) of Gemma2-2b model \cite{team2024gemma}. More specifically, the interface corresponds to a single SAE feature (index $12{,}332$) from the $16{,}000$ features obtained by projecting $h^{20}_i$ into a sparse representation ($20\text{-}\text{GEMMASCOPE}\text{-}\text{RES}\text{-}16k$) via an SAE. The interface consists of a collection of \emph{intermediate explanations}, produced using various MI techniques, to guide human evaluators in the \emph{Explain} step, as listed below:  
 \begin{itemize}
     \item Auto-Interp Explanation (upper-left panel): A concise natural language description (e.g., ``French language text and dialogue'' or ``Program-related syntax and command structure'') generated by prompting LLMs of your choice (e.g, GPT-4~\citep{achiam2023gpt}) with the top-activating text snippets for a given neuron or feature~\citep{bills2023language}. Multiple, potentially unrelated, descriptions may be produced for the same SAE feature, as SAE features can sometimes be polysemantic. Accordingly, a quantitative measure (ranging from 0 to 1) that indicates how well the automatically generated explanations capture a feature's behavior can be computed by employing another scoring LLM (``No Scores'' currently shown in the demo interface). Following \citet{bills2023language}, the scoring is performed by comparing the feature's actual activation pattern with the activation predicted by the scoring LLM based on the auto-interp explanation. A higher score means the explanation consistently and accurately predicts the feature’s true activations, while a lower score suggests the explanation is incomplete or inaccurate.
     % \item Auto-Interp Score: A quantitative measure (ranging from 0 to 1) that indicates how well an \emph{auto-interp explanation} captures a feature’s behavior. To compute this score, an LLM of your choice (e.g., GPT-4~\citep{achiam2023gpt}) is given the explanation and asked to predict whether the feature will activate on a set of unseen text snippets. The predicted activation pattern is then compared to the feature’s actual activation pattern~\citep{bills2023language, paulo2024automatically}. A higher score means the explanation consistently and accurately predicts the feature’s true activations, while a lower score suggests the explanation is incomplete or inaccurate.
     \item Top Features by Cosine Similarity (middle-left panel): Features ranked based on their cosine similarity with the feature under investigation (i.e., feature of index $12{,}332$). The cosine similarity is calculated by comparing the target feature's vector (i.e., the corresponding row vector of the SAE decoder, $W_{dec}[12332,:] \in \mathbb{R}^d$) with other features (i.e., $W_{dec}[j,:] \in \mathbb{R}^d, j \neq 12332$).

     \item {Configuration (bottom-left panel): Configurations of the SAE architecture and the dashboard visualization, such as the dataset used to create the activation density plot and feature activation visualizations.
     % Configuration used to train the SAE, including details such as the training data and SAE architecture. It also provides information about additional settings used to generate the dashboard’s explanatory visualizations—for example, the datasets used to create activation density plots and feature activation visualizations. 
     The dataset can either be the original SAE training data or a different public dataset, such as pile-uncopyrighted~\citep{pile} shown in the example interface. This flexibility is especially useful when the SAE training corpus is proprietary or confidential.}
     % Features ranked by measuring the cosine similarity between their activation vectors and other feature directions, helping identify other features with potentially similar behavior. 
     \item {Statistical Information (upper-right panel): Upper-right panel consists of three intermediate explanations - \emph{vocabulary projection} (left), \emph{activation density} (top-right), and \emph{logit density histogram} (bottom-right).
     \begin{itemize}
         \item Vocabulary Projection (left): 
         % Each SAE feature corresponds to a feature vector, a row vector of the decoder matrix $W_{dec} \in \mathbb{R}^{s \times d}$, which can be 
         The feature vector $W_{dec}[12332,:]$ is projected to a vocabulary space or logit distribution by multiplying with $W_U$. The interface then displays the top-10 tokens with the highest and lowest logit scores to investigate the effect of the feature on the model output.  
         \item Activation Density (top-right): Histogram of randomly sampled non-zero activation values when provided with text input sampled from the dataset used to train SAEs or other public datasets (e.g., pile-uncopyrighted~\citep{pile}). 
         \item Logit Density Histogram (bottom-right): Logit density histogram shows the logit distribution obtained from vocabulary projection of the feature vector. In the figure, we can observe that the feature vector assigns a negative logit score to most output tokens, which could indicate that their primary role could be the suppression of the set of output tokens.
     \end{itemize}}
     % \zyc{@Daking, can you merge vocab proj and activation density here, as they are organized in one panel?} \daking{Explain how SAE features are projected!}
     % \item Vocabulary Projection: Top-10 tokens promoted and suppressed by the feature as next-token prediction, obtained using the vocabulary projection method.\zyc{I do not understand, based on your description, what is projected using logit lens. The "feature" is only one neuron, right?}
     % \item Activation Density: Histogram of randomly sampled non-zero activations. It shows how often and how strongly a feature is activated. 
     % \item Logit density histogram \zyc{which part do you refer to? I don't see a correspondence in the figure. If I misunderstood it, please revise this item and indicate the panel position.}: This shows how strongly a neuron promotes or suppresses the output tokens. It may be useful for deriving various insights about the feature. For example, a long-tailed distribution skewed toward negative values suggests that the neuron generally suppresses most tokens, offering insight into the feature effect in the output.
     \item Test activation with custom text (middle-right panel): An interactive form that allows practitioners to input custom text and observe whether the neuron activates. This helps validate the evaluator’s hypothesis. For example, if the feature is believed to detect French text, the evaluator can enter French and non-French texts to test whether the feature's activation pattern aligns with the hypothesis. 
     \item Steer (middle-right panel): {The interface includes a steering option that allows you to manually activate a feature with chosen activation strength and observe its influence on the model’s output. The {Steer button} is located next to the input form for testing activations with custom text, and it links to a separate interface designed for steering experiments.} 
     \item Feature Activation Visualization (bottom-right panel): The visualization highlights tokens in input texts, {randomly sampled from the dataset used to train SAE or other public datasets (e.g., pile-uncopyrighted~\citep{pile}).}
% \zyc{is the evaluator expected to provide the texts, or do you mean the texts from SAE's pre-training corpus?} 
that fall within specific activation intervals. To do this, activation values are divided into evenly spaced fractions of the feature’s maximum activation, and the corresponding texts that fall within each interval are extracted. This allows us to explore how the interpretability of a feature changes with varying activation strength. 
 \end{itemize}
 }

\begin{figure}[t!]
    \centering
    \includegraphics[width=0.9\linewidth]{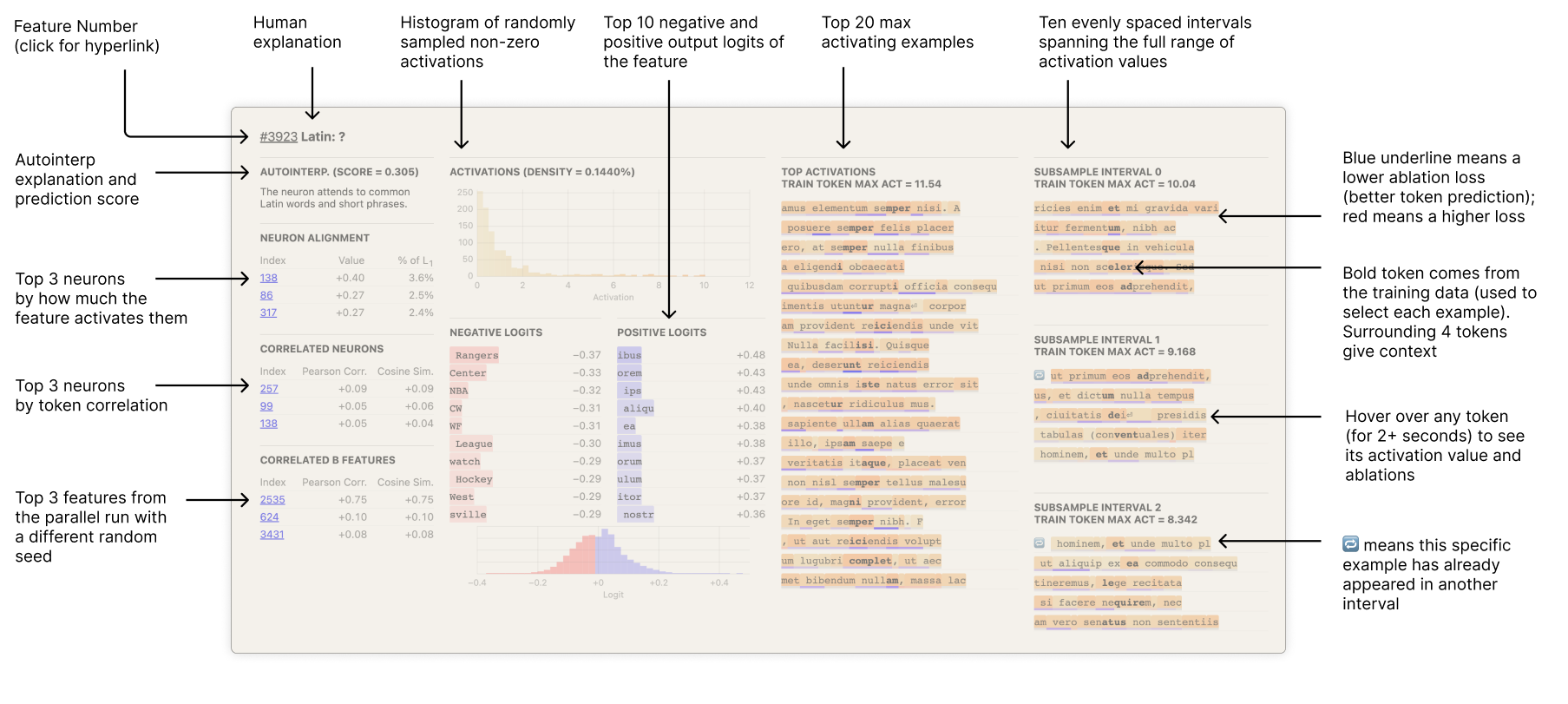}
    \caption{Example interface designed by \citet{bricken2023monosemanticity} for SAE feature interpretation, similar to the interface of Neuronpedia. However, the interface was designed mainly for demonstration purposes and is not a general-purpose SAE interpretation tool. It is not open-sourced either.}
    % \zyc{move fig to the same page of the main-text discussion}}
    \vspace{-2mm}
    \label{fig: anthropic-interface}
\end{figure}
 
{Now, based on these intermediate explanations, a human evaluator annotates the feature with a description in the \emph{Explain} step. Neuronpedia already consists of completed \emph{Observe} step for multiple open-weight models, including GPT-2~\citep{radford2019language}, Llama-2~\citep{touvron2023llama}, and Gemma-2~\citep{lieberum2024gemma}, allowing users to begin directly from the \emph{Explain} step. However, one can also upload new models along with trained SAE and other meta information (e.g., dataset for training SAE) to perform open-ended feature discovery on new models as well. {Similar to Neuronpedia, Figure~\ref{fig: anthropic-interface} is another example of an interface designed by Anthropic~\cite{bricken2023monosemanticity}. It is created to perform an open-ended feature discovery on the SAE trained for the one-layer toy model analyzed in the paper. The interface presents overlapping intermediate explanations as Neuronpedia, though with fewer explanation types and slight differences in presentation. However, the interface is mainly designed for demonstration purposes. It is not open-sourced and is unavailable for use on new models by the broader research community.}
 % \zyc{does it work in practice? new models will need new SAEs. Relevant to my earlier question, I wonder if the user needs to provide input texts or not.}

 % {\paragraph{Anthropic Interface for SAE Feature Interpretation.} Figure~\ref{fig: anthropic-interface} is another example of an interface designed for interpreting SAE features by Anthropic \cite{bricken2023monosemanticity}. The interface is interactive and can be accessed in \cite{bricken2023monosemanticity}. While the tool provides a valuable demonstration, its scope is restricted to the two-layer toy model analyzed in the paper and cannot be used for studying new models like Neuronpedia. Functionally, it mirrors Neuronpedia in offering intermediate explanations of SAE features to facilitate open-ended feature studies, with most explanations overlapping with those described above. 
 % \zyc{This is too brief. Can you add more details, such as what models this interface supports, is it a platform open for practical research or only for demonstration purposes? Also briefly comment on the difference compared to Neuronpedia.}  
 }

 % \paragraph{Evaluation} 
 \subsubsection{Evaluation of Feature Study}
 {The evaluation of a feature study is typically performed in two dimensions: \emph{faithfulness} and \emph{interpretability}. Specifically, faithfulness measures whether the discovered feature truly exists in the LM representation, while interpretability assesses how easily a human can understand the feature description. Measuring faithfulness is inherently challenging due to the absence of ground truth. To this end, most studies often rely on technique-specific proxy measures, such as probing accuracy on a held-out test set for targeted feature studies and reconstruction loss when using SAEs for open-ended feature studies, where a higher accuracy or a lower reconstruction loss indicates greater faithfulness. In addition, some studies~\cite{templeton2024scaling, marks2024sparse} also proposed manually altering the feature value, referred to as \emph{activation steering}, during model inference to measure the direct effect of a feature on the next-token distribution of the model and causally measure the faithfulness of the feature study. On the other hand, interpretability of discovered features is typically conducted manually by a human or automatically by an LM. For instance, in the work of \citet{bricken2023monosemanticity}, a human is instructed to provide a score based on how interpretable the discovered features are, following a guideline ``on a scale of 0–3, rate your confidence in this interpretation''. Alternatively, one can also use LMs to calculate an automatic explanation score~\cite{bills2023language} as a proxy for manual human evaluation.}

\subsection{Circuit Study}\label{subsec:roadmap-circuit-study}
\begin{figure*}[t!]
    \centering
    \includegraphics[width=\linewidth]{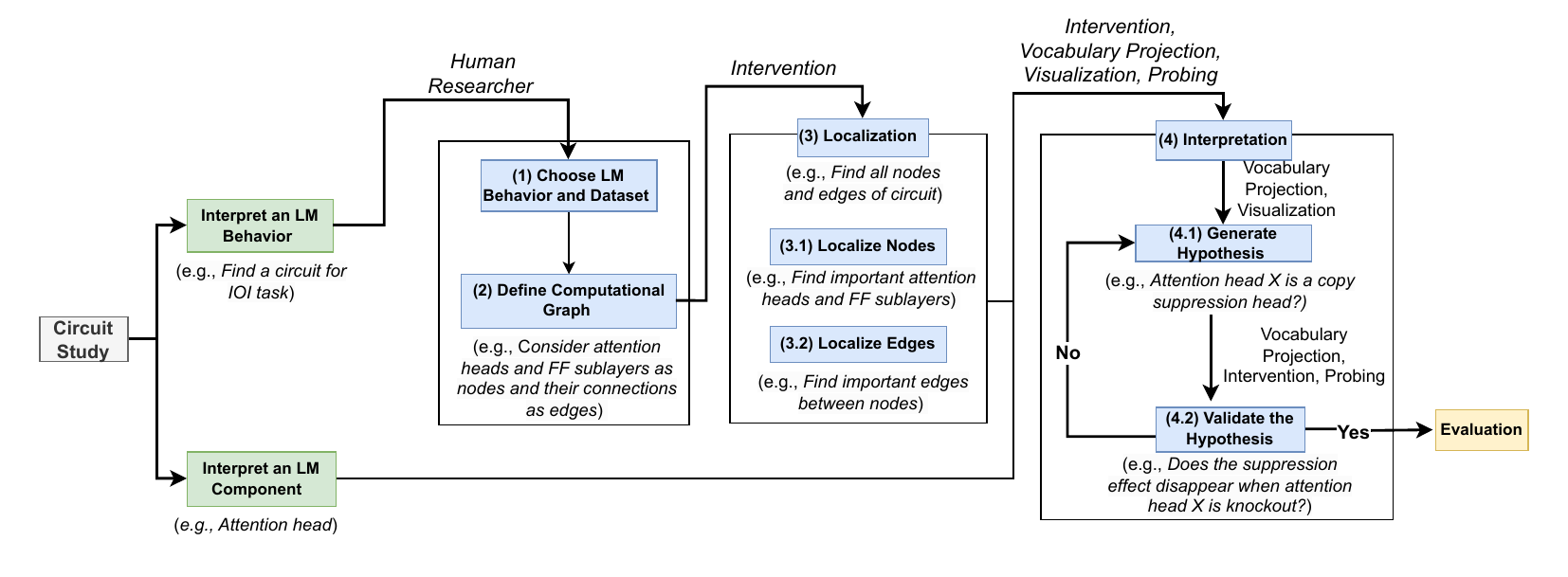}
    \caption{A task-centric beginner's roadmap for circuit study.}
    % \zyc{Capitalize all words in each term. Term use should be consistent with the main text -- Step 4 is called "Interpretation" in the main text. (4.2) change to "Validate the Hypothesis"}}
    % \vspace{-3mm}
    \label{fig: circuit-study-roadmap}
\end{figure*}
Similarly, the study of circuits (Figure~\ref{fig: circuit-study-roadmap}) can be broadly divided into two categories, i.e., interpreting \textit{an LM behavior} and interpreting \textit{an LM component}.

\paragraph{Interpreting an LM Behavior} This category of study involves identifying the circuit responsible for a specific LM behavior. For example, \citet{elhage2021mathematical} investigated the circuits to explain the two-layer toy LM's capability in sequence completion tasks and
% \zyc{what are the tasks? can they be more specific?};
\citet{wang2022interpretability} discovered the circuit in GPT-2 in the task of indirect object identification.
The two key questions in the circuit include: (1) \emph{Localization: Which LM components are responsible for this behavior?} and (2) \emph{Interpretation: How do these components implement the behavior?} Answering these two questions provides an algorithmic-level explanation of how the LM implements the observed behavior. Additionally, it is crucial to focus on behaviors where the LM performs with high accuracy or exhibits the capability clearly, as we can only investigate mechanisms that are already reliably present in the model. In Section~\ref{sec: discussion-future-works}, we will provide a more careful discussion about the feasibility of circuit studies.

\paragraph{Interpreting an LM Component} This second category of circuit study aims to develop a thorough understanding of a specific LM component (e.g., attention head or FF sub-layer). The goal is often to develop a general understanding of the LM component that is independent of any specific task, making it essential for its interpretation to remain consistent across different tasks (Section~\ref{subsec:roadmap-universality}).
% The goal is often to obtain a generic understanding of the LM component agnostic to the specific task it is applied to, making the universality (Section~\ref{subsec:roadmap-universality}) of its interpretation across tasks a typical necessity \abu{we should rephrase; i don't know what is meant by ``typical necessity''}.
% The goal is to understand the component's role not just in a single task, but across a range of tasks.\zyc{"across a range of tasks" -- this is the ideal case, but is it fair to claim it as part of the goal?} \daking{Shouldn't the goal be always like that, for example, the goal of MI is to reverse engineer the LM but each MI paper only achieves the goal partially?}
The process of interpreting an LM component overlaps with the ``component interpretation'' step of interpreting an LM behavior; however, to obtain a comprehensive interpretation of an LM component, one often needs to iterate the interpretation process across multiple tasks, which goes beyond the scope of a particular behavior.

Considering the overlap in the procedures for the two sub-categories of circuit study, in what follows, we will mainly focus on presenting the workflow for interpreting an LM behavior.

\begin{table*}[t!]
    \centering
    \resizebox{\textwidth}{!}{%
    \begin{tabular}{
        >{\centering\arraybackslash}m{4.2cm}
        >{\raggedright\arraybackslash}m{7cm}
        >{\raggedright\arraybackslash}m{7cm}
    }
    \toprule
    \textbf{Techniques} & \textbf{Advantages} & \textbf{Limitations} \\
    \toprule

    \multicolumn{3}{c}{\textbf{Step 3: Localization}} \\
    \midrule

    Zero, mean, and resampling ablation (Node/Edge localization) &
    Faithful localization  &
    Computationally expensive and inefficient \\
    \midrule
    Path patching (Edge localization) &
    Faithful localization &
    Computationally expensive and inefficient \\
    \midrule

    ACDC (Auto Circuit DisCovery) \cite{conmy2023towards} &
    Faithful localization; Computationally efficient &
    Computationally expensive \\
    \midrule

    EAP~\cite{syed2023attribution}  &
    Low computational requirements  &
    Less faithful than ACDC and activation patching \\
    \midrule

    EAP-IG~\cite{hanna2024have} &
    Faithful localization; Low computational requirements  &
    N/A \\
    \midrule

    \multicolumn{3}{c}{\textbf{Step 4.1: Interpretation (Hypothesis Generation)}} \\
    \midrule

    Visualization &
    Intuitive; useful for rapid hypothesis generation about the function of the LM component &
    Qualitative and easy to over-interpret; require follow-up causal tests \\
    \midrule

    Vocabulary Projection Methods &
    Provide important signals for what output tokens are promoted or suppressed by an LM component or activation &
    Only decode features that can be represented in vocabulary space; early-layer projections can be unreliable without translators. \\
    \midrule

    \multicolumn{3}{c}{\textbf{Step 4.2: Interpretation (Hypothesis Validation)}} \\
    \midrule

    Vocabulary Projection Methods &
    Quick verification on the effect of the LM component or activation on LM output &
    Unreliable on early layers \\
    \midrule

    Intervention-based Methods &
    \emph{Causal} evidence of necessity/sufficiency; quantify edge/node contributions. &
    Out-of-distribution risks from corruption; compute-heavy at scale; require careful metric and corruption design \\
    \midrule

    Probing &
    Standard practice &
    Require training data; correlational evidence only; detect feature presence but not usage \\
    \bottomrule
    \end{tabular}
    }
    \caption{Comparison of techniques for \emph{circuit study} across Step 3 (Localization), Step 4.1 (Interpretation—Hypothesis Generation), and Step 4.2 (Interpretation—Hypothesis Validation).}
    \label{tab:circuit-study-comparison}
\end{table*}

\subsubsection{General Workflow for Circuit Study} \label{subsec: circuit-workflow}
A circuit study typically involves the following steps: \textbf{(1) Choose an LM behavior and a dataset:} Select the LM behavior for which we are searching the circuit. \textbf{(2) Define the computational graph:} Describe the nodes and edges of the computational graph. \textbf{(3) Localization:} Identify all the important nodes and edges connecting them. \textbf{(4) Interpretation:} Explain the role of all nodes and edges in implementing the LM behavior. \textbf{(5) Evaluation:} Evaluate the faithfulness of the discovered circuit. {Various MI techniques are used for conducting each step; we summarize their limitation and advantages in Table~\ref{tab:circuit-study-comparison}.}

\paragraph{Choose an LM Behavior and a Dataset} The first step in circuit discovery involves selecting a specific LM behavior and curating a dataset that showcases the behavior. The LM should have high task performance on the dataset, as this suggests the presence of internal mechanisms or a circuit that supports the behavior. In contrast, low task performance indicates that the mechanism may be poorly defined or absent, preventing further investigation. 

\begin{figure*}[t!]
  \centering
  % Row 1
  \begin{subfigure}[b]{0.49\textwidth}
    \centering
    \includegraphics[width=\textwidth]{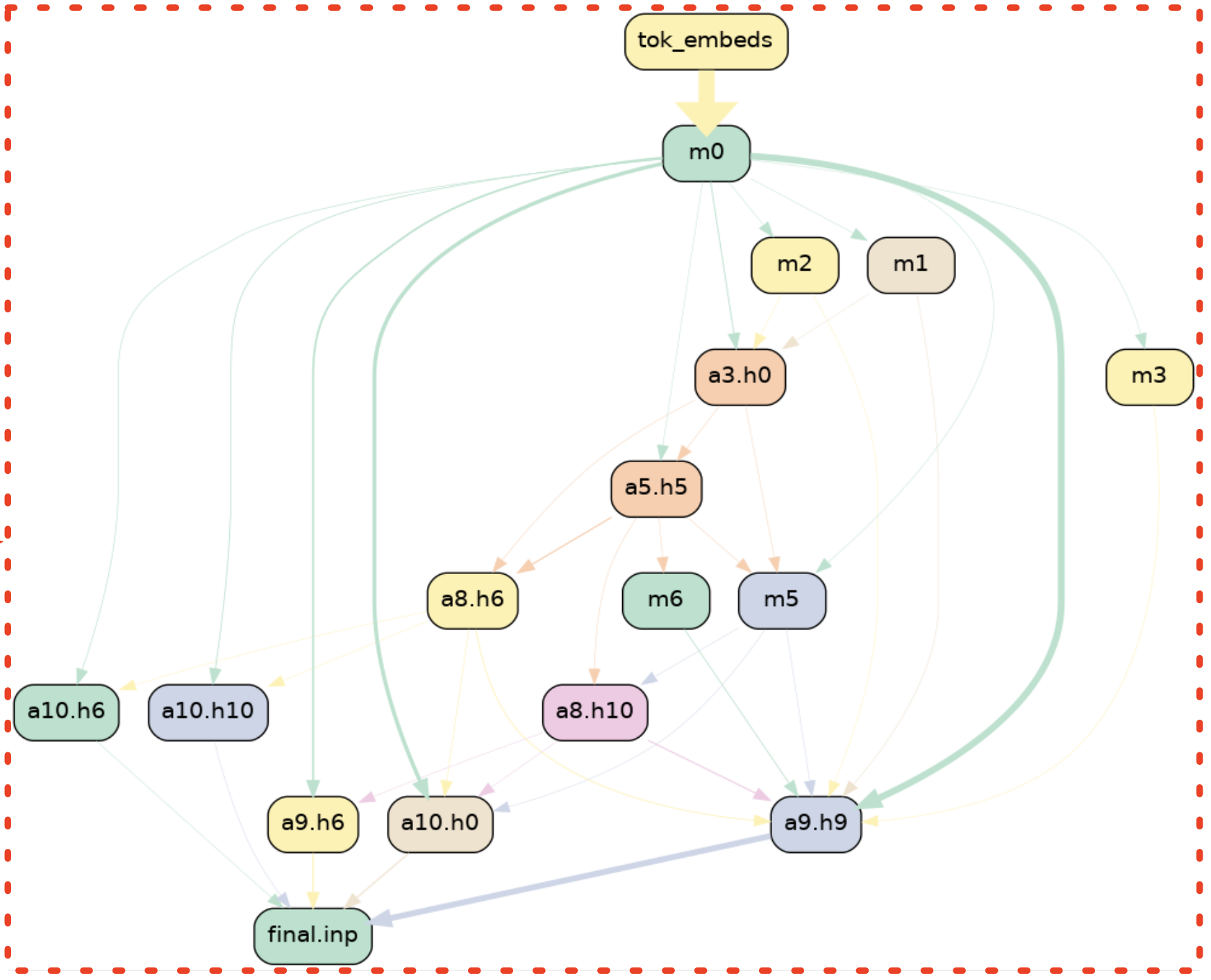}
    \caption{IOI circuit in GPT-2 Small discovered by \citet{conmy2023towards}.}
    % \label{fig:main-three-attn}
  \end{subfigure}
  \hfill
    \begin{subfigure}[b]{0.49\textwidth}
    \centering
    \includegraphics[width=\textwidth]{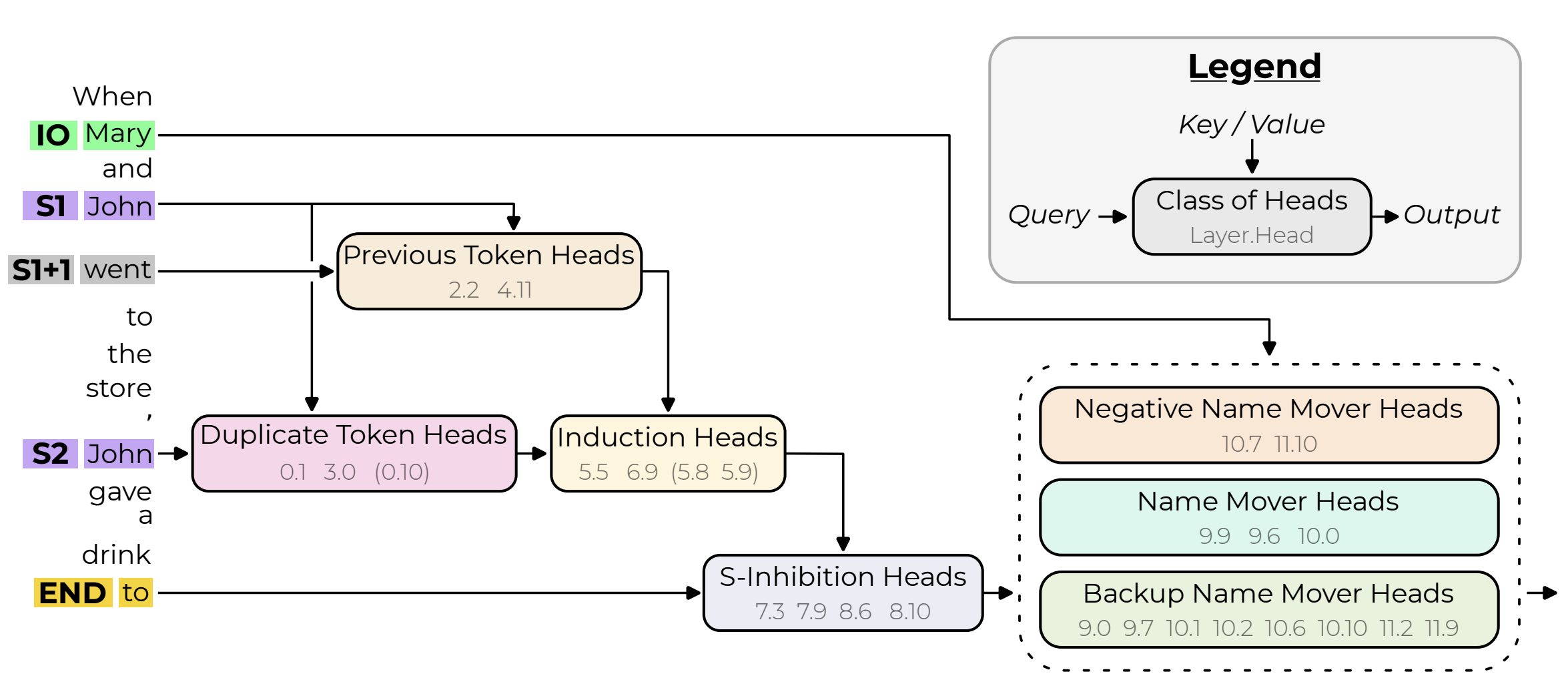}
    \caption{IOI circuit in GPT-2 Small discovered by \citet{wang2022interpretability}.}
    % \label{fig:main-three-circuit}
  \end{subfigure}  
    \caption{Example illustrating computational graphs defined with different abstractions for the same IOI circuit. (a) The IOI circuit, discovered by \citet{conmy2023towards}, defines a computational graph to include LM components \emph{independent from input positions} as nodes: $ml$ denotes the FF sublayer at layer $l$, $al.hj$ denotes the $j$-th attention head at layer $l$, and $final.inp$ refers to the logit score. One LM component (e.g., $a5.h5$) corresponds to only one node in the computational graph since they are position-independent. 
    % For nodes that are not directly connected across adjacent layers (e.g., $m0$ and $a9h9$), they are still considered connected in effect by writing to and reading from the RS. 
    (b) The IOI circuit, discovered by \citet{wang2022interpretability}, is represented as a computational graph where nodes corresponding to \emph{the same LM component at different input positions} are considered distinct nodes, highlighting that one component may assume different functional roles depending on its positional context. Only attention heads are considered in this circuit, where $l.j$ denotes the $j$-th attention head at layer $l$.}
  \label{fig: circuit-type}
\end{figure*}
% \begin{figure*}[t!]
%     \centering
%     \includegraphics[width=0.8\linewidth]{Figures/circuits-type.png}
%     \caption{Examples illustrating computational graphs of circuits defined at different levels of abstraction and granularity. (A) The IOI circuit, discovered by \citet{conmy2023towards} as a reproduction of \citet{wang2022interpretability}, defines a computational graph to include LM components \emph{independent from input positions} as nodes: $ml$ denotes the FF sublayer at layer $l$, $al.hj$ denotes the $j$-th attention head at layer $l$, and $final.inp$ refers to the logit score. Only one presence of each node is considered in the computational graph since they are position-independent. For nodes that are not directly connected across adjacent layers (e.g., $m0$ and $a9h9$), they are still considered connected in effect by writing to and reading from the RS. (B) The arithmetic circuit, discovered by \citet{nikankin2024arithmetic}, is represented as a computational graph where nodes corresponding to \emph{the same LM component at different input positions} are considered distinct nodes, capturing how a single component can serve distinct functional roles depending on positional context. 
%     % In addition, unlike the IOI circuit in (A), the graph does not model connections between LM components across physically non-adjacent layers.\zyc{double check if the last sent is correct.} 
%     }
%     \label{fig: circuit-type}
% \end{figure*}

\paragraph{Define the Computational Graph} 
{The second step of circuit discovery involves defining a computational graph of the model, which can be defined at different abstraction levels depending on the desired level of explanation detail for the model behavior. For example, Figure~\ref{fig: circuit-type} shows two types of definition: \citet{conmy2023towards} defined the computational graph such that a single component is represented by a single node across all token positions, whereas \citet{wang2022interpretability} treated the same component as different nodes, one for each decoding position. In addition, the definition of computational graphs can also vary in the granularity of their nodes. For example, \citet{hanna2024have} defined the GPT-2 model as a computational graph, where 144 attention heads (12 heads/layer $\times$ 12 layers) and the 12 FF sub-layers (1 per layer) in a GPT-2 model were considered nodes in its computational graph and discovered circuits as sub-graphs from it. In contrast, \citet{marks2024sparse} defined a computational graph of Pythia-70M~\cite{biderman2023pythia} and Gemma-2-2B~\cite{team2024gemma} model by considering the SAE features as nodes instead of attention heads, defining the graph at a more granular level.  Beyond node definitions, an additional consideration in computational graphs lies in how edges are determined. Most existing literature \cite{conmy2023towards, syed2023attribution, hanna2024does, hanna2024have}  considers the connections between LM components in \emph{non-adjacent} layers as valid edges in the LM's computational graph, due to the additive nature of the RS (Eq~\ref{eq: rs}). That is, even if two components (e.g., the FF sub-layers in Layer 2 and Layer 5) are not connected through direct computations across adjacent layers, they are still connected in effect by writing to and reading from the RS, as shown in Figure~\ref{fig: circuit-type} (a).}

\paragraph{Localization}
% \zyc{TODO: consider splitting this paragraph into "Localization of Nodes" and "Localization of Edges". For edge patching, four runs in implementation are needed -- can we include a picture to illustrate the practical procedure? Talk about the consideration of "position" -- discuss \cite{conmy2024towards, hanna2024does, wang2022interpretability}; I will check other papers similarly.}
Once the computational graph for the LM is defined, the next step is to identify all the important nodes and edges responsible for implementing the behavior. This process is termed ``localization''. Localization can be broken down into \emph{Localization of Nodes} and \emph{Localization of Edges}. \textbf{(1) Localization of nodes}: Intervention methods, such as zero ablation~\cite{olsson2022context} and mean ablation~\cite{wang2022interpretability}, are commonly used to measure the importance of individual nodes in the model. These methods involve intervening in a node's output and observing its impact on the model's behavior, with important nodes being those whose intervention results in a deviation above a certain threshold in model performance. \textbf{(2) Localization of edges}: Path patching~\cite{wang2022interpretability} is a commonly used intervention method to localize edges. Other intervention-based techniques, such as zero ablation, mean ablation, and resampling ablation, can also be applied for localization. However, these methods can be computationally-intensive
% \abu{do you mean ``computationally-intensive''? ``labor'' sounds like it requires human annotation effort} 
as we need to measure the importance of each edge individually.
% While both noising and denoising-based intervention methods can be used for localization, noising-based interventions, e.g., path patching~\cite{wang2022interpretability}, are more commonly used\zyc{any reason? then when should denoising-based methods be used?}. 
% At a high level, the noising-based intervention methods, such as zero ablation~\cite{} and mean ablation~\cite{wang2022interpretability}, can be used to measure the importance of all nodes of the model, where only the important nodes are considered as the nodes of the circuit. Similarly, noising-based intervention methods, such as path patching~\cite{wang2022interpretability} can be used to identify the edges of the circuit. 
% This intervention-based method can be manually intensive as we need to measure the importance of each node and edge of the model individually. 
% To localize all the important nodes and edges, one will have to apply the intervention to every node in the LM, which is labor-intensive.
To address this issue, automated localization techniques such as ACDC~\cite{conmy2023towards}, EAP~\cite{syed2023attribution}, and EAP-IG~\cite{hanna2024have} have been proposed, as discussed in Section~\ref{subsec:intervention-tech-advancements}.

\paragraph{Interpretation}
Once all the important LM components and edges have been identified during the localization step, we interpret the functional role of each component. This {interpretation} step can be further divided into two sub-steps -- \textbf{(a) Generating a hypothesis:} The first step involves generating a hypothesis about the function of each LM component. This process is often guided by analyzing the behavior of the component under relevant inputs using various techniques, such as visualization (e.g., attention visualization) and vocabulary projection methods. A human can then generate a plausible hypothesis based on the analysis (e.g., attention head $H$ was observed to decrease the logit of the token it attends to and thus is hypothesized to function as a copy suppression head). 
% \zyc{What does the following mean? "a more principled"?
% Additionally, techniques such as causal mediation analysis and interchange interventions offer a more principled approach to formalizing hypotheses. 
% }
\textbf{(b) Validating the hypothesis:} The hypothesis can be validated using various techniques, including vocabulary projection, intervention-based methods, and probing. For instance, if the hypothesis suggests that attention head $H$ is a copy suppression head, then knocking out (or noising) attention head $H$ should remove the suppression effect. If the hypothesis is validated, the model component is considered interpreted; otherwise, further analysis and new hypotheses are needed. 

\paragraph{Evaluation} The discovered circuits are evaluated for three criteria -- faithfulness, minimality, and completeness.
\textit{Faithfulness} is evaluated by comparing the full model vs. the partial one with the localized circuit alone \cite{olsson2022context, wang2022interpretability, marks2024sparse}; the circuit is considered faithful when the performance gap between the full and the partial LM is small. On the other hand, \textit{minimality} measures whether all LM components in the circuits are necessary, often by randomly ablating components in the circuit and computing the change in the LM behavior \citep{wang2022interpretability}. A significant change in behavior would indicate that all parts of the circuit are necessary and the explanation is minimal. Minimality is also referred to as \textit{sparsity}~\cite{bhaskar2024finding}. Finally, \textit{completeness} checks if the circuit includes all the LM components used by the LM for exhibiting the behavior of interest~\cite{wang2022interpretability, marks2024sparse}. Specifically, \citet{wang2022interpretability} measures the completeness by comparing the full model vs identified circuit behavior under random ablations of the circuit components. If the circuit is complete, LM behavior should remain similar to the whole model even under random ablations.

% \paragraph{Interpreting LM component} The second category of the circuit study, i.e., \textbf{interpreting an LM component}, follows similar steps as interpreting an LM behavior, but starts with the second step of the latter, given that the target component to interpret has been provided. 

% Next, $h_{name}^0$ was multiplied by the OV matrix of the $a_9^{9}$ and multiplied by the unembedding matrix, and the final layer norm was applied to obtain logit probabilities. Finally, a copy scores metric was proposed which is the proportion of samples ($N=1000$) that contain the input name token in the top 5 logits. $a_9^{9}$ had a copy score above 95\% which provides strong evidence that $a_9^{9}$ is name mover head. 

% \input{Roadmap/interpret-lm-behavior}
% \input{Roadmap/interpret-lm-component}
% \input{Roadmap/circuit-study-eval}

% \input{Roadmap/circuit-study-findings}

\subsection{Study of Universality}\label{subsec:roadmap-universality}

\begin{figure*}[t]
    \centering
    \includegraphics[width=\linewidth]{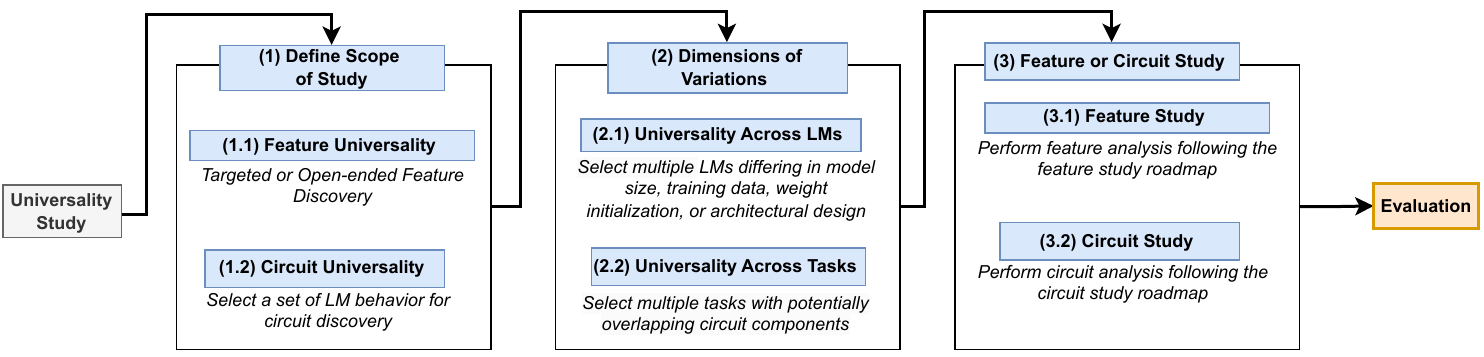}
    \caption{A task-centric beginner's roadmap for the study of universality.}
    % \vspace{-3mm}
    \label{fig: universality-roadmap}
\end{figure*}

The study of universality (Figure~\ref{fig: universality-roadmap}) investigates whether similar features and circuits can be found across different LMs or tasks. To this end, the study of universality involves investigating the internal mechanisms of several LMs and tasks with a primary research question as -- \emph{``Do the same features and circuits exist across LMs and tasks?''}

\subsubsection{General Workflow for Universality Study}
The general workflow for universality consists of the following steps. 
\paragraph{Scope of Universality} The study of universality can be mainly studied under two dimensions -- (1) \emph{Universality of features:} the degree of similarity of the features contained within representations across different LMs; (2) \emph{Universality of circuits:} whether a particular LM behavior is implemented using similar circuits across LMs, and whether individual model components perform the same specialized functions (e.g., induction heads [\citealt{olsson2022context}], successor heads [\citealt{gould2023successor}]) across multiple LMs and tasks. The universality of features can be studied in both the targeted and open-ended feature study settings. On the other hand, the study of circuit universality only makes sense when a particular LM behavior is consistently observed across multiple LMs and/or tasks, as the goal is to examine whether various LMs or a single LM in various tasks implement the same behavior using similar circuits.

\paragraph{Dimension of Variations} 
To measure \textbf{universality across LMs}, it is essential to analyze multiple LMs that differ in aspects such as model size, training data, weight initialization, or architectural design.
% \zyc{do people have to pick one dimension? does it make sense to randomly pick two LMs (say GPT vs. llama)?} \daking{They don't necessarily have to pick one dimension. For instance, if they randomly pick GPT vs Llama, they differ in multiple aspects such as model size, training data, architectural design, and so on}. 
For instance, \citet{gurnee2024universal} trained five GPT2-Small and GPT2-Medium models from different random seeds to study the universality of neurons.
In the case of studying \textbf{universality across tasks}, the dimension of variation can be the task choice and the specific task setting (e.g., data distribution). For instance, \citet{merullo2023circuit} studied two different tasks, an indirect object identification task and a colored objects task, to determine whether certain LM components are reused across tasks while maintaining the same function.

\paragraph{Feature and Circuit Study} The feature and circuit study can then be performed across LMs and/or tasks, as discussed in Sections~\ref{subsec: feature-workflow} and \ref{subsec: circuit-workflow}. 

\paragraph{Evaluation of Feature Universality}  
{To measure similarity between features across different models, a common approach involves running each model on a common dataset (e.g., Pile test set~\citep{pile}) and recording the activation patterns of individual features (i.e., neurons or SAE features) across all tokens in the dataset~\cite{gurnee2024universal}. These activation patterns are treated as activation vectors representing each feature's behavior. Similarity is then quantified by computing the Pearson correlation (e.g., Pairwise Pearson Correlation) between activation vectors from different models. For each feature in one model, the most similar counterpart in another model is identified by selecting the feature with the highest correlation. This process can be repeated across all features to construct a distribution of similarity scores. To account for sources of variation such as random initialization, architecture, or training noise, comparisons are often made against appropriate baselines or control settings. This method offers a principled way to identify and analyze features that are universal across models.}

\paragraph{Evaluation of Circuit Universality}
{To determine whether a similar circuit is present across different models, we assess whether the models implement the same underlying algorithm for the behavior under study. Specifically, a circuit can be interpreted as an algorithm that is implemented by the components within the circuit. By interpreting the functional roles of these components, we can identify whether components with similar roles are present across models. Circuit universality is then evaluated by examining the overlap in functionally equivalent components across models under study~\cite{merullo2023circuit}.}

% \paragraph{Evaluation} Finally, we compare the similarities and differences of the discovered circuits and features across different LMs and/or tasks. Specifically, the features and circuits that are found across various LMs are considered to be universal, while those unique to specific LMs are considered model-specific. Likewise, one can draw conclusions about task-agnostic vs. task-specific circuits.

It is important to note that the question of ``do similar features and circuits exist across LMs and tasks'' may not be binary; the experiments in practice instead reflect the degree to which universality holds. For instance, \citet{gurnee2023finding} found that only 1-5\% of the neurons were universal across five different GPT2-Small and GPT2-Medium models. 

% \zyc{I feel the argument about "binary question" is ambiguous. In your example, to each neuron, it is either universal or not universal across LMs, which clearly has a binary answer to the question. I guess that you were referring to the question of ``do similar features and circuits exist across LMs and tasks''; to this question, yes the answer can be in a percentage. I've tried to clarify it.}

\section{Case Studies of Beginner's Roadmap}\label{sec: case-studies}
We present case studies for research work in MI, detailing how the high-level workflows in our roadmap were employed in papers that conducted these analyses. These case studies serve as \emph{concrete examples} to help readers new to the field effectively map research questions to specific research activities outlined in our roadmap.
% and the insights gained from the experiments. 

\subsection{Case Study for Feature Study} 

\subsubsection{Targeted Feature Study with Probing}
\citet{gurnee2023finding} proposed and employed a probing technique, \emph{sparse probing}, to conduct targeted feature discovery. 
% for 100 unique features across 7 different LMs. 
\textbf{(1) Hypothesis Generation:} \citet{gurnee2023finding} sought to discover 100 pre-defined features across seven LMs of varying sizes.
Furthermore, they also aimed to localize features to a specific neuron or sets of neurons within the activations. For instance, to determine whether an LM encodes the ``is\_french\_text'' feature in the FF activations in the first layer ($f_i^1$'s) of the LM, the following steps were taken for \textbf{(2) Verifying the Hypothesis}: \textbf{(2.1) Dataset preparation:} A labeled dataset consisting of a train and test set was created for training and evaluating the probe. This dataset included examples with the ``is\_french\_text'' feature (i.e., French text) and examples without this feature. \textbf{(2.2) Activation Extraction:} The corresponding $f_i^1$ activations for each example in the dataset were obtained when they were used as input text of the LM. Whenever the examples consisted of multiple tokens, elementwise aggregation (e.g., mean or max) of the activations over the token
span were considered. \textbf{(2.3) Training a Sparse Probe:} A sparse probe was trained using the extracted activation $a_i^1$ and their corresponding labels in the train set. To localize the feature to a specific set of neurons, the activations were constrained to have at most 
$k$-non-zero coefficients. To determine which neurons should have non-zero coefficients, a technique called \emph{adaptive thresholding} was employed which involves training a series of classifiers that iteratively decrease the value of $k$. At each step, the probe is retrained to use only the top-$k$ neurons with the highest coefficient magnitudes from the previous step. \textbf{(3) Evaluation:} Finally, the trained sparse probe was evaluated using F1 score, where if the probe achieves a high F1 score in the test set, it suggests that the neurons with non-zero coefficients have ``is\_french\_text'' feature encoded within them. Furthermore, they also measure the precision and recall to get more insight into the granularity of features encoded by the set of neurons. For instance, high precision and low recall of the probing classifier may indicate that the identified subset of neurons may encode
a more specific feature than the feature being probed for (e.g. ``is\_french\_noun'' instead of ``is\_french\_text''). In their work, \citet{gurnee2023finding} repeated Steps 2-3 multiple times to determine whether activations at other layers also encode the ``is\_french\_text'' feature, and applied the same to multiple LMs to study the universality of this feature.
% \textbf{(5) Iterating Across Other Activations:} Steps 2-4 were then repeated to determine if other activations, aside from $a_i^1$, encode the ``is\_french\_text'' feature. \textbf{(6) Discovering other features across LMs:} The entire process (Steps 1–5) was repeated to localize other features across other LMs.

\subsubsection{Open-ended Feature Study with SAEs} \citet{bricken2023monosemanticity} conducted open-ended feature discovery on a one-layer toy LM using SAE, visualization, and human and automatic interpretation techniques. The following steps were taken to discover features encoded in the LM activations: \textbf{(1) Observe:} SAEs were first trained on the activations obtained by feeding the LM sentences from its training set as input. After training, a higher-dimensional, sparse activation was obtained for each LM activation. The next step was to accurately label each element of the sparse activations, referred to as \emph{features} by \citet{bricken2023monosemanticity}, with a correct description. For instance, a feature might activate in response to phrases related to music. A user interface was then developed for each element of the sparse activations to provide useful information for accurate labeling. This included text examples of when each sparse feature activates, the effects they have on the logits when active, examples of how they influence token probabilities if the feature is ablated, and other relevant details. All of these observations provide the intermediate explanations of each activation. \textbf{(2-3) Explain and Evaluation:} Human evaluators were tasked with labeling the features using all the information provided in the interface, guided by a scoring rubric that included instructions such as ``On a scale of 0–3, rate your confidence in this interpretation'' for evaluating the interpretability of the feature discovery. Additionally, automatic explanations~\cite{bills2023language} were also used.

% \textbf{(1) Choose LM behavior and Dataset:} Select the LM behavior for which we are searching the circuit. \textbf{(2) Define the Computational Graph:} Describe the nodes and edges of the computational graph. \textbf{(3) Localization:} Identify all the important nodes and edges connecting them. \textbf{(4) Interpretation:} Explain the role of all nodes and edges in implementing the LM behavior. \textbf{(5) Evaluation:} Evaluating the faithfulness of the discovered circuit.

\subsection{Case Study for Circuit Study} \citet{wang2022interpretability} conducted one of the first circuit analyses based on GPT-2 Small, following these steps. \textbf{(1) Choose an LM Behavior and a Dataset:} The study examined a circuit in GPT-2 Small responsible for solving the indirect object identification (IOI) task: Given a sentence such as ``When Mary and John went to the store, John gave a drink to'', complete the sentence with ``Mary'' (IO) and not ``John'' (S). To investigate the behavior, a dataset was synthesized to showcase the IOI task. \textbf{(2) Define the Computational Graph:} The study defined the computational graph with attention heads as nodes and connections between them as edges. Notably, components such as FFs, LayerNorms, and embedding matrices were excluded from the analysis, allowing the investigation to focus solely on understanding the attention heads. In addition, the study considered the same attention heads at different token positions as distinct components; as a result, the computational graph to be discovered is \emph{position dependent}. \textbf{(3) Localization:} Using the synthesized dataset, path patching was applied to identify the important nodes and edges in the circuit. Specifically, to determine whether an attention head $a_i^l$ is important, path patching for the pathway $a_i^l \rightarrow logits$ was applied and the effect on the logit difference between the direct (John) and indirect (Mary) object was measured, where a high logit difference indicates that the $a_i^l$ plays a crucial role in the IOI task. After performing path patching across all attention heads in GPT-2 Small, 26 attention heads were identified as important nodes for the IOI circuit. \textbf{(4) Interpretation:} This step involves investigating the role of each of the 26 important attention heads identified in the previous step. For instance, to understand the role of $a_9^{9}$, the following two substeps were performed. \textbf{(4.1) Generating a hypothesis:} To generate a plausible hypothesis on the role of $a_9^{9}$, attention visualization was analyzed, which showed that $a_9^{9}$ strongly attends to the indirect object token (e.g., ``Mary'').
% attention visualization (e.g., tokens attended by $a_9^{9}$) was analyzed. The visualization showed $a_9^{9}$ strongly attends to the IO token.
Based on this observation, a hypothesis was formed: $a_9^{9}$ is a \emph{name mover head} that attends to the correct name and copies whatever it attends to.\textbf{(4.2) Validating the hypothesis:} 
% \zyc{The following description is misleading. (1) The single hypothesis actually contains two parts: "attends to the correct name" and "copies whatever it attends to". In the IOI paper, the authors validated these two sub-hypotheses separately. What you described was about the second sub-hypothesis. (2) The description for validating the "copy" sub-hypothesis is also imprecise. The authors applied the logit lens to $h_{name}^1 W_V^{9,9}W_O^{9}$ (i.e., $h_{name}^1$ multiplied by the OV matrix of the head $a_9^9$), rather than $h_{name}^1$. The projection is to isolate the two sub-hypotheses and to confirm that when the attention is perfect (hence $h_{name}^1$ for the name in consideration), whether there's a copy effect observed from the output of the head.}
{They validated the hypothesis that attention head $a_9^{9}$ is a \emph{name mover head} by demonstrating that $a_9^{9}$ (1) attends to the correct name and (2) copies whatever name it attends to. First, they analyzed attention patterns and found that these heads strongly focused on the indirect object (IO) token, with higher attention correlating ($\rho > 0.81$) with increased logit output in the direction of the correct name token (IO token). To further confirm their copying behavior, they studied what values were written via the OV matrix of $a_9^{9}$ by simulating the perfect attention score of $a_9^{9}$ to the IO token and measuring the resulting logit probabilities. Specifically, they applied the logit lens to $h_{name}^1 W_V^{9,9}W_O^{9}$ (i.e., $h_{name}^1$ multiplied by the OV matrix of the head $a_9^9$) to examine what values these heads wrote into the RS at the position of each name token. The ``copy score'' that quantified how often the input name appeared in the top 5 logits, exceeded 95\% for Name Mover Heads, compared to less than 20\% for average heads. These results confirm that Name Mover Heads play a crucial role in copying and transferring name information through the model’s computation.}
% Logit lens was employed to validate a hypothesis that $a_9^{9}$ is a \emph{name mover head}. Specifically, they applied the logit lens to $h_{name}^1 W_V^{9,9}W_O^{9}$ (i.e., $h_{name}^1$ multiplied by the OV matrix of the head $a_9^9$). The projection is to isolate the two sub-hypotheses and to confirm that when the attention is perfect (hence $h_{name}^1$ for the name in consideration), whether there's a copy effect observed from the output of the head. Specifically, they validate the hypothesis by checking if tokens with the top-5 largest logits include the name token. In \citet{wang2022interpretability}'s experiments, for over 95\% of test samples, the name token appeared among the top-5 tokens, which provided strong evidence that $a_9^{9}$ was a name mover head. 
\textbf{(5) Evaluation:} The study assessed the discovered circuit using three evaluation metrics—faithfulness, minimality, and completeness—as discussed in Section~\ref{subsec: circuit-workflow}.

\section{Findings and Applications} \label{sec: findings-application}
\subsection{Feature Study} \label{findings: feature}

\paragraph{Monosemantic vs. Polysemantic Neurons} \label{para: polysemantics}
\begin{table*}[!t]
\centering\resizebox{0.9\textwidth}{!}{
\begin{tabular}{
    >{\arraybackslash}p{2.5cm}
    >{\arraybackslash}p{12.5cm}
}
\toprule
\textbf{Neurons} & \textbf{Description} \\
\midrule
Knowledge neurons~\cite{dai2021knowledge} & FF neurons that store a relational fact (e.g., ``The capital of Ireland is -> Dublin''); increasing or decreasing their activations directly affects whether the model recalls or suppresses the associated fact. \\
\midrule
Skill neurons~\cite{wang2022finding} &  A small set of FF neurons whose activation patterns are highly predictive of whether the model is performing a specific task (e.g., sentiment analysis); ablation of these neurons leads to a decrease in task performance. \\
\midrule
Language-specific neurons~\cite{tang-etal-2024-language} &  Neurons that activate much more strongly for one or two languages compared to others; ablation of these neurons leads to a sharp drop in the model’s ability to understand and generate text in the language they encode. \\
\midrule
Confidence regulation neurons~\cite{stolfo2024confidence} & Neurons in the final layer of LMs that adjust the model’s confidence (entropy of predictions) on what they are predicting, but don't directly change which token is predicted.  \\
\midrule
Arithmetic neurons~\cite{rai2024investigation} & Neurons that are crucial for arithmetic operations; ablating these neurons leads to a significant decrease in the performance of the arithmetic task. \\
\midrule
Positional neurons~\cite{voita2024neurons} & Neurons that activate for specific token position ranges, i.e., they always activate for tokens within certain position ranges regardless of their meaning. \\
\midrule
Dead neurons~\cite{voita2024neurons} & Neurons that never activate across a wide and diverse set of inputs, making them essentially unused by the model.  \\
\midrule
Safety-specific neurons~\cite{zhao2025understanding} & A small specialized set of neurons, found mainly in MHA layers, that are dedicated to recognizing and blocking harmful prompts (e.g., ``how to build a bomb?'').\\

\bottomrule
\end{tabular}
}
\caption{A list of neurons identified in LMs, each associated with specific features or roles, along with their descriptions.}
\label{tab:neuron-list}
\end{table*}

{Earlier work studied neurons as a natural candidate for encoding features, where a high value at a particular position would indicate the presence of that feature. This led to the discovery of several interesting neurons such as \emph{sentiment neurons}~\cite{tigges2023linear}, \emph{knowledge neurons}~\cite{dai2021knowledge},  \emph{skill neurons}~\cite{wang2022finding}, and \emph{positional neurons}~\cite{voita2023neurons}, {among others, as summarized in Table~\ref{tab:neuron-list}.} The question of whether neurons encode specific features has been explored extensively, not only in the context of transformer-based LMs but also in earlier work on word embeddings~\cite{mikolov2013distributed}, recurrent neural networks \cite[RNNs;][]{radford2017learning}, and vision models~\cite{cammarata2020curve}, predating the field of MI itself. However, many of these studies found that neurons within LMs are not \emph{monosemantic}, i.e., they do not activate only for a single feature. In contrast, they are \emph{polysemantic}, i.e., they activate in response to multiple unrelated features~\cite{elhage2022solu, gurnee2023finding, elhage2022superposition}. For instance, a single neuron could activate for both French texts and texts encoded in Base64. This polysemantic nature of neurons shows that the initial hypothesis that each neuron encodes a single feature is not always true.}

\paragraph{Superposition}
{The discovery of polysemantic neurons has led to the hypothesis of \emph{superposition}, i.e., a model can represent a greater number of independent features than the number of available neurons~\cite{elhage2022superposition}. In this framework, features are encoded as directions formed by a linear combination of multiple neurons rather than being tied to a single neuron. Although superposition enables the encoding of a larger number of features, it also introduces potential interference among overlapping representations, where activating one feature may unintentionally activate others that share the same representation space. Empirical findings suggest that the natural sparsity of feature activations helps mitigate this issue of interference~\cite{elhage2022superposition}. Since only a small subset of features is typically active at any given time, the risk of unintended activations is reduced, allowing the neural model to maintain distinguishable representations despite the constraints imposed by superposition. Furthermore, models appear to prioritize encoding the most important features in a more monosemantic fashion, while less critical features are either distributed across multiple neurons or omitted altogether, thereby minimizing performance degradation caused by interference~\cite{scherlis2022polysemanticity}.  Beyond its role in feature representation, superposition has also been observed to aid in data memorization for smaller datasets and in learning more generalized features for larger ones, with a notable transition between these regimes marked by the phenomenon of double descent~\cite{henighan2023superposition}.}

% Superposition has been empirically studied in various toy settings. For instance, \citet{elhage2022superposition} introduced the first examples of toy models that utilized superposition to achieve low loss. They found that superposition is feasible because not all features are active simultaneously i.e. only a small subset of features is active at any given time. This sparsity in feature activations helps minimize interference caused by superposition. In other words, superposition is possible because not all features are active at any one time. Similarly, \citet{henighan2023superposition} further explored the role of superposition while overfitting and found that superposition are being used by models not only to encode features but also to memorize data points. Specifically, they found that superposition are being used to memorize data points when datasets are small or to learn more general features when datasets are large, with a transition between these two regimes that is marked by double descent. Similarly, \citet{scherlis2022polysemanticity} demonstrate that models tend to encode the most important features monosemantically, represent less but equally important features polysemantically, and disregard the least important features altogether. Finally, \citet{hanni2024mathematical} provides a theoretical foundation for understanding how neural networks can perform computation in superposition.}

\paragraph{Tackling Superposition with SAEs}
{Recently, SAEs have gained popularity as a method for interpreting representations that encode features in superposition~\cite{bricken2023monosemanticity, riggs2023sae, cunningham2023sparse, lieberum2024gemma, marks2024enhancing, he2024llama}. Various SAE variants with different architectures have been proposed, as discussed in Section~\ref{sec: sae}. Early studies on SAEs have shown promising results, with the extracted features from SAEs appearing more interpretable than those from neurons, according to both human analysis and automatic explanation scores \cite{bricken2023monosemanticity, makelov2024sparse}. Despite these promising findings, several challenges persist. For instance, SAEs do not perfectly reconstruct the activation which is often attributed to the trade-off between sparsity and reconstruction loss, raising concerns regarding the faithfulness of the extracted representations~\cite{gurnee2024sae, muhamed2024decoding, chanin2024absorption, anders_bloom_2024_gpt2saeacts}. For instance, \citet{templeton2024scaling} speculated that SAEs may have only captured a fraction of features encoded in model activations, with many rare or highly specific concepts remaining undetected due to their infrequent activation. Similarly, recent studies~\cite{braun2024identifying, makelov2024towards} have also raised concerns regarding the functional usefulness of the extracted features, i.e., whether these features are actively used by the model for inference or merely reflect patterns present in the training data used to train the SAE. Although SAEs with improved architectures have been proposed to mitigate some of these limitations~\cite{muhamed2024decoding, braun2024identifying}, further advancements in SAEs are needed to ensure that the extracted features are both representative of the model’s underlying computation and practically useful for interpretability.}

\paragraph{Linear Representation Hypothesis}
 {Most MI studies rely on the \emph{linear representation hypothesis}~\cite{mikolov2013distributed, pennington2014glove} for performing feature studies. The hypothesis states that \emph{features are encoded linearly in the representation space of the model}. For example, SAEs adopt this hypothesis by assuming that activations can be expressed as sparse linear combinations of feature vectors, while probing assumes that a linear model can detect the presence or absence of a feature in activations. Specifically, the linear representation hypothesis posits that neural networks have two properties: \emph{linearity}, i.e., the network's activation space consists of meaningful (linear) vectors, each representing a feature, and \emph{decomposability}, i.e., network activations can be decomposed and described in terms of these independent features. Notably, this hypothesis predates MI and has been extensively studied to understand how high-level concepts are stored in word embeddings~\cite{mikolov2013distributed}, latent representations of variational autoencoders~\cite{wang2024concept}, and LM representations~\cite{dai2021knowledge, radford2017learning}. A well-known example of the hypothesis is the presence of semantic relationships such as $\overrightarrow{man\vphantom{k}} - \overrightarrow{woman\vphantom{k}} \approx \overrightarrow{king} - \overrightarrow{queen\vphantom{k}}$ within word embeddings, providing evidence that high-level concepts (e.g., gender) are represented as specific directions in the activation space. An increasing body of work has found evidence supporting the hypothesis by discovering features encoded as directions within LM representations, such as sentiment~\cite{tigges2023linear}, space and time~\cite{gurnee2023language}, language~\cite{bricken2023monosemanticity}, truth~\cite{marks2023geometry}, and refusal~\cite{arditi2024refusal}. 
 {However, some studies have also found the existence of non-linear features that contradict the linear representation hypothesis. For instance, \citet{engels2024not} discovered multi-dimensional features, such as circular representations for days of the week and months of the year, demonstrating that some features are inherently irreducible to a single dimension. This finding contradicts the previous notion of linear representation hypothesis where each feature is represented by one-dimensional directions. Subsequently, \citet{olah2024multidim} pointed out that since multi-dimensional features remain mathematically linear, existing techniques should be capable of detecting them with a slight change. However, \citet{csordas2024recurrent} found that gated recurrent neural networks (RNNs) encode sequential information using non-linear ``onion representations'', where tokens are represented by magnitude rather than direction, leading to layered features that do not reside in distinct linear subspaces. Given these contradictory findings, further research is crucial to either validate or refute the linear representation hypothesis, as it forms the foundation of much feature analysis.}

\subsection{Circuit Study}
\label{subsec: findings-circuits}
\subsubsection{Interpreting LM Behaviors} \label{para: LM behaviors} 
% \zyc{can we get back the v1 paragraph summarizing existing circuits?}
{Circuit studies have identified a range of behaviors across LMs of different sizes, as listed in Table~\ref{tab: Circuits}. These circuits are defined at varying levels of granularity, where nodes in the computation graph are defined as the outputs of MHA and FF sublayers \citep{olsson2022context, wang2022interpretability} or as SAE features \cite{marks2024sparse, cunningham2023sparse, kissane2024interpreting}. In addition, \citet{merullo2023circuit} and \citet{quirke2024increasing} showed that the same components (e.g., induction heads) are reused by different circuits (e.g., IOI and induction circuits) to implement different tasks, demonstrating the generalizability of interpreted components. Beyond circuit discovery, some studies~\cite{heimersheim2023circuit, marks2024sparse} have shown that the identified circuits can be enhanced for better performance, highlighting the potential for practical downstream applications. For instance, \citet{marks2024sparse} proposed a technique, Spurious Human-interpretable Feature Trimming (SHIFT), which improves the generalization of a classifier by removing the spurious features in the discovered circuit. This approach paves the way for new possibilities in human-AI collaboration.  

% \zyc{The following discussion seems to be more about "approaches" rather than "findings and applications"}
However, these studies have also revealed new challenges in circuit discovery. For instance, circuit discovery methods were found to generalize poorly to adversarial examples, indicating that existing approaches are sensitive to the datasets used in circuit identification and evaluation
% issue\abu{``issue'' is a typo? delete it?}
~\cite{uit2024adversarial}. In addition, some studies~\cite{zhong2024clock, nanda2023progress} have also shown that LMs implement different algorithms in tandem to solve the same task and the interactions between these parallel algorithms are highly sensitive to training and architectural hyperparameters. Finally, scalability remains a concern for larger LMs. To demonstrate the scalability of current techniques for identifying circuits in LLMs, \citet{lieberum2023does} identified the circuit used for the multiple-choice question-answering task on the 70B Chinchilla LLM \cite{hoffmann2022training}.}

\begin{table*}[t!]
    \centering
    \resizebox{\textwidth}{!}{%
    \begin{tabular}{>{\centering\arraybackslash}m{4.0cm}
    m{7.6cm}>{\centering\arraybackslash}m{5cm}>{\centering\arraybackslash}m{2.4cm}}
    \toprule
    \textbf{Circuits} & \textbf{Circuit Descriptions} & \textbf{Circuit Nodes} & \textbf{LMs}  \\
    \toprule
    Induction \par \cite{elhage2021mathematical} & Completes sentences like ``Mr D urs ley was thin and bold. Mr D'' with ``urs'' & 2 attention heads & 2-layer toy LM, GPT2-Small \\
    \midrule

    Indirect Object Identification \par \cite{wang2022interpretability} & Completes sentences like ``When John and Mary went to the store, John gave a drink to'' with ``Mary'' as opposed to ``John'' & 26 attention heads & GPT2-Small \\
    \midrule

    Docstring \par \cite{heimersheim2023circuit} & Predicts argument names in the docstring:\par
    % def port(self, load, size, files):
    %     ''' Oil column piece
    %     :param load: crime population
    %     :param size: unit dark
    %     :param
\begin{lstlisting}[language=Python]
def port(self, load, size, files):
    ''' Oil column piece
    :param load: crime population
    :param size: unit dark
    :param
\end{lstlisting}
    \par
    is completed by ``files''
    & 8 attention heads & 4-layer toy LM \\
    \midrule

    Greater-Than Operation \par \cite{hanna2024does} & Completes sentences such as ``The war lasted from the year 1732 to the year 17'' predict valid two-digit end years greater than 32 & 8 attention heads and 4 FF sub-layers & GPT2-Small \\
    \midrule

    Gender Bias \par \cite{chintam-etal-2023-identifying} & Completes sentences such as ``The \{profession\} said that'' with gendered pronouns reflecting common stereotypes & 21 attention heads and FF sublayers & GPT2-Small  \\
    \midrule

    Subject-verb Agreement \par \cite{marks2024sparse} & Completes sentences like ``The keys in the cabinet'' with ``are'' as opposed to ``is'' &  $<100$ SAE features & Pythia-70M  \\
    \midrule

    Arithmetic Calculation \par \cite{nikankin2024arithmetic} & $1+4 \xrightarrow{} 5$; $8-4 \xrightarrow{} 4$; $2*4 \xrightarrow{} 8$; $8/4 \xrightarrow{} 2$ & 6, 6, 20, and 6 attention heads, along with all FF sub-layers, correspond to Addition, Subtraction, Multiplication, and Division, respectively & Llama3-8B \\

    \bottomrule
    \end{tabular}
    }
    \caption{List of circuits identified in various MI studies. Note that the list does not encompass all circuits discovered by the MI community.} 
    \label{tab: Circuits}
\end{table*}

% \subsubsection{Interpreting LM behaviors}
% Discussion points:
% \begin{itemize}
%     \item Table of circuits
%     \item Circuits at various levels of granularity 
%     \item Scalability of circuit study
%     \item Circuit for the similar task (four-digit versus eight-digit addition)
%     \item Circuit training dynamics
% \end{itemize}

\subsubsection{Interpreting LM components} \label{subsec: findings-lm-comp}
\paragraph{Interpreting Transformer Components} \label{para: components} The study of circuits has also yielded insights into the functionalities of transformer components. 

The \textbf{RS} can be viewed as a one-way communication channel that transfers information from earlier to later layers. Furthermore, \citet{elhage2021mathematical} hypothesized that MHA and FF in different layers write their output in different subspaces of the RS, which prevents interference of information. In addition, \citet{nostalgebraist2020blog} proposed to view the RS as an LM's current ``guess'' for the output, which is iteratively refined layer-by-layer.

 The \textbf{MHA sublayers} are responsible for moving information between tokens, which enables information from other tokens (i.e., context) to be incorporated into each token's representation. \citet{elhage2021mathematical} showed that each attention head in a layer operates independently and can be interpreted independently. Several MI studies have shown that these attention heads seem to have specialized roles. For instance, ``negative heads'' discovered in GPT2-small by \citet{mcdougall2023copy} are responsible for reducing the logit values of the tokens that have already appeared in the context. Other notably identified attention heads include previous token heads \cite{wang2022interpretability}, duplicate token heads \cite{wang2022interpretability}, copying heads \cite{elhage2021mathematical}, induction heads \cite{olsson2022context}, and successor heads \cite{gould2023successor}.

The \textbf{FF sublayers} are attributed for the majority of feature extraction \cite{gurnee2023finding}, storing and recalling pre-trained knowledge \cite{meng2022locating} and arithmetic computation \cite{stolfo2023mechanistic}. More generally, \citet{geva2020transformer} viewed FF sublayers as key-value stores where the outputs of the first layer ($W_k^l$) of the FF sublayer serves as keys that activate values (stored knowledge) within the weight matrices of the second layer ($W_v^l$). Furthermore, they demonstrated that earlier FF layers typically process shallow (syntactic or grammatical) input patterns, while later layers focus more on semantic patterns (e.g., text related to TV shows).

{The \textbf{LayerNorms}~\cite{ba2016layer} are primarily employed to stabilize and accelerate the training of LMs. However, their role in model computation during inference remains less understood. Many MI studies assume that LayerNorm layers do not play a significant role during inference and thus often ignore their computation. To investigate this assumption, \citet{heimersheim2024you} demonstrated that the LayerNorm in a pre-trained GPT2-small can be removed by fine-tuning the model on a small fraction of the training data without the LayerNorm, suggesting that it may not be essential for the core computational process. In contrast, other studies have revealed that LayerNorm can have additional functionalities; for instance, \citet{winsor2020blog} showed that LayerNorm can serve as a general-purpose non-linear function capable of performing various classification tasks, while \citet{stolfo2024confidence} proposed that it is instrumental in implementing confidence regularization within LMs. These findings highlight that while LayerNorm is critical during training, its exact contributions during inference merit further exploration.} 
% \zyc{how is MI used to study LayerNorm?}

\subsection{Universality}
\label{subsec: findings-universality}
We summarize the findings on the universality of features and circuits below. 

\subsubsection{Universality of Features}
{The extent of feature universality remains largely unexplored, with no definitive conclusions. \citet{gurnee2024universal} found that only 1-5\% of neurons in randomly initialized GPT-2 models exhibit universality, though the polysemantic nature of neurons may have obscured the identification of universal features. To address the problem, recent studies~\cite{lan2024sparse, wang2024towards} have used SAEs to analyze feature universality across LMs, revealing a high degree of similarity in SAE features across models of varying architectures, sizes, and training regimes. Investigating the degree of feature universality and its dependence on factors such as initialization, model size, and loss function remains a critical open challenge.}

\subsubsection{Universality of Circuits}
{Similar to feature universality, studies on circuit universality have yielded mixed results. Early circuit analyses identified components such as induction heads \cite{olsson2022context}, successor heads \cite{gould2023successor}, and duplication heads \cite{wang2022interpretability}, across multiple LMs. 
Similarly, \citet{merullo2023circuit} found that different circuits implementing different tasks (IOI and colored objects tasks) reuse the same components (e.g., induction heads), suggesting a degree of universality of circuits across tasks studied.
% \abu{perhaps we should temper this claim a bit, since this doesn't show the circuits are universal across all tasks, but at least a handful of tasks that were investigated}. 
Additionally, \citet{tigges2024llm} found that specialized attention heads responsible for tasks like indirect object identification (IOI)~\cite{wang2022interpretability}, and greater-than task~\cite{hanna2024does}
% \abu{i think we should use a noun such as `comparison' or something similar, since using an adjective `greater-than' make it a bit more difficult to read}
 consistently emerge while training models of different sizes after a similar number of tokens are processed during training. However, \citet{zhong2024clock} discovered that two LMs trained with different initializations can develop qualitatively different circuits for the modular addition task. Similarly, \citet{chughtai2023toy} found that LMs trained to perform group composition on finite groups with different random weight initializations on the same task do not develop similar representations and circuits. }
\subsection{Findings on Model Capabilities} \label{subsec: model capabilities}

\subsubsection{In-Context Learning (ICL)}
{
ICL is an emergent ability of LLMs that enables them to adapt to new tasks based solely on instructions or a few demonstrations {at inference time} \cite{wei2022emergent}. \citet{elhage2021mathematical} studied a simplified case of ICL
and discovered an \emph{induction circuit} composed of attention heads with specialized roles (e.g., \emph{induction heads}), which were then found to be crucial even for general cases of ICL \cite{olsson2022context, bansal2022rethinking}. Following this, \citet{bietti2024birth} examined how a transformer model balances memorizing general knowledge (\emph{global bigrams}) with adapting to sequence-specific information (\emph{in-context bigrams}). They observed that global bigrams are learned first, followed by the gradual development of an induction head mechanism for handling in-context bigrams. Similarly, \citet{reddy2023mechanistic} showed that the distributional properties of training data (e.g., items in natural data tend to occur in clusters rather than being uniformly distributed over time [\citealt{chan2022data}]) are crucial for the development of induction heads. Further investigations in various synthetic settings have uncovered additional variants of induction heads. For instance, \citet{edelman2024evolution} investigated ICL in bigram models trained on samples from a Markov chain and demonstrated that transformers learn \emph{statistical induction heads} when trained on such data. \citet{ren2024identifying} further investigated few-shot ICL and identified \emph{semantic induction heads}, which, unlike prior induction heads, model the semantic relationship between the input and the output token (e.g., \emph{``I have a nice pen for writing. The pen is nice to'' $\rightarrow$ ``write''}). {Besides the study of induction heads, \citet{hendel2023context} demonstrated that an LM maps the set of demonstrations $\mathcal{S}$ in ICL to a \emph{task vector} that essentially represents the mapping/rule described in $\mathcal{S}$. This task vector is then used by the model to generate task-relevant outputs.}
% \abu{unclear what ``create'' here means; it suggests the vector did not exist before the prompt which i dont think is true; we should rephase} 
} 

\subsubsection{Reasoning}
{
 Recently, sufficiently large LMs have been shown to exhibit reasoning capabilities~\cite{wei2022emergent, huang2022towards}. MI studies have investigated LMs on various reasoning tasks to understand how and to what extent LMs perform reasoning. For instance, \citet{stolfo2023understanding} studied arithmetic reasoning and found that attention heads are responsible for transferring information from operand and operator tokens to the RS of the answer or output token, with FF modules subsequently calculating the answer token. \citet{rai2024investigation} employed a neuron activation analysis to study how Chain-of-Thought (CoT) prompting~\cite{wei2022chain} elicits the arithmetic reasoning capability of LMs. Their study reveals that specific FF neurons encode arithmetic reasoning concepts and that their activation is necessary but not sufficient for arithmetic reasoning. Besides arithmetic reasoning, \citet{dutta2024think} studied CoT multi-step reasoning over fictional ontologies and found that LMs seem to deploy multiple pathways in parallel to compute the final answer. 
Additionally, \citet{brinkmann2024mechanistic} discovered an interpretable algorithm in LM for the task of path-finding in trees. On the other hand, \citet{saparov2024transformers} showed that transformers learn to perform graph search/multi-hop reasoning using a parallel ``path-merging'' algorithm by extracting the circuit for this algorithm; so the embeddings of each fact will include information about facts provable in $k$ hops, where $k$ increases exponentially with the number of layers.
% \abu{plug for a recent paper of mine (https://arxiv.org/pdf/2412.04703) which we could add here; we show that transformers learn to perform graph search/multi-hop reasoning using a parallel ``path-merging'' algorithm by extracting the circuit for this algorithm; so the embeddings of each fact will include information about facts provable in $k$ hops, where $k$ increases exponentially with the number of layers}
{Similarly, \citet{biran2024hopping} also investigated multi-hop reasoning on factual information by employing Patchscopes and showed that while the early layers effectively handle the first-hop reasoning, later layers often fail because they may no longer contain the necessary mechanisms to perform the second-hop reasoning required to extract the correct answer.} 
% \zyc{the following finding reads like an empirical observation, rather than an insight discovered by MI. replace the sentence with, e.g., any mechanism leading to LMs' worse performance in multi-hop reasoning.}
 Finally, \citet{men2024unlocking} identified that the look-ahead planning mechanism in LMs is primarily facilitated by the multi-head attention in middle layers at the last token.} 

\subsubsection{Knowledge Mechanisms} 
{LMs acquire a vast amount of knowledge (e.g., 
facts, grammar, commonsense, concepts, etc.) during pretraining; however, our understanding of how they store, recall, and utilize the knowledge remains incomplete~\cite{wang2024knowledge}. \citet{geva2020transformer} posit that \emph{knowledge is stored in the FF sub-layers}, where the first transformation layer of FF generates keys that activate corresponding values stored in the second FF transformation layer, functioning as a key-value memory system (more detail in Section~\ref{subsec: findings-lm-comp}). In addition, they also employed vocabulary projection methods to discover various kinds of semantic and syntactic knowledge stored in the FF sub-layers. Similarly, recent studies have also shown that factual~\cite{meng2022locating} and commonsense knowledge~\cite{gupta2023editing} are stored in the middle-layer FF sub-layers by employing intervention-based techniques. At a more granular level, specific FF neurons such as arithmetic neurons~\cite{rai2024investigation} and entropy neurons~\cite{stolfo2024confidence} have also been identified to store specific knowledge. In addition to FF sub-layers, recent studies indicate that \emph{knowledge is also stored in the MHA sub-layers}, where each attention head seems to encode a specific type of knowledge~\cite{gould2023successor, geva2023dissecting}. Besides localizing knowledge storage within LMs, \citet{geva2023dissecting} further studied the circuit for the recall of facts or knowledge, i.e., given a subject (e.g., ``Beats music'') and relation (e.g., ``is owned by''), predicting the fact or attribute (e.g., ``Apple''). Their study revealed a three-step mechanism: subject enrichment via FF sub-layers, relation propagation at the end token position via MHA sub-layers, and attribute extraction by later MHA sub-layers. Moreover, \citet{chughtai2024summing} found that the recall of facts in LMs leverages several distinct, independent, and qualitatively different mechanisms or circuits. Although each mechanism alone might not suffice, their additive combination creates constructive interference that ultimately leads to the correct answer.} 

 \subsubsection{Others} As listed in Section~\ref{subsec: findings-circuits}, prior work has also studied LM capabilities in tasks such as IOI \cite{wang2022interpretability}, modular addition \cite{nanda2023emergent}, and greater-than operations \cite{hanna2024does}, leading to the discovery of circuits that implement these tasks. Compared with the interpretation of ICL and reasoning, these studies not only justified the rationale of a capability but also revealed its underlying algorithm through circuits.

 \subsection{Findings on Learning Dynamics} \label{subsec: learning dynamics}

\subsubsection{Phase Changes during LM Training}
Prior studies have observed sudden shifts in LMs' capabilities, called ``phase changes'' \cite{olsson2022context, power2022grokking, wei2022emergent}. These changes are considered key steps during LM training.
MI has been applied to examine the relationship between the emergence of features and circuits and these phase changes. For example, \citet{olsson2022context} found correlations between phase changes and the formation of induction circuits, suggesting that the development of these circuits underlies the phase change. 
In the task of symbol manipulation, \citet{nanda2023progress, varma2023explaining} discovered a similar correlation contributing to LM grokking \cite{power2022grokking}, a phenomenon of LM generalizing that can occur when they are trained beyond overfitting. 
\citet{chen2024learningdynamics} found that sudden drops in the loss during training correspond to the acquisition of attention heads that recognize specific syntactic relations. {Similarly, \citet{wang2024grokked} examined the grokking phenomenon in implicit reasoning tasks and found that LMs can perform implicit reasoning through grokking, where the gradual formation of a generalizing circuit drives the process.} Finally, \citet{huang2024unified} provided a unified explanation for grokking, double descent \cite{nakkiran2021deep}, and emergent abilities \cite{wei2022emergent} as a competition between memorization and generalization circuits. 

% New papers: \citet{wang2024grokked, zhu2024critical, thilak2022slingshot, liu2022towards, doshi2023grok}

% \paragraph{Learning Dynamics during LM Fine-Tuning}
\subsubsection{Effects of Post-training}
{MI studies have investigated how fine-tuning enhances LM capabilities by analyzing underlying mechanistic changes in task-relevant circuits, LM representations, etc. These studies suggest that fine-tuning does not fundamentally change the mechanisms but enhances existing ones. For instance, \citet{jain2023mechanistically} observed that these changes are often localized in a few model weights and that simply pruning them can restore lost pre-training capabilities. Similarly, \citet{jain2024makes} examined the effects of safety fine-tuning and found that it minimally alters the model, with the fine-tuning primarily adjusting FF weights to push unsafe inputs into a ``null space'', thereby making the model significantly less sensitive to unsafe inputs. \citet{jiang2024interpretable} examined catastrophic forgetting in fine-tuning, suggesting that it occurs not by erasing pre-existing skills but by introducing specialized reasoning patterns that overshadow the original capabilities. {On the other hand, \citet{lee2024mechanistic} investigated
% \abu{we should go through the paper carefully and make sure that we are using tense consistently; should we use past tense everywhere? or present tense? i think present is more common in similar papers, but past everywhere is fine too} 
how direct preference optimization (DPO)~\cite{rafailov2023direct} alters the internal mechanisms of LMs to reduce toxicity. They found that DPO does not eliminate the LM's ability to generate toxic outputs but instead reduces toxicity by learning to bypass regions in the model associated with toxic behavior. These findings imply that models aligned with DPO remain vulnerable to being ``jailbroken'' or reverted to an unaligned state, as the underlying toxic vectors persist and can be reactivated.}} 
% \zyc{add \citet{lee2024mechanistic} (work published in 2024; highly cited)}

% New works: \citet{kotha2023understanding, jain2023mechanistically, lee2024mechanistic, jain2024makes, jiang2024interpretable} more work listed in 
% https://www.alignmentforum.org/posts/mFAvspg4sXkrfZ7FA/deep-forgetting-and-unlearning-for-safely-scoped-llms

\subsection{Applications of MI} 

\subsubsection{Model Enhancement} \label{subsec: enhancement}
\paragraph{Knowledge Editing} \label{para: editing}
LMs are known to store factual knowledge encountered during pre-training \citep{petroni2019language, cohen2023crawling}. For instance, when an LM is prompted with ``The space needle is in the city of'', it may retrieve the stored facts and correctly predict ``Seattle''. However, these stored facts may be incorrect or outdated over time, leading to factually incorrect generation \cite{cohen2024evaluating}. 
MI has been found to be a helpful tool for addressing the problem, including understanding where and how facts are stored within LMs, how they are recalled during inference time, and providing approaches for knowledge editing~\cite{meng2022locating, meng2022mass, geva2023dissecting, sharma2024locating}.
For instance, \citet{meng2022locating} used activation patching to localize components that are responsible for storing factual knowledge, and then edited the fact (e.g., replacing ``Seattle'' with ``Paris'') by only updating the parameters of those components.
 
\paragraph{LM Generation Steering} \label{para: steering}
LM generation steering involves controlling an LM's output by manipulating its activations at inference time. For instance, \citet{geva2022transformer} proposed a method to suppress toxic language generation by identifying and manually activating neurons in FF layers responsible for promoting non-toxic or safe words. Similarly, \citet{templeton2024scaling} identified safety-related features (e.g., unsafe code, gender bias) and manipulated their activations to steer the LM towards (un)desired behaviors (e.g., safe code generation, unbiased text generation). Additionally, \citet{nanda2023emergent} demonstrated that an LM's output can be altered (e.g., flipping a player turn in the game of Othello from YOURS to MINE) by pushing its activation in the direction of a linear vector representing the desired behavior, which was identified using a linear probe.

\subsubsection{AI Safety} \label{subsec: safety}
{AI safety is an important concern that MI aims to address. Although LMs undergo multiple rounds of safety fine-tuning, they frequently exhibit low reliability and susceptibility to harmful prompts~\citep{casper2023open, macdiarmid2024sleeperagentprobes}. Within MI, ``enumerative safety'' aims to address AI safety by enumerating all features in LMs and inspecting those related to dangerous capabilities or intentions \cite{elhage2022superposition, jermyn2023anthropic}. To this end, \citet{templeton2024scaling} identified several safety-relevant features that not only activate when the LM exhibits specific behaviors but also causally influence the LM’s output; however, the specific circuits that use these features to implement the behavior have not yet been identified. Similarly, \citet{geva2022transformer} discovered several non-toxic neurons and promoted them to steer the LM's generation of non-toxic tokens. Recently, a new line of research in MI has centered on \emph{latent space monitoring} to detect and prevent harmful model behaviors. Specifically, rather than inspecting the input and output texts for harmful model behaviors, this approach analyzes the latent representation in the model. To this end, these approaches have identified several safety-related linear directions within model latent representations that are useful for monitoring and limited control of AI safety-related behaviors such as truthfulness~\cite{burger2024truth}, refusal~\cite{arditi2024refusal}, and jailbreaking~\cite{ball2024understanding, li2024inference}. Finally, circuit-level insights have proven useful for detecting prompt injection attacks~\citep{belrose2023eliciting} and identifying adversarial examples in tasks such as IOI~\citep{wang2022interpretability}. }

% At present, the exact role MI can play in addressing AI safety is unclear \cite{casper2023interpretability}. Within MI, ``enumerative safety'' aims to address AI safety by enumerating all features in LMs and inspecting those related to dangerous capabilities or intentions \cite{elhage2022superposition, jermyn2023anthropic}. To this end, \citet{templeton2024scaling} identified several safety-relevant features that not only activate when the LM exhibits specific behaviors but also causally influence the LM’s output; however, the specific circuits that use these features to implement the behavior have not yet been identified. 
% As we have discussed, \citet{geva2022transformer} encouraged language safety by steering the LM's generation of non-toxic tokens.
% Finally, insights from circuit studies were used to detect prompt injection \cite{belrose2023eliciting} and find adversarial examples for the IOI task \cite{wang2022interpretability}.

\subsubsection{Others}
Insights from MI have also been used for other downstream tasks. For example, \citet{marks2024sparse} improved the generalization of classifiers by identifying and ablating spurious features that humans consider to be task-irrelevant. \citet{geva2022transformer} proposed self-supervised early exit prediction for efficient inference, drawing insights from their investigations of the FF sublayers' role in token prediction.

\subsection{Benchmarks for Evaluating Interpretability Techniques} \label{sec: tech-benchmarks}

{\subsubsection{Benchmarks for Feature Study Techniques} For feature study, several benchmarks have been developed to evaluate how well various techniques (discussed in Section~\ref{sec: techniques}) can identify meaningful feature vectors for specific concepts, i.e., \emph{targeted feature study}. For instance, CausalGym~\citep{arora-etal-2024-causalgym} introduces a suite of linguistic tasks that test the effectiveness of the technique in discovering the feature vector corresponding to a given concept, and shows that DAS outperforms alternative approaches such as probing. Similarly, Resolving Attribute–Value Entanglements in Language Models (RAVEL)~\citep{huang2024ravel}  
% \yilun{Maybe spell out the full name in the first mentioning, and same for MIB and FIND below.} 
benchmarks targeted feature study methods on their ability to localize and disentangle specific attributes of entities (e.g., ``Paris is on the continent of''). Their results again show that DAS achieves the strongest performance, outperforming methods like SAE and probing. Mechanistic Interpretability Benchmark (MIB)~\citep{mueller2025mib} is another benchmark that provides a broader benchmark covering both targeted and open-ended feature study. Their benchmark also finds DAS to be the best technique for isolating the feature vector for a given concept. For open-ended feature study, however, MIB reports a surprising finding: SAE features are no more effective than individual neurons at isolating meaningful features, contrary to prior claims. SAEBench~\citep{karvonen2025saebench} is another benchmark that provides a comprehensive evaluation suite for comparing various SAE architectures and training setups. Specifically, it proposes to evaluate SAEs across eight diverse metrics, including interpretability, feature disentanglement, and practical applications such as unlearning. Their benchmark results indicate that Matryoshka SAEs achieve the best overall performance, particularly in feature disentanglement. Finally, Function INterpretation and Description (FIND)~\cite{schwettmann2023find} is a benchmark that focuses on the \emph{Explain} step of open-ended feature study, evaluating the correctness of interpretability LM agents in describing the latent functions implemented by model components.}

{\subsubsection{Benchmarks for Circuit Study Techniques} Several benchmarks have also been proposed to evaluate the effectiveness of techniques for the \emph{localization step} of circuit study. Some works employ collections of synthetic~\citep{lindner2024tracr} or semi-synthetic~\citep{gupta2024interpbench} transformers with known circuits, where the availability of ground truth makes evaluation more straightforward. Results from these studies show that ACDC and EAP-IG achieve the strongest performance. However, concerns remain that such synthetic benchmarks may not faithfully capture the behavior of standard pre-trained transformers. To address this, MIB~\citep{mueller2025mib} evaluates circuit localization techniques on both semi-synthetic and standard pre-trained transformer models, where they also found that EAP-IG outperforms other approaches like edge patching.}
\section{Discussion and Future Work}
\label{sec: discussion-future-works}

% \zyc{I suggest making the subsections broader and more generic for easier extension in the future. I personally don't like using a very specific topic as the subsection title, unless you are determined to sell this topic. What I suggest to include in this section: 
% - Connection to broader field: make sense, but i don't know how it will be different from what you had in the background section;
% - Advancement in Techniques: echo what we had in Techniques section
% - Automated hypothesis gen: same as now
% - Practical utility: consider merging the study on complex tasks and LLms in? 
% - benchmarks and metrics
% - human-AI collaboration
% - beyond MI: illusion and proposition interp there?
% }

% \input{Discussions-and-Future-Work/connection-broader-field}

% \subsection{Interpretability illusion}

\subsection{Advancement in MI Techniques}
Despite several promising recent advancements in MI techniques, the current set of techniques still faces a number of critical limitations that hinder their effectiveness and general applicability. One of the primary challenges is \textbf{scalability}, as many existing techniques (e.g., SAE, intervention-based techniques) require significant computational resources for interpreting larger LMs and increasingly complex model behaviors (e.g., reasoning). Furthermore, the \emph{heavy reliance on human interpretation} of current techniques (e.g., interpreting SAE results) exacerbates scalability concerns while also introducing subjectivity, raising issues about the reproducibility and reliability of the findings. To address these challenges, developing more \emph{automated techniques} for feature and circuit analysis is a promising direction, as it could reduce human effort while enhancing both the rigor and scalability of interpretability research. Additionally, exploring techniques that are \emph{less computationally demanding}, such as attribution patching~\cite{nandaattribution}, is another promising research direction. 
Beyond the scalability concern, the field
has often focused on individual techniques in isolation rather than \textbf{combining complementary techniques} to improve their applicability and mitigate limitations of individual techniques. For instance, Patchscopes~\cite{ghandeharioun2024patchscope} combines intervention-based and vocabulary-based techniques to obtain more expressive techniques, as discussed in Section~\ref{tech: intervention-methods}. 
% Finally, each MI technique comes with its own set of challenges, 
Finally, \textbf{addressing challenges of individual techniques},
as outlined under \emph{Technical Advancements} in Section~\ref{sec: techniques}, is crucial for these techniques to reliably and effectively serve MI studies. For instance, while intervention-based methods frequently suffer from out-of-distribution issues due to input corruption methods, probing techniques are inherently correlational and lack causal explanatory power. Exploring new techniques that are more efficient, automated, and integrated, while also addressing the limitations of existing techniques is essential for advancing the field and ensuring the reliability and applicability of MI insights.

\subsection{Practical Utility of MI Studies}
While the importance of deriving actionable insights from MI studies for downstream applications is widely recognized \cite{doshi2017towards}, most MI studies often showcase these applications only as extrinsic evaluations on toy tasks, lacking rigorous comparisons against alternative methods or baselines. Furthermore, most MI studies, especially circuit studies, involve investigating simple LM behavior without downstream applications, often criticized as ``streetlight interpretability'' \cite{bereska2024mechanistic, casper2023streetlight, wang2022blog}. 
% For example, \citet{wang2022blog} deliberately selected the IOI task \cite{wang2022interpretability} due to its simplicity as an algorithmic problem. 
% \zyc{I don't see the streetlight interpretability definition in Casper 2023; is this is a wrong reference? Also, the "For example" sentence reads incomplete. I also don't think that we should cite a non-published blogpost when discussing a general phenomenon of a study field.} 
Similarly, although some studies have been conducted on ``production-level'' LMs \cite{lieberum2023does, templeton2024scaling}, the majority of studies still investigate smaller LMs, raising concerns about generalizability given the mixed findings on universality. Despite these challenges, a few MI studies have demonstrated promising practical downstream applications. For instance, recent studies have shown the potential of MI techniques in enhancing AI safety, notably through red-teaming efforts to identify model vulnerabilities~\cite{arditi2024refusal, casper2023red}. Additionally, several studies have also leveraged MI to develop new techniques~\cite{zou2024improving, ashuach2024revs, guo2024mechanistic, pochinkov2024dissecting} and perform evaluation~\cite{deeb2024unlearning, hong2024intrinsic, lynch2024eight} on tasks such as machine unlearning~\cite{liu2025rethinking} and knowledge editing~\cite{wang2024knowledge}. Finally, \citet{kitouni2024neurons} explored whether mechanistic approaches can uncover scientific knowledge embedded in models trained on prediction tasks. In summary, the advancement of MI research will benefit from a stronger emphasis on practical utility, which is crucial for reinforcing the field’s credibility and demonstrating its real-world impact.

\subsection{Standardized Benchmarks and Metrics}
Evaluating interpretability results is inherently challenging due to the lack of ground truth \citep{zhou2022feature}, and current MI studies also employ various \textit{ad hoc} evaluation approaches, potentially leading to inconsistent comparisons \cite{zhang2023towards}. On the other hand, there have been proposals for standardized evaluation benchmarks. RAVEL \cite{huang2024ravel} evaluates techniques (e.g., SAEs) that disentangle polysemantic neurons into monosemantic features. {Tracr \cite{lindner2024tracr} converts human-readable RASP programs \cite{weiss2021thinking} into transformer models, enabling the construction of models with known ground-truth features and algorithms. These synthetic models provide a controlled setting to validate feature and circuit discovery algorithms, allowing researchers to assess whether these algorithms can accurately identify the known ground-truth features and mechanisms.}
% \abu{what does it mean to ``enable'' a feature/algorithm? did you mean that it enables the identification of features/algorithms?} \daking{} 
 However, these proposed benchmarks are still insufficient \cite{rauker2023toward}. {For instance, features and algorithms identified on synthetic transformers in the Tracr benchmark may not generalize to naturally-trained transformers.}
 % \abu{what does success mean here? successful identification of features/algorithms?} 
  Thus, more effort is needed to develop standard evaluation techniques and to ensure their wide adoption.

\subsection{Automated Hypothesis Generation in MI Practices}
% At a high level, MI research typically involves two key stages: generating hypotheses about the underlying mechanisms of LM behavior and validating those hypotheses. 
{MI research often requires researchers to propose a hypothesis about the underlying mechanisms of LM behavior. However, this can be a laborious and non-scalable process. For example, to formulate plausible hypotheses while interpreting LM components in a circuit study, researchers need to manually analyze the activation patterns or intervention results (Section~\ref{subsec: circuit-workflow}).}
% For example, interpreting LM components in a circuit study, as discussed in Section~\ref{subsec: circuit-workflow}, requires human researchers to formulate plausible hypotheses by analyzing activation patterns or intervention results. 
This reliance on human intuition poses a potential scalability bottleneck, particularly when studying large-scale LLMs and complex behaviors. To mitigate this challenge, developing automated hypothesis-generation methods, potentially with human oversight, will be crucial for improving efficiency and scalability in MI research.
% At a high level, MI study is a process with two stages: generating hypotheses on the underlying mechanisms of LM behavior and validating them. Although various techniques have automated hypothesis validation (Section~\ref{subsec: automated-techniques}), the generation part is mainly left to humans, a potential scalability bottleneck to LLMs and complex behaviors. To address it, automated hypothesis generation, potentially with humans in the loop, is instrumental.

% \subsection{Studies on Complex Tasks and LLMs} Current MI studies are mostly performed on simpler tasks, often criticized as ``streetlight interpretability'' \cite{casper2023interpretability, wang2022blog}. For instance, \citet{wang2022blog} intentionally selected the IOI task \cite{wang2022interpretability} because it is a simple algorithmic task.
% Similarly, although a few studies were done on ``production-level'' LMs \cite{lieberum2023does, templeton2024scaling}, most still used small LMs, which may have limited generalizability, given the mixed results on universality. 

\subsection{Generalizing Beyond MI for More Intuitive Applications} \label{propositional-interp}
% Enumerative safety in mechanistic interpretability aims to enumerate all features in a model and inspect whether there are any features related to unsafe capabilities or intentions encoded in the model representations~\citep{elhage2022superposition}. However, a feature is not inherently safe or unsafe. For example, if a model activates the features representing ``kill'' and ``humans''
% \abu{using simple strings for each feature is very non-descriptive, and very incongruent with our definition of ``feature'' in 3.1.1; there, we define it as a property, for example, of an input token/concept. but perhaps this definition is too narrow? to be honest, i am not sure i understand the definition of a ``kill'' or ``human'' feature, other than that the input may describe an object that has the property ``human''}
% , it could be representing ``kill a human'' or ``don’t kill a human''—two concepts with opposite meanings. 

{While MI was initially defined around the three objects as we described in Section~\ref{sec:objects of study}, recent research has generalized beyond this scope for more intuitive applications. For instance, while existing feature studies have focused on decoding low-level input properties as features from the activations of LMs, recent works by \citet{feng2023language, chalmers2025propositional} have called for decoding higher-level \emph{proposition} from the activations. Consider finding safety-relevant features for enumeration safety~\cite{elhage2022superposition} as an example. While being able to discover features encoding properties such as ``kill'' is helpful, this discovery alone cannot describe the safety propensity of the model; rather, higher-level propositions such as ``I support killing humans'' or ``I oppose killing humans'' can more directly allow one to measure the safety concern of an LM.}
A notable example of work in this direction is \citet{feng2023language}, which investigates how individual features are combined to form higher-level concepts and how these concepts are represented within model activations. Specifically, it 
identified \emph{binding vectors}—a mechanism that enables LMs to encode associations such as \texttt{lives(Alice, Paris)} and \texttt{lives(Bob, Bangkok)} when given input like ``Alice lives in the capital city of France. Bob lives in the capital city of Thailand.'' Understanding how models represent features at a higher level of abstraction seems critical for a more comprehensive and intuitive interpretation of LM behavior, highlighting the necessity for further research in this area.

\subsection{A New Paradigm of Human-AI Collaboration Driven by MI}
{Several studies have explored the utility of post-hoc feature-attribution explanations~\citep{ribeiro2016should, lundberg2017unified, sundararajan2017axiomatic} for improving human-AI team performance. However, most of these studies~\citep{bansal2021does, chen2023understanding, carton2020feature, rai2024understanding} have found that these feature-attribution explanation fails to improve human-AI team performance and instead lead to a decline in performance compared to stand-alone AI.
% , primarily due to users' over-reliance on explanations. 
{This decline in performance, however, may be attributed to the limitations of feature-based explanations, which simply highlight important input tokens as the model explanation without offering deeper insights into the model’s decision-making process. In contrast, MI provides a more comprehensive understanding by analyzing the model’s internal mechanisms, uncovering what features are extracted from the input tokens and how they influence the final output. Besides post-hoc explanation methods, textual explanations such as chain-of-thought (CoT) have also been proposed to be leveraged for model explanation; however, recent research suggests that CoT explanations can be unfaithful and misleading~\cite{turpin2023language}. In light of these findings, MI not only has the potential to provide the most faithful explanations as it mostly relies on causal techniques but also presents a new type of ``mechanistic explanation'', opening up new potential for human-AI collaboration.} 
% \zyc{If you consider "users' over-reliance on explanations" is the primary limitation, how would MI mitigate this issue? I also suggest discussing a bit about textual explanations (e.g., CoT) vs. MI. Textual explanations are naturally easier for people to digest and many people have questioned the need for MI. We can cite \url{https://arxiv.org/pdf/2305.04388} as an example showing that textual explanations may not be faithful. }
 However, limited work has been done to investigate whether the mechanistic explanation generated through MI analysis can be used to improve human-AI team performance. One notable effort in employing the mechanistic understanding of LM behavior to improve human-AI team collaboration is provided by \cite{marks2024sparse}. In this work, the authors proposed Spurious Human-interpretable Feature Trimming (SHIFT), a technique that enhances classifier generalization by removing the spurious features with the help of human evaluators. Their experiments demonstrate that SHIFT not only eliminates unintended biases but also improves the classifier's overall performance. However, the practical application of MI findings such as SHIFT have only been considered as an extrinsic evaluation to validate MI studies, and as a result, they have been tested in only limited domains and benchmarks.}

\section*{Acknowledgements}
DR and ZY were sponsored by the National Science Foundation (\#2311468/\#2423813) and the Department of Computer Science at the College of Computing and Engineering at George Mason University. We appreciate comments from readers of the earlier version of this paper, and the clarification from Johnny Lin on the usage of Neuronpedia.

\bibliography{tmlr}
\bibliographystyle{tmlr}

\appendix

\end{document}